\documentclass[final]{cvpr}

\usepackage{times}
\usepackage{epsfig}
\usepackage{graphicx}
\usepackage{amsmath}
\usepackage{amssymb}

\usepackage{dsfont}
\usepackage{multirow}
\usepackage{array}
\usepackage{subcaption}
\usepackage{longtable}

\usepackage[pagebackref=true,breaklinks=true,colorlinks,bookmarks=false]{hyperref}

\def\cvprPaperID{5433} 
\def\confYear{CVPR 2021}

\begin{document}

\title{Sewer-ML: A Multi-Label Sewer Defect Classification Dataset and Benchmark}

\author{Joakim Bruslund Haurum\qquad Thomas B. Moeslund\\
Visual Analysis and Perception (VAP) Laboratory, Aalborg University, Denmark\\
{\tt\small joha@create.aau.dk, tbm@create.aau.dk}

}

\maketitle

\begin{abstract}
Perhaps surprisingly sewerage infrastructure is one of the most costly infrastructures in modern society. Sewer pipes are manually inspected to determine whether the pipes are defective. However, this process is limited by the number of qualified inspectors and the time it takes to inspect a pipe. Automatization of this process is therefore of high interest. So far, the success of computer vision approaches for sewer defect classification has been limited when compared to the success in other fields mainly due to the lack of public datasets. To this end, in this work we present a large novel and publicly available multi-label classification dataset for image-based sewer defect classification called Sewer-ML.

The Sewer-ML dataset consists of 1.3 million images annotated by professional sewer inspectors from three different utility companies across nine years.
Together with the dataset, we also present a benchmark algorithm and a novel metric for assessing performance. The benchmark algorithm is a result of evaluating 12 state-of-the-art algorithms, six from the sewer defect classification domain and six from the multi-label classification domain, and combining the best performing algorithms. The novel metric is a class-importance weighted F2 score, $\text{F}2_{\text{CIW}}$, reflecting the economic impact of each class, used together with the normal pipe F1 score, $\text{F}1_{\text{Normal}}$. The benchmark algorithm achieves an $\text{F}2_{\text{CIW}}$ score of 55.11\% and $\text{F}1_{\text{Normal}}$ score of 90.94\%, leaving ample room for improvement on the Sewer-ML dataset. The code, models, and dataset are available at the project page \url{http://vap.aau.dk/sewer-ml}

\end{abstract}

\section{Introduction}

\begin{figure}
\centering
\begin{subfigure}{.5\linewidth}
  \centering
  \includegraphics[width=.99\linewidth]{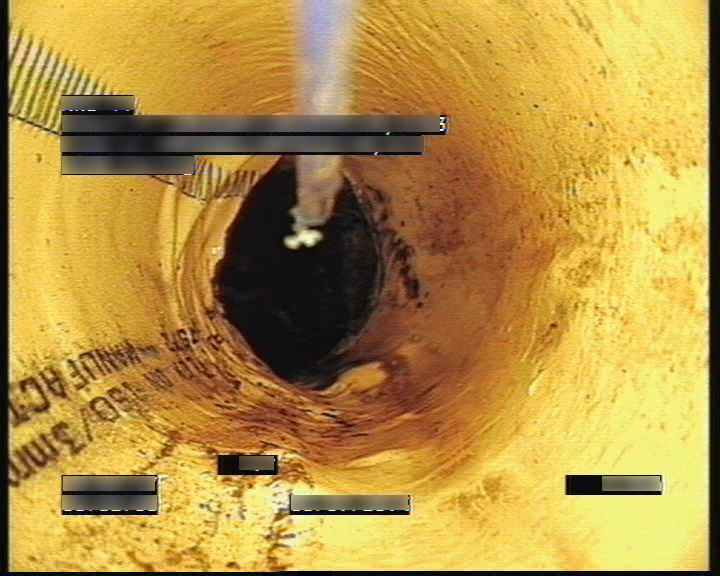}
  \caption{DE, PF}
  \label{fig:DEPF}
\end{subfigure}%
\begin{subfigure}{.5\linewidth}
  \centering
  \includegraphics[width=.99\linewidth]{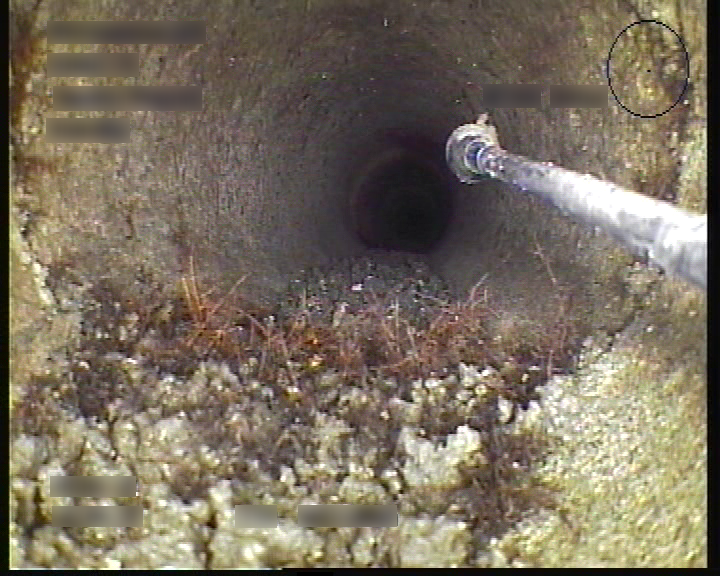}
  \caption{RO, FS, AF}
  \label{fig:ROFSAF}
\end{subfigure} %
\begin{subfigure}{.5\linewidth}
  \centering
  \includegraphics[width=.99\linewidth]{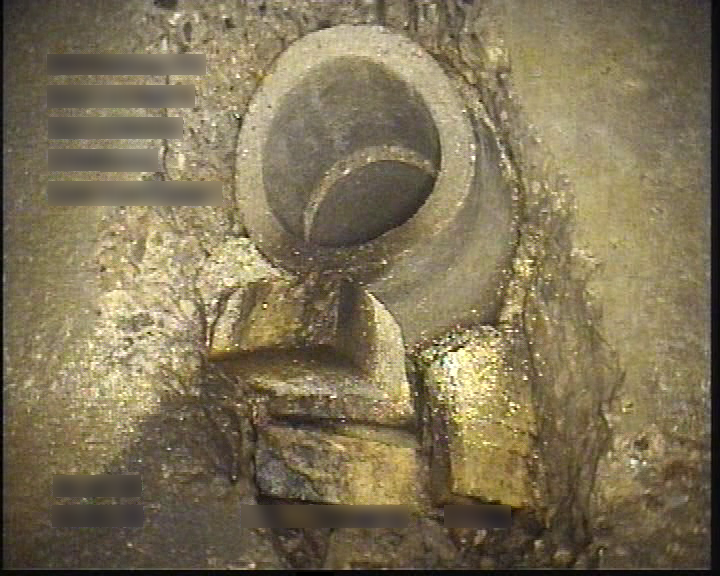}
  \caption{PH, RB}
  \label{fig:PHRB}
\end{subfigure}%
\begin{subfigure}{.5\linewidth}
  \centering
  \includegraphics[width=.99\linewidth]{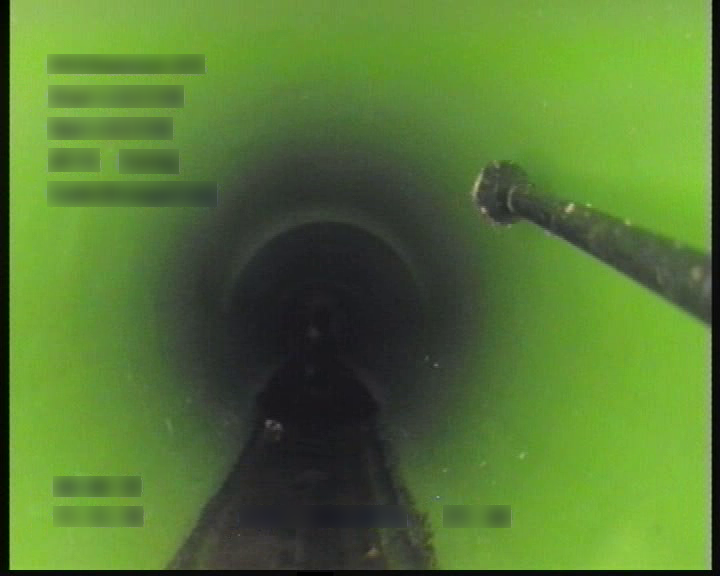}
  \caption{Normal pipe}
  \label{fig:ND}
\end{subfigure}
\caption{\textbf{Sewer-ML data examples.} Images showcasing a subset of the classes and the visual variation in the dataset. The class codes below each image are described in Table~\ref{tab:ClassOverview}.}
\label{fig:examples}
\end{figure}

The sewerage infrastructure is an important but often unnoticed infrastructure. 240 million US citizens are serviced by 1.28 million kilometers of public sewer pipes and 800,000 kilometers of privately owned pipes \cite{ASCEReportCard2017}. In order to maintain public health and sanitation, and avoid \eg unintentional sewer overflows, a 271 billion dollar investment is needed within the next 10 years in order to service an additional 56 million US citizens \cite{ASCEReportCard2017}. Additionally, all of these sewer pipes have to be regularly inspected to avoid sudden pipe collapse or reduced sewer capabilities.

Sewer inspections are currently performed on location by a professional inspector, who simultaneously maneuvers a remote controlled vehicle with a movable camera through the sewer pipe. This is hard and tiresome work, as the inspectors must look at a video feed for a prolonged amount of time. This can lead to flawed inspections, which in the worst case can result in damage to the sewerage infrastructure. Furthermore, the variance in visual appearance within sewer pipes further complicates the task, see Figure~\ref{fig:examples}.

Therefore, the field of automated sewer inspection has been researched by industry and academia for the last three decades, through the development of different robot platforms and specialized algorithms \cite{HaurumSurvey2020}. However, there are at the moment no means to determine which method is the best. Haurum and Moeslund \cite{HaurumSurvey2020} found that there are no open-source benchmark datasets, little to no open-source code, and no agreed upon metrics or evaluation protocol. Instead, many researchers utilize their own datasets from different countries and follow different inspection guides. This leads to stagnation in the field when compared to other computer vision fields and a lack of reproducibility in the automated sewer inspection field.

For these reasons, we present the open-source \textit{Sewer-ML} multi-label defect dataset, containing 1.3 million images annotated by professional sewer inspectors. The dataset is collected from three different Danish water utility companies over a period of nine years. Our contributions are fourfold:
\begin{itemize}
    \itemsep0em
    \item A publicly available multi-label sewer inspection dataset with 1.3 million annotated images.
    \item An open-source comparison of state-of-the-art methods using the new dataset.
    \item A novel, class-importance weighted F2 metric, $\text{F}2_{\text{CIW}}$.
    \item A benchmark algorithm combining knowledge from sewer defect and multi-label classification domains.
\end{itemize}

The paper is structured as follows. In Section~\ref{sec:related}, we review the related works within the multi-label image classification and automated sewer inspection fields. In Section~\ref{sec:dataset}, the proposed dataset is introduced and described in detail. In Section~\ref{sec:benchmark}, we introduce our novel metric, test several state-of-the-art methods on the new dataset, and conduct an ablation study on the obtained results leading to our benchmark algorithm. Finally, in Section~\ref{sec:conclusion}, we summarize our findings and conclude the paper.
\section{Related Works}\label{sec:related}
\textbf{Multi-label Image Classification.}  Through the years, the field of multi-label classification has experienced several different trends. Classically, the naive way to approach the problem has been to use an ensemble of binary classifiers and ignore label correlations \cite{MLReview}. This approach has been replaced by methods consisting of a single model incorporating the label correlations into the method itself. These trends have included ranking the label predictions \cite{DLRanked13, Ranked16, Ranked17}, utilizing object localization techniques and attention mechanisms \cite{MCAR20, MLEvid18, Guo2019, BiModalML20,  WeaklyMLDet18, LSTMAtt17, HCP16, Yang2016ML,RLSD18, Zhu2017}, or incorporating a recurrent sub-network to encode label dependencies \cite{OrderFreeRNN,  MLRL18,BiModalML20,CNNRNN16,LSTMAtt17, Orderless20}. 

Current state-of-the-art networks focus on utilizing the inherent graph nature of the multi-label problem \cite{ SSGRML19, ICME19, MLGCN19, PartialLabel20, ZLML18,KSSNet20,You2020}, by using the co-occurrence matrix between labels in combination with graph convolutional networks (GCNs) \cite{GCN}. Chen \etal \cite{MLGCN19} proposed the ML-GCN method, which combines the output of a two-layer GCN with the last feature map of a ResNet-101 \cite{ResNet} network to achieve a well performing multi-label classifier. Wang \etal \cite{KSSNet20} built upon this idea in their KSSNet model. KSSNet improves the performance over ML-GCN by fusing features from a GCN into the final feature map of each residual block in a ResNet-101 model, using a novel lateral connection module. Furthermore, the GCN adjacency matrix is created by combining the label correlation matrix with a label knowledge graph. Lastly, it is also possible to simply take a network which has been proven to work well on a multi-class classification task and instead train it with a relevant loss objective, often the binary cross-entropy loss. This is the case with the recent work of Wu \etal \cite{TencentML19} who utilized the ResNet-101 architecture and Ridnik \etal \cite{TResNet20} who proposed a modified variation of the ResNet architecture, called TResNet. The TResNet network has outperformed several models designed for the multi-label task. 

The multi-label image classification field has classically worked on smaller datasets such as PASCAL VOC \cite{VOC09}, NUS-WIDE \cite{NUSWIDE09}, and COCO \cite{COCO14}, each containing between 5 to 80 thousand training images and 20-80 classes. Therefore, the applied methods have often relied on pre-training the backbone network on ImageNet \cite{ILSVRC15}. However, recently the Tencent-ML \cite{TencentML19} and Open Images datasets \cite{OpenImages20}, each containing between approximately 6 and 12 million training images and 11 to 20 thousand classes, have been proposed. These datasets allow for training methods directly on the multi-label task and not pretraining on ImageNet. All of these datasets focus on natural scene images with ``common'' objects. Furthermore, these datasets are often severely imbalanced, as \eg the class ``person'' occurs more frequently than the class ``sheep''. Therefore, there have been attempts to counteract the data imbalance through weighting the loss objective. This has classically been achieved by utilizing some variant of the inverse class frequency, though custom loss objectives have been proposed specifically for the imbalanced data problem \cite{ASL20, EfficientSamples19, FocalLoss17,  DistributedLoss20}.

\textbf{Automated Sewer Inspection.} For several decades there has been an increasing industrial and academic interest in automating the sewer inspection process. This line of research builds heavily upon the general computer vision field, though the current state-of-the-art does not fully utilize the recent advances within computer vision.

Several different types of sensors \cite{Duran2002Survey, Liu2013Survey}, such as acoustic sensors \cite{Iyer2012, Khan2018_2, Khan2018_1}, laser scanners \cite{Lepot2017, Tezerjani2015}, and depth sensors \cite{Alejo2017, Alejo2020, SynthPointClouds2, SynthPointClouds}, have all been utilized for sewer pipe reconstruction and detecting specific defects but have not seen widespread usage in more generalized tasks. Conversely, image and video based approaches have been utilized to detect, segment, and classify a wide variety of sewer defects. Traditionally, hand-crafted features and small, model based classifiers or heuristic decision rules have been utilized \cite{Myrans2018MultiClass, Myrans2018Binary,Ye2019}. However, in recent years deep learning based methods have gained traction within the field. This has led to advances within video processing \cite{ Fang20, Moradi2017, Moradi20, WangKumar21}, water level estimation \cite{HaurumWater, Ji2020}, defect detection \cite{ Cheng18, KumarWang20, Yin20}, segmentation \cite{Pan20, Piciarelli19, Wang20}, and classification in multi-class and multi-label settings \cite{Chen2018, Hassan2019AlexNet,Kumar2018EnsembleCNN, Li2019HierCNN, Meijer2019MultiLabel,  Myrans2018MultiClass, Xie2019HierCNN}. For a full review of the field we refer to Haurum and Moeslund \cite{HaurumSurvey2020}.

Within sewer defect classification there has been a recent increase in interest, focused on three different system settings: a single end-to-end classifier, a two-stage approach consisting of a binary classifier and a multi-class/label classifier, and an ensemble of binary classifiers.  Kumar \etal \cite{Kumar2018EnsembleCNN} utilized an ensemble of binary classifiers to categorize four types of defects using a small, two-layer CNN trained in a one \vs all manner. Hassan \etal \cite{Hassan2019AlexNet} used AlexNet \cite{AlexNet} and Li \etal \cite{Li2019HierCNN} a modified ResNet-18 network \cite{ResNet}, trained in an end-to-end manner. Similarly,  Meijer \etal \cite{Meijer2019MultiLabel} built upon the work of Kumar \etal using a small, three-layer CNN for multi-label defect classification, trained end-to-end. Lastly, Xie \etal \cite{Xie2019HierCNN}, Chen \etal \cite{Chen2018}, and Myrans \etal \cite{Myrans2018MultiClass} all used two-stage approaches. Xie \etal trained two small, three-layer CNNs, where the first CNN determines whether a defect is present, while the second CNN, a fine-tuned version of the first, classifies the defects. Chen \etal, on the other hand, use the lightweight SqueezeNet \cite{SqueezeNet} network for the initial binary defect classification and the deeper InceptionV3 \cite{InceptionV3} network for predicting the defect class. Myrans \etal differ from the other recent methods by using the GIST feature descriptor \cite{GIST} and two Extra Trees \cite{ExtraTrees} classifiers in sequence. 

All prior methods utilize separate private dataset with different classes and class distributions, due to the inherent commercial interest involved in the field \cite{HaurumSurvey2020}.  The datasets are typically either balanced such that the number of observations per class is balanced, the number of normal and defect observations are balanced, or the dataset is not balanced but inherently skewed. For example, Meijer \etal \cite{Meijer2019MultiLabel} utilized a dataset consisting of 2.2 million images, but only 17,663 of those images contain defects. In order to counteract this large imbalance, Meijer \etal increased the number of defective observations by a factor of five through oversampling. 
Additionally, there are no common metric nor evaluation protocol, making fair comparison between methods impossible \cite{HaurumSurvey2020}. All of these factors severely hinder the reproducibility and progress within the field.

\section{The Sewer-ML Dataset}\label{sec:dataset}
In this section, we present how the data was collected (Section~\ref{sec:collection}), how the multi-label ground truth annotations are obtained (Section~\ref{sec:groundTruth}), how the dataset is constructed (Section~\ref{sec:construction}), and how we redact information which is present in the images (Section~\ref{sec:anonymization}). Further dataset insights are presented in the supplementary materials.

\subsection{Data Collection}\label{sec:collection}
A total of 75,618 annotated sewer inspection videos were obtained from three different Danish water utility companies from the period 2011--2019. All videos were annotated by licensed sewer inspectors following a common Danish standard \cite{Standard2010} containing 18 specific classes listed in Table~\ref{tab:ClassOverview}. According to the inspection standard, each class is given a point score representing the economic consequence of the class, which is determined by professionals involved in the sewer inspection field \cite{Standard2005}. We normalize the point scores to the interval $[0, 1]$ by dividing all point scores by the largest one, denoting the new values as the \textit{class-importance weight} (CIW). The collected data span a large variety of materials, shapes, and dimensions from both main and lateral pipes. This leads to a large variety in the available data, reflecting the natural variance observed during actual sewer inspections. 

\begin{table}[!t]
\centering
\caption{\textbf{Sewer inspection classes.} Overview and short description of each annotation class \cite{Standard2010} and the class-importance weights (CIW) \cite{Standard2005}.}
\label{tab:ClassOverview}
\begin{tabular}{l|ll}\hline
\textbf{Code} & \textbf{Description} & \textbf{CIW}                         \\ \hline
VA     & Water Level (in percentages) & 0.0310      \\
RB     & Cracks, breaks, and collapses  & 1.0000      \\
OB     & Surface damage         & 0.5518         \\
PF     & Production error        & 0.2896             \\
DE     & Deformation             &  0.1622            \\
FS     & Displaced joint           & 0.6419           \\
IS     & Intruding sealing material   & 0.1847        \\
RO     & Roots               &  0.3559               \\
IN     & Infiltration       &  0.3131                \\
AF     & Settled deposits     &  0.0811               \\
BE     & Attached deposits    &  0.2275               \\
FO     & Obstacle             & 0.2477                \\
GR     & Branch pipe           & 0.0901               \\
PH     & Chiseled connection    &  0.4167              \\
PB     & Drilled connection     & 0.4167               \\
OS     & Lateral reinstatement cuts & 0.9009         \\
OP     & Connection with transition profile & 0.3829   \\
OK     & Connection with construction changes & 0.4396 \\ \hline
\end{tabular}
\end{table}

\subsection{Multi-Label Ground Truth}\label{sec:groundTruth}
The dataset is constructed by extracting a single frame at each class annotation in a sewer inspection video. Each annotation corresponds to a ground truth annotation of a single class at a specific second in the video, with an associated location within the pipe. We obtain the multi-label representation by combining annotations close to each other in the pipe. This is a noisy approach as the camera can rotate in a hemisphere and does not guarantee that all annotations will be visible. For each annotation in an inspection video, we aggregate the annotated class with all other annotated classes which are up to 0.3 meters earlier in the pipe or 1.0 meters ahead in the pipe. These values have been decided through manual inspection as the position measurement can be noisy. This is necessary in order to include nearby and upcoming, visible classes. 
Lastly, some entries are noted as \textit{continuous}, which means the class occurs frequently within a specified stretch of the pipe, but are not explicitly annotated at each occurrence. We handle this edge case by adding the continuous class to all other annotated class occurrences within the defined pipe stretch.

The 18 classes are not all instances of pipe defects but can also indicate important information such as a change in pipe shape or material, occurrence of a branch pipe or pipe connections. The VA class is a special class, as it is annotated at the start and end of an inspection video, as well as when the water level changes within a $10\%$ step interval. This means all annotations have an associated water level.

\begin{table}[!t]
\centering
\caption{\textbf{Split between defective and normal observations.} Number of images containing normal and defective observations in the three dataset splits.}
\label{tab:binaryOcc}
\resizebox{\linewidth}{!}{%
\begin{tabular}{l|rrr|r}\hline
\textbf{Type}          & \textbf{Training}   & \textbf{Validation} & \textbf{Test}   & \textbf{Total}   \\ \hline
Normal    & 552,820  & 68,681      & 69,221  & 690,722  \\
Defective & 487,309  & 61,365      & 60,805  & 609,479  \\ \hline
Total     & 1,040,129 & 130,046     & 130,026 & 1,300,201 \\\hline
\end{tabular}%
}
\end{table}

Additionally, we obtain observations of cases with no annotated classes, denoted non-defective (ND), using a set of heuristic rules. First, we apply a one meter buffer zone around each annotated class, such that there is at least two meters between annotated classes before ND images can be extracted. If there are any active continuous class between the annotated classes, no ND images are extracted. Furthermore, we enforce that the inspection vehicle may at maximum move 0.25 m/s, calculated based on the time and distance difference between the two classes. This restriction is based on the maximum speed the inspectors are allowed to move the inspection vehicle during an inspection. Lastly, ND images are only extracted when the inspection vehicle is moving forward through the pipe. This condition is checked using the distance information associated with each annotation. If these conditions are met we can extract ND images. In order to avoid duplicate images of the same pipe area, we extract one ND image per meter uniformly sampled between the two annotated classes. The video timestamps of the ND images are calculated using a constant velocity assumption. Examples from the dataset are shown in Figure~\ref{fig:examples} and the supplementary materials.

Moreover, the VA class is special, as it is a continuous entity throughout the video. 
The VA annotations are grouped together with the ND class if there are no other co-occurring labels. This leads to a total of 690,722 images of ``normal'' pipes with no annotated classes and 609,479 images with one or more annotated classes which we call ``defective'', resulting in a total of 1,300,201 images. Lastly, we pose the multi-label classification problem as predicting the class labels in Table~\ref{tab:ClassOverview}, except for the VA class. This means a normal pipe with no class annotations is the absence of any classes. Therefore, it is an \textbf{implicit} class. 

\begin{table}[!t]
\centering
\caption{\textbf{Sewer dataset comparison}. A comparison of datasets used for sewer defect classification and the proposed Sewer-ML dataset. We report  whether the dataset is publibly available (P), the annotations are multi-label (ML), the number of images with defects (DI), images with normal pipes (NI), annotated classes (C), and the Class Imbalance (CI) for each dataset rounded to the nearest integer.}
\label{tab:dataStats}
    \resizebox{\linewidth}{!}{%
\begin{tabular}{l|ccrrrr} \hline
\textbf{Dataset}& \textbf{P} & \textbf{ML} &  \textbf{DI}  &  \textbf{NI}     & \textbf{C} & \textbf{CI}   \\ \hline
Ye \etal \cite{Ye2019}&  &       & 1,045 & 0  & 7 & 13\\
Myrans \etal \cite{Myrans2018MultiClass}&  &     & 2,260   & 0 & 13 & 102\\
Chen \etal \cite{Chen2018} & &      & 8,000  & 10,000  & 5 & 5\\
Li \etal \cite{Li2019HierCNN} & &    & 8,455   & 9,878 & 7 & 19\\
Kumar \etal \cite{Kumar2018EnsembleCNN} &&   & 11,000 & 1,000  & 3 & 4\\
Meijer \etal \cite{Meijer2019MultiLabel} &&\checkmark       & 17,663  & 2,184,919  & 12 & 12,732 \\
Xie \etal \cite{Xie2019HierCNN} & &      & 22,800 & 20,000  & 7 & 8 \\
Hassan \etal \cite{Hassan2019AlexNet} & &    & 24,137  & 0  & 6 & 3\\
\textbf{Sewer-ML} &\checkmark &\checkmark & 609,479 & 690,722 & 17 & 123 \\ \hline
\end{tabular}%
}
\end{table}

\subsection{Dataset Construction}\label{sec:construction}
We construct the dataset by first splitting the data into three splits: training, validation and test. We randomly select videos until 80\% of all annotations are in the training split and the remaining 20\% equally split between the validation and test splits. This leads to 60,356 videos for training, 7,692 videos for validation, and 7,570 videos for testing. This way it is ensured that no images from the same pipe are present between splits.  These splits lead to a near even split of normal and defective observations, see Table~\ref{tab:binaryOcc}.

\begin{figure}[!t]
 \centering
     \begin{subfigure}[b]{\linewidth}
         \centering
         \includegraphics[width=0.9\linewidth]{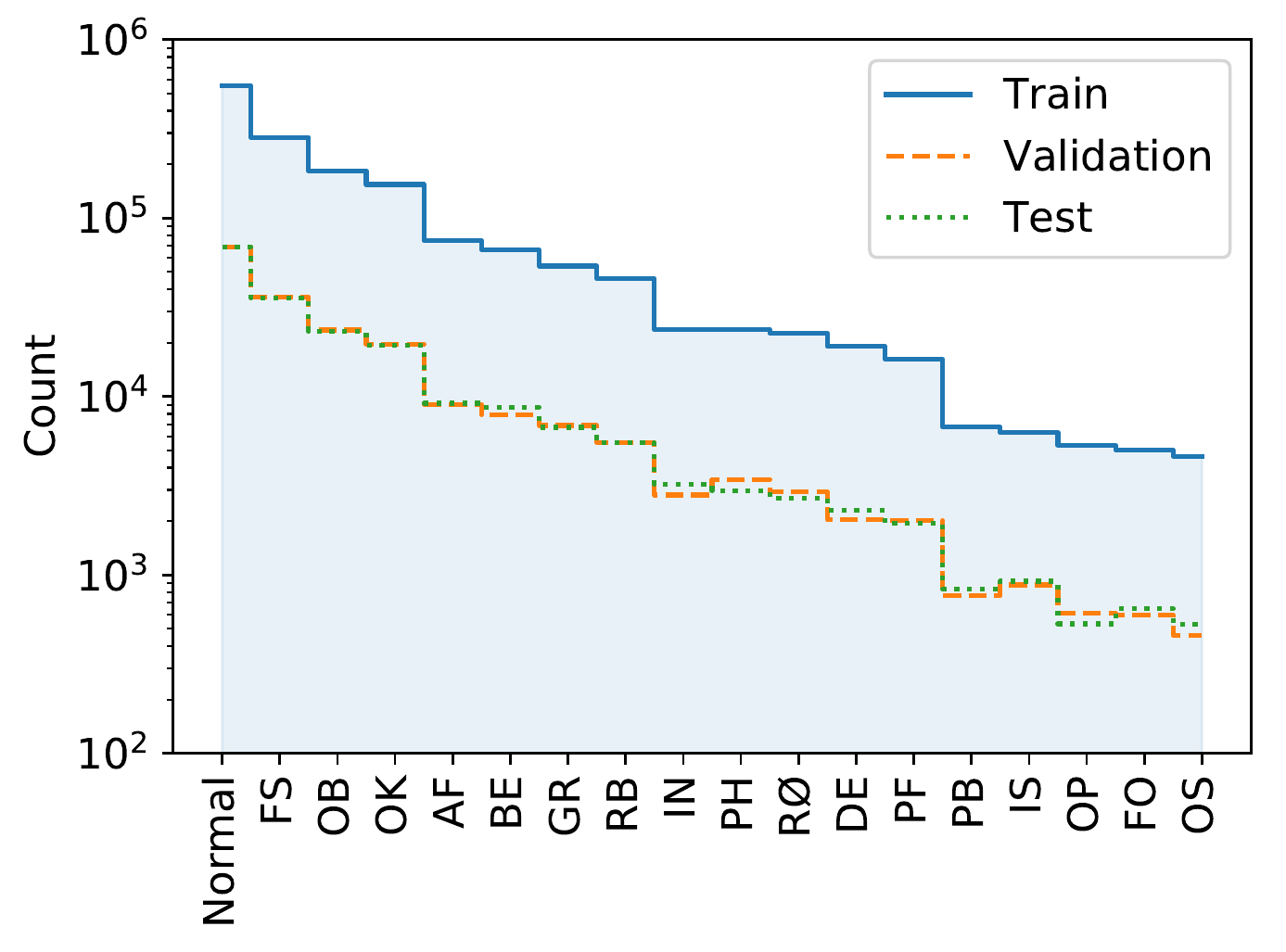}
     \end{subfigure}
     \begin{subfigure}[b]{\linewidth}
         \centering
         \includegraphics[width=0.9\linewidth]{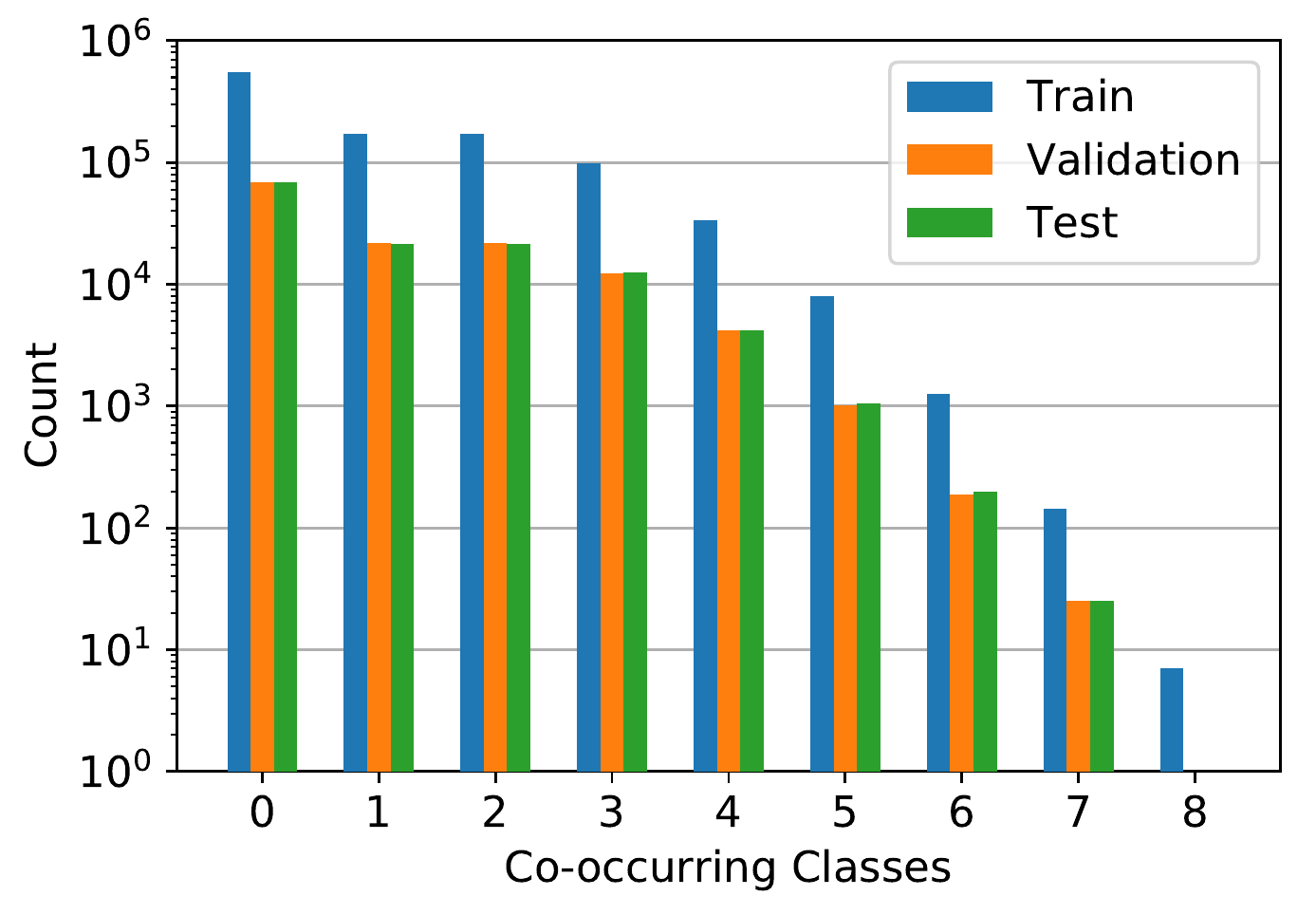}
     \end{subfigure}
     \caption{\textbf{Dataset label statistics.} The frequencies of the annotated classes and the normal class are shown in the top plot in descending order. The frequencies of the number of labeled classes per split are shown in the bottom plot, where ``Normal'' pipes have zero labeled classes. Note that the y-axes are log-scaled.}
     \label{fig:LabelStats}
\end{figure}

Looking at the distribution of the class occurrences, as shown in Figure~\ref{fig:LabelStats}, the occurrences are evenly represented in each split, suggesting a similar class distribution in each of the splits. Moreover, it is evident that the constructed dataset is skewed towards a few major classes, such as the ``Normal'' and ``FS'' classes. This visually shows the large imbalance in the dataset, representative of the real life distribution of the classes. Unlike prior sewer inspection datasets, we do not manually balance the classes.  
We quantify the class imbalance (CI) in the dataset by calculating the ratio between the largest and smallest class and compare with the previously used sewer datasets, see Table~\ref{tab:dataStats}.  Meijer \etal have a large CI due to sampling every five centimeters, resulting in a large number of normal images. Uniquely, Sewer-ML contains a large number of defect images, which are needed to train discriminative classifiers.

Similarly, it is interesting to see how often several classes are present at the same time. For each split, we plot the distribution of the number of labels in the observations in Figure~\ref{fig:LabelStats}. In this plot we count the normal observations as having zero labels as it is an implicit label. We see that there is an equal number of observations with one or two classes and the number of observations reducing as more classes are present. We quantify this using the label cardinality (LC) using Equation~\ref{eq:cardinatlity} \cite{LabelCardinality} for each split. For these measures, we count the normal pipe observations as having one label.

\begin{equation}\label{eq:cardinatlity}
    \text{LC} = \frac{1}{N}\sum_{i=1}^N\sum_{c=1}^{C+1}y_c^{(i)}
\end{equation}
where $N$ is the number of observations in the split, $C$ is the number of annotated classes, and ${y}_c^{(i)}$ is the ground truth value for class $c$ in observation $i$.

We find that across splits, the LC is 1.49-1.50, indicating that on average there are 1.5 labels per observation. We cannot compare this with the LC of the datasets in Table~\ref{tab:dataStats}, as the datasets and ground truth data are not public.

\subsection{Data Anonymization}\label{sec:anonymization}
\begin{figure}[!t]
 \centering
     \begin{subfigure}[b]{0.48\linewidth}
         \centering
         \includegraphics[width=\linewidth]{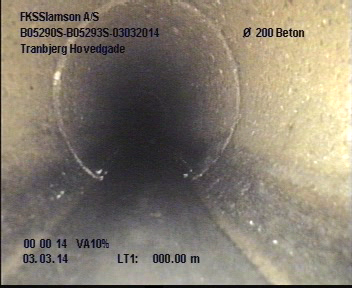}
         \caption{Before text redaction.}
     \end{subfigure}
     \begin{subfigure}[b]{0.48\linewidth}
         \centering
         \includegraphics[width=\linewidth]{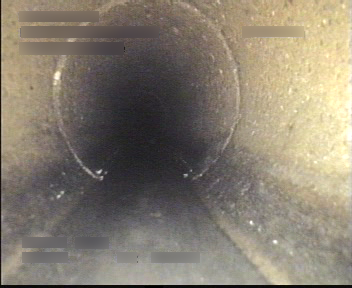}
         \caption{After text redaction.}
     \end{subfigure}
     \caption[]{\textbf{Effect of the data anonymization process.} By applying our text redaction pipeline the system is capable of detecting and blurring all text information on the images.}
     \label{fig:redaction}
\end{figure}

The raw data provided by the water utility companies have all been post-processed by the inspection software to include metadata and annotation text information on the video itself. In order to avoid including ground truth information in the images and any potential privacy issues, the text has been redacted as shown in Figure~\ref{fig:redaction}. Since the overlaid information is not static through the inspection video, due to \eg class codes appearing on screen or pipe material changing, a single redacting mask cannot be used. This leads to a large annotation task, which would be long and tiresome to do manually. Instead, inspired by Borisyuk \etal \cite{Rosetta}, we train a Faster-RCNN \cite{FRCNN} model on examples from the overlaid text data. 23,044 videos are used, with one frame extracted per video. The data is split into a training split of 20,739 images and a validation split of 2,305 images. All text information is manually annotated with bounding boxes. The Faster-RCNN backbone is a ResNet-50 FPN \cite{FPN} pre-trained on ImageNet \cite{ILSVRC15}. We fine-tune the last three residual blocks. As the text data is distinctly different from the data present in the COCO dataset, we use custom anchor boxes. We choose to use three anchor box ratios and five scales, based on the bounding box ratio and area information from the training split.  Full details are available in the supplementary materials. 

Using the COCO metrics \cite{COCO14}, we achieve an mAP@[0.75] of 96.39\% and mAP@[0.5:0.95] of 89.10\%. The Faster-RCNN model is applied on all 1.3 million images in the dataset, and the detected text is removed by applying a Gaussian blur kernel with a radius of 51 pixels.
While this is not a perfect metric score, looking at the detections tells another story. We find that the model detects strings of text, annotated with several bounding boxes, as a single bounding box. An example of this is the ``Ø 200'' in Figure~\ref{fig:redaction}. Similarly, text annotated with a single bounding box, are at times detected with several boxes. This leads to a lower metric score even though the redactions are correct. Therefore, we conclude that data leakage is not an issue in the dataset.

\section{Benchmark}\label{sec:benchmark}
In this section, we present an approach that can be used as benchmarking for future work on the dataset. To this end, we first select (Section~\ref{sec:MLMethods}), train (Section~\ref{sec:train}), and test current state-of-the-art algorithms from the sewer defect classification and the general multi-label classification domains in order to see how they perform on the dataset (Section~\ref{sec:test}), using our novel class-importance weighted F2 metric (Section~\ref{sec:metrics}). Finally, we conduct an ablation study leading to our benchmark algorithm (Section~\ref{sec:ablation}), and discuss the per-class performance (Section~\ref{sec:per-class}). 

\subsection{Methods}\label{sec:MLMethods}

From the sewer inspection domain we compare the methods proposed by Kumar \etal \cite{Kumar2018EnsembleCNN}, Meijer \etal \cite{Meijer2019MultiLabel}, Xie \etal \cite{Xie2019HierCNN}, Chen \etal \cite{Chen2018}, Hassan \etal \cite{Hassan2019AlexNet}, and Myrans \etal \cite{Myrans2018MultiClass}. These six methods are chosen to represent the recent advances within sewer defect classification \cite{HaurumSurvey2020}. For the ensemble of binary classifiers and the end-to-end methods, we train using the full dataset. However, for the two-stage classification approach, the first stage is trained on the full dataset to predict the presence of \textit{any} annotated class, while the second stage is trained to predict the classes from a subset of the data containing annotated classes.

From the general multi-label classification domain, we choose four of the current best performing methods on the COCO and VOC datasets \cite{ASL20}. We choose two state-of-the-art graph-based methods, ML-GCN by Chen \etal \cite{MLGCN19} and KSSNet by Wang \etal \cite{KSSNet20}, each utilizing a ResNet-101 \cite{ResNet} backbone. Furthermore, we also test the vanilla ResNet-101 model, as used by Wu \etal \cite{TencentML19}, and the TResNet architectures from Ridnik \etal \cite{TResNet20}, where we compare the medium, large, and extra-large versions of the model. All models are trained using an end-to-end approach. 

While the normal/defect classification task is intrinsically related to the anomaly detection task, we do not compare with the state-of-the-art anomaly detection methods \cite{AnomDet2021}. This is due to the sewer pipe owners requiring the defect classes in order to correctly manage their assets.

\subsection{Training Procedure}\label{sec:train}
\textbf{Hyperparameters.} In order to ensure comparability, we train all networks from scratch using the exact same training procedures. We base our training procedure on the methodology proposed by Goyal \etal \cite{Goyal17} for efficiently training models on ImageNet. We train each network for 90 epochs, with a batch size of 256 using SGD with momentum. We utilize a learning rate of 0.1, momentum of 0.9, weight decay of 0.0001, and multiply the learning rate by 0.1 at epochs 30, 60, and 80. The ML-GCN and KSSNet networks utilize re-weighted correlation matrices in the GCN subnet, where the hyperparameters stated by the original authors are used. We do not construct a knowledge graph for KSSNet, as the class labels are abbreviations containing little semantic information. We use a one-hot encoding for the initial input to the GCN. For the Myrans \etal \cite{Myrans2018MultiClass} system, the standard GIST hyperparameters are used and the first and second stage classifiers use 100 and 250 trees, respectively, a maximum depth of 10, and $\log_2(d)$ features when splitting nodes, where $d$ is the dimensionality of the GIST feature vector. We find the hyperparameters through a small grid search, described in the supplementary materials.

\textbf{Data augmentation.} The training data are pre-processed by resizing the images to $224\times 224$, horizontally flipping with a 50\% chance, jittering the brightness, contrast, saturation, and hue by $\pm 10\%$ of the original values, and normalizing the data using the training split channel mean and standard deviation. During inference the images are simply resized to $224\times 224$ and normalized. For the InceptionV3 network used by Chen \etal, the images are resized to $299\times 299$ \cite{InceptionV3}. For the GIST features the images are converted to grayscale and resized to $128\times 128$ \cite{Myrans2018MultiClass}.

\textbf{Loss objective.} We train using the standard binary cross-entropy loss, see Equation~\ref{eq:loss}, which is commonly used in the multi-label image classification domain. 
\begin{equation}\label{eq:loss}
    L(\mathbf{x},\mathbf{y}) = \frac{1}{C} \sum_c^C - [w_c y_c \log(\sigma(x_c)) + (1-y_c) \log(1-\sigma(x_c))]
\end{equation}
where $C$ is the number of annotated classes in the dataset, $y_c$ denotes whether class $c$ is present in the current image, $x_c$ is the raw output of the model for class  $c$, $\sigma$ is the sigmoid function, and $w_c$ is the weight for class $c$ if it is present in the current image.

As the dataset is imbalanced, we weight each positive class observation by the negative-to-positive class observation ratio, $w_c$, calculated using Equation~\ref{eq:weight}. This way the loss of minority classes are weighted higher when present in the images, while the loss of majority classes are weighted lower when present. For the InceptionV3 network, a lower weighted loss from the auxiliary classifier is added. 

\begin{equation}\label{eq:weight}
    w_c = \frac{N - N_c}{N_c}
\end{equation}
where $N$ is the number of images in the training split, and $N_c$ is the number of images in the split containing class $c$. 
\label{sec:methods}
\subsection{Metrics}\label{sec:metrics}
Currently, there is no consensus on how sewer defect classification methods should be evaluated \cite{HaurumSurvey2020}. Commonly, the accuracy is used, but this is a poor metric when working with skewed datasets. Moreover, the metrics do not include domain knowledge. Therefore, we evaluate the model performance using two metrics incorporating domain knowledge, based on the $\text{F}\beta$ metric \cite{FBeta},
\begin{equation}\label{eq:FBeta}
    \text{F}\beta = (1 + \beta^2)\frac{\text{Prc} \cdot \text{Rcll}}{\beta^2\text{Prc} + \text{Rcll}}
\end{equation}
where $\text{Prc}$ and $\text{Rcll}$ are the precision and recall of the classifier, respectively, and $\beta$ is a weighting of recall, such that the recall $\beta$ times more important than precision. 

When performing sewer inspections, false negatives have a larger economic impact than false positives. This is due to false negatives possibly leading to faulty pipes going unnoticed, whereas a human will verify the predicted classes before a renovation decision is made. Therefore, it is more important to have a high recall than high precision, if both cannot be achieved. To incorporate this domain knowledge into the evaluation, we set $\beta = 2$ when evaluating the annotated classes. This is similar to previous tasks where recall is weighted higher than precision \cite{Inclusive2018, Planet2017, iMet2019}. The per-class F2-scores are averaged using a novel, class-importance weighted F2-score, $\text{F}2_{\text{CIW}}$. The classes are weighted by the associated CIW, see Table~\ref{tab:ClassOverview}, as classes with a high CIW will be of larger importance for the pipe owners. $\text{F}2_{\text{CIW}}$ is calculated as shown in Equation~\ref{eq:CIWF2}.
\begin{equation}\label{eq:CIWF2}
    \text{F}2_{\text{CIW}} = \frac{\sum_{c=1}^C \text{F}2_c \cdot \text{CIW}_c}{\sum_{c=1}^C \text{CIW}_c} 
\end{equation}
where $\text{CIW}_c$ and $\text{F}2_c$ are the CIW and F2-score for class $c$, respectively, and $C$ is the number of annotated classes.

However, the normal pipes are not included in the $\text{F}2_{\text{CIW}}$ computation, as normal pipes do not have a CIW. In order to quantify whether the tested methods can handle the absence of classes, and not simply maximize the $\text{F}2_{\text{CIW}}$ score by predicting one or more classes at all times, we use the F1-score for the normal pipes, $\text{F}1_{\text{Normal}}$.

\subsection{Model Performances}\label{sec:test}

We report the validation and test split results of each model in Table~\ref{tab:mainRes}. Unless otherwise noted, a threshold of $0.5$ is used to binarize the predictions.  
For the two-stage approaches, the prediction score from the first stage is used for all classes if the binary classifier detects no classes, and otherwise, the score of the second stage network is used. The results are obtained using the model weights from the epoch with the lowest validation loss. In most cases, the lowest validation loss is obtained after 30-40 epochs, whereafter the networks start overfitting. This indicates that while we utilize a dataset nearly the size of ImageNet, it might not be necessary to train for as long, due to all images being from the same visual domain. We also see that the small CNNs from Kumar \etal and Meijer \etal immediately diverge during training. This is possibly due to only applying two or three pooling layers before connecting to dense layers, leading to a parameter count of 269 and 135 million, respectively. Comparatively, the small CNN used by Xie \etal, uses three pooling layers as well as a pixel stride of two in the last two convolutional layers, leading to a parameter count of nine million parameters. Similarly, we observe that the ML-GCN method also diverges immediately, whereas the KSSNet method manages to train. We hypothesize that this is due to the lateral connections in the KSSNet adding stability during training. The loss curves are reported in the supplementary materials.
We observe that the methods from the multi-label classification domain are better at classifying the specific classes, with TResNet-L achieving a $\text{F}2_{\text{CIW}}$ test score of 54.75\%.  However, the simple two-stage approach by Xie \etal achieves the highest $\text{F}1_{\text{Normal}}$ score of 90.62\%. This indicates that the approach by Xie \etal excels at distinguishing whether there are \textit{any} classes, but not which one. The results are not solely due to the two-stage approach. Chen \etal also utilize a two-stage approach, but this produces significantly worse results. It is observed that the first stage simply predicts a ``defect'' in all images, which the later stage cannot properly handle. This is reflected by a low $\text{F}1_{\text{Normal}}$ score. Therefore, it appears there is value in using a small CNN for the first stage.

\begin{table}[!t]
\centering
\caption{\textbf{Performance metrics for each method.} We present the different metrics for each method. The metrics are presented as percentages, and the highest score in each column is denoted in bold. The Kumar \cite{Kumar2018EnsembleCNN}, Meijer \cite{Meijer2019MultiLabel} and ML-GCN \cite{MLGCN19} methods are not shown as they diverged during training. The ``Sewer'' and ``General'' identifiers indicate whether the method is from the sewer defect or multi-label classification domains, respectively. The classic multi-label metrics \cite{PartialLabel20} are reported in the supplementary materials.}
\label{tab:mainRes}
\resizebox{\linewidth}{!}{%
\begin{tabular}{lc|cc|cc}
\hline
\multicolumn{1}{c}{\multirow{2}{*}{}} & \multirow{2}{*}{\textbf{Model}} & \multicolumn{2}{c|}{\textbf{Validation}} & \multicolumn{2}{c}{\textbf{Test}} \\ \cline{3-6} 
\multicolumn{1}{c}{}                        &                        & \textbf{$\text{F}2_{\text{CIW}}$} $\uparrow$        & \textbf{$\text{F}1_{\text{Normal}}$} $\uparrow$        & \textbf{$\text{F}2_{\text{CIW}}$} $\uparrow$      & \textbf{$\text{F}1_{\text{Normal}}$} $\uparrow$  \\ \hline
\parbox[t]{2mm}{\multirow{4}{*}{\rotatebox[origin=c]{90}{Sewer}}}                                             & Xie \cite{Xie2019HierCNN}                   & 48.57        & \textbf{91.08}            & 48.34     & \textbf{90.62}         \\
                                             & Chen \cite{Chen2018}                   & 42.03        & 3.96             & 41.74     & 3.59         \\
                                             & Hassan \cite{Hassan2019AlexNet}                 & 13.14        & 0.00             & 12.94     & 0.00         \\
                                             & Myrans \cite{Myrans2018MultiClass}       &   4.01 & 26.03           & 4.11 & 27.48                 \\ \hline
\parbox[t]{2mm}{\multirow{5}{*}{\rotatebox[origin=c]{90}{General}}}                 & ResNet-101 \cite{ResNet}            & 53.26        & 79.55            & 53.21     & 78.57        \\

                                             & KSSNet \cite{KSSNet20}                 & 54.42        & 80.60            & 54.55     & 79.29        \\
                                             & TResNet-M \cite{TResNet20}             & 53.83        & 81.23            & 53.79     & 79.91        \\
                                             & TResNet-L \cite{TResNet20}              & \textbf{54.63}        & 81.22            & \textbf{54.75}     & 79.88        \\
                                             & TResNet-XL \cite{TResNet20}              & 54.42        & 81.81            & 54.24     & 80.42 \\ \hline    
\end{tabular}%
}
\end{table}

\subsection{Ablation Studies and Benchmark Algorithm}\label{sec:ablation}
Looking at the results in Table~\ref{tab:mainRes}, there is merit to both the end-to-end and two-stage approaches. We investigate whether the results can be improved further by combining end-to-end and two-stage methods. In the supplementary materials we report two additional ablation studies focused on getting a better understanding of the two-stage results.

\textbf{Effect of different second stage classifiers.} Based on our results in Table~\ref{tab:mainRes}, we look into whether combining the general multi-label methods with two-stage approaches would lead to state-of-the-art performance. Specifically, we combine the first stage of Xie \etal with each of the multi-label classifiers in Table~\ref{tab:mainRes}. The results are shown in Table~\ref{tab:combRes}. 
We observe that by utilizing the first stage of Xie \etal both the $\text{F}2_{\text{CIW}}$ and $\text{F}1_{\text{Normal}}$ scores are improved when compared to the best results in Table~\ref{tab:mainRes}. Moreover, the performance is improved for all tested methods. Specifically, by using the first stage to filter out normal pipes, all general multi-label methods increase their $\text{F}2_{\text{CIW}}$ scores by approximately 0.5-1 percentage points, and the $\text{F}1_{\text{Normal}}$ by up to 10-12 percentage points. For the sewer domain methods their $\text{F}2_{\text{CIW}}$ scores are increased by 5-13 percentage points, and the $\text{F}1_{\text{Normal}}$ by 65-90 percentage points. From these results we can conclude that using a two-stage approach with the binary classifier from Xie \etal \cite{Xie2019HierCNN} and the TResNet-L model \cite{TResNet20} is the \textit{Benchmark} algorithm on Sewer-ML, with an $\text{F}2_{\text{CIW}}$ score of $55.11\%$ and $\text{F}1_{\text{Normal}}$ score of $90.94\%$.

\begin{table}[!t]
\centering
\caption{\textbf{Two-stage classifier permutations.} We evaluate each of the tested multi-label classifiers in a two-stage setup together with the first stage used by Xie \etal \cite{Xie2019HierCNN}.}
\label{tab:combRes}
\resizebox{\linewidth}{!}{%
\begin{tabular}{lc|cc|cc}
\hline
\multicolumn{1}{c}{\multirow{2}{*}{}} &\multirow{2}{*}{\begin{tabular}[c]{@{}c@{}}\textbf{Second}\\\textbf{Stage}\end{tabular}} & \multicolumn{2}{c|}{\textbf{Validation}} & \multicolumn{2}{c}{\textbf{Test}} \\ \cline{3-6} 
\multicolumn{1}{c}{}         &                                          \multicolumn{1}{c|}{}               & \textbf{$\text{F}2_{\text{CIW}}$} $\uparrow$        & \textbf{$\text{F}1_{\text{Normal}}$} $\uparrow$       & \textbf{$\text{F}2_{\text{CIW}}$} $\uparrow$     & \textbf{$\text{F}1_{\text{Normal}}$} $\uparrow$   \\ \hline
\parbox[t]{2mm}{\multirow{3}{*}{\rotatebox[origin=c]{90}{Sewer}}}  &Chen \cite{Chen2018}            & 48.67 & 91.06            &  48.19 & 90.60           \\
&Hassan \cite{Hassan2019AlexNet}           & 18.08 & 91.08            &  17.89 & 90.62           \\
&Myrans \cite{Myrans2018MultiClass}     & 27.87 & 91.08  & 27.83 & 90.62          \\ \hline
\parbox[t]{2mm}{\multirow{5}{*}{\rotatebox[origin=c]{90}{General}}}                                            &ResNet-101 \cite{ResNet}            & 54.45 & 91.28            &  54.01 & 90.88          \\
&KSSNet \cite{KSSNet20}                & \textbf{55.37} & 91.30            & 55.09 & \textbf{90.95}        \\
&TResNet-M \cite{TResNet20}              & 54.58 & 91.33             & 54.32 & 90.93         \\
&TResNet-L \cite{TResNet20}              & 55.36 & 91.32             & \textbf{55.11} & 90.94        \\
&TResNet-XL \cite{TResNet20}             & 54.97 & \textbf{91.37}            & 54.51 & \textbf{90.95}  \\ \hline    
\end{tabular}%
}
\end{table}

\subsection{Per-Class Performance}
\begin{figure}
    \centering
    \includegraphics[width=\linewidth]{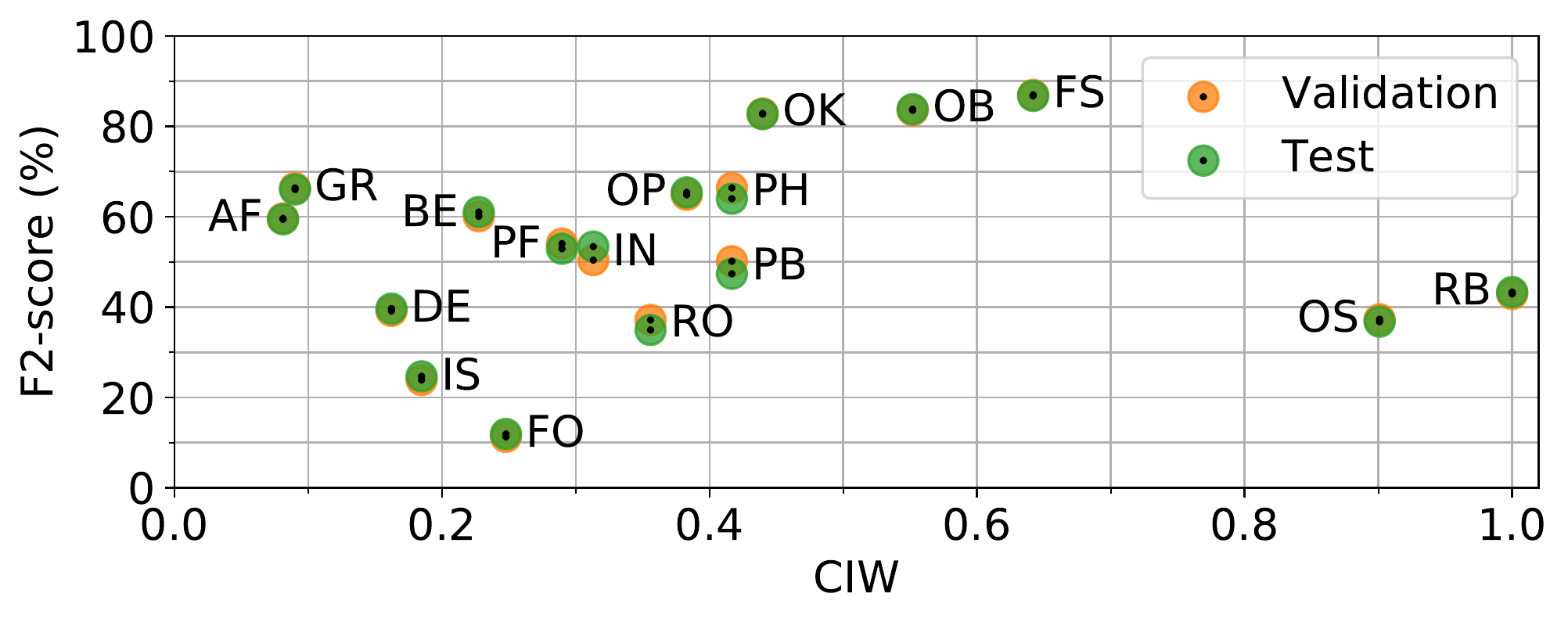}
    \caption{\textbf{Per-class performance.} Per-class F2 scores of the \textit{Benchmark} algorithm  (TResNet-L + Xie \etal), plotted against the corresponding CIW values.}
    \label{fig:CIWScatter}
\end{figure}

To gain a better understanding of the difficulty of detecting the different defects compared to their economical impact, we compare the F2 score for each defect with the corresponding CIW scores, see Figure~\ref{fig:CIWScatter}. We find that each of the classes with a high F2 score exhibit low intra-class and high inter-class variance, as well as more frequently occurring in the dataset. The displaced joint class FS exhibits limited intra-class variance due to limitations in where the defect can occur within the pipe, while being distinct from the other classes. Similarly, the surface damage class OB occurs frequently in the dataset and exhibits high inter-class variance due to the distinct visual appearance of the class.

Contrarily, the lower scoring defect classes exhibit a larger intra-class variance, lower inter-class variance, and are less frequently occurring. The obstacle class FO consists of a wide span of objects, \eg a soda can, a leftover hammer, or another pipe which goes through the main pipe. The RB class exhibits large intra-class variance, due to the class encompassing cracks, breaks, and collapses, and low inter-class variance, due to the similarity in appearance between \eg cracks and the fine roots in the RO class.

We observe that most of the lower scoring defects do not have a large economic impact. However, the two defects with the highest economic impact, OS and RB, are among the lowest scoring classes. Therefore, in order to improve the performance of the classification system, the detection rate on these two classes should be the main priority.  

\label{sec:per-class}

\section{Conclusion}\label{sec:conclusion}
Sewerage infrastructure is a fundamental part of modern society and is continuously expanded. However, current manual inspections are tedious and slow when compared to the immense number of pipes that have to be inspected. Therefore, automated sewer inspection technologies are crucial to ensuring the quality of our sewerage infrastructure. However, current state-of-the-art sewer defect classification methods have not yet adopted recent advances within computer vision. In order to facilitate this transition, we present the first public, multi-label sewer defect classification dataset called Sewer-ML.

Sewer-ML consists of 1.3 million images of a large variety of sewer pipes annotated by professional sewer inspectors. The data is acquired from 75,618 inspection videos conducted over nine years. 12 methods from the sewer defect classification and multi-label classification domains are compared on Sewer-ML. 
Methods are evaluated using a novel, class-importance weighted F2 score, $\text{F}2_{\text{CIW}}$, which incorporates the economic impact of each class, and the F1 score for pipes with no annotated classes, $\text{F}1_\text{Normal}$. We present a benchmark algorithm by combining the best two-stage approach from the sewer domain with the best classifier from the multi-label domain, achieving a state-of-the-art performance with an $\text{F}2_{\text{CIW}}$ of 55.11\% and $\text{F}1_\text{Normal}$ of 90.94\%. The code, data, and trained models are open-sourced in order to lower the barrier of entry and encourage further development within sewer defect classification.\\
\noindent\textbf{Acknowledgments.} This research was funded by Innovation Fund Denmark [grant number 8055-00015A] and is part of the Automated Sewer Inspection Robot (ASIR) project.

\appendix
\section{Supplementary Materials Content}
In these supplementary materials we describe in further detail aspects of the dataset and the training process and performance of the tested methods. Specifically, the following will be described:
\begin{itemize}
    \itemsep0em
    \item Additional examples from the Sewer-ML dataset (Section~\ref{sec:dataExamples}).
    \item Further insights into the Sewer-ML dataset (Section~\ref{sec:dataInsights}).
    \item Full training details and metric performance for the Faster-RCNN text detector  (Section~\ref{sec:textDetector}).
    \item Full details on the Extra Trees hyperparameter grid search (Section~\ref{sec:ETSearch}).
    \item The loss curves of the trained multi-label classification methods (Section \ref{sec:lossCurves}).
    \item Ablation study of the two-stage methods (Section~\ref{sec:ablationSup}).
    \item Results when evaluating using the common multi-label performance metrics (Section~\ref{sec:metricsSup}).
\end{itemize}

\section{Sewer-ML Dataset Examples}\label{sec:dataExamples}
\begin{figure*}[!t]
\centering
\begin{subfigure}[b]{0.3\linewidth}
  \centering
  \includegraphics[width=0.98\linewidth]{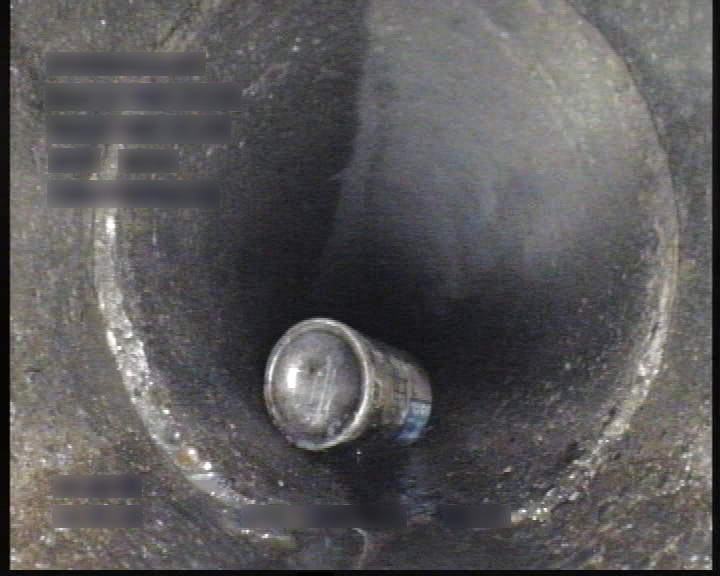}
  \caption{FO, FS}
\end{subfigure}%
\begin{subfigure}[b]{0.3\linewidth}
  \centering
  \includegraphics[width=0.98\linewidth]{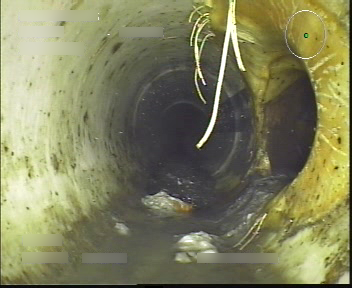}
  \caption{DE, AF, FO, OP}
\end{subfigure}%
\begin{subfigure}[b]{0.3\linewidth}
  \centering
  \includegraphics[width=0.98\linewidth]{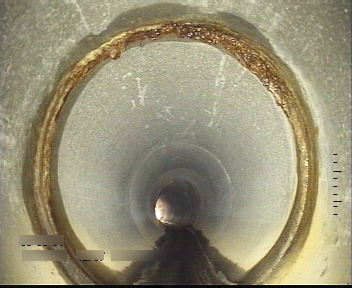}
  \caption{IN, OB, BE}
\end{subfigure}

\begin{subfigure}[b]{0.3\linewidth}
  \centering
  \includegraphics[width=0.98\linewidth]{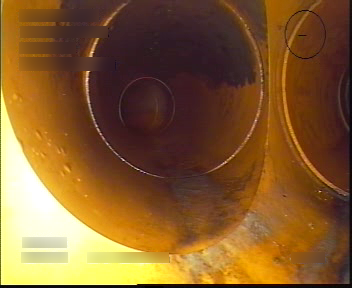}
  \caption{GR, OB, FS, OK}
\end{subfigure}%
\begin{subfigure}[b]{0.3\linewidth}
  \centering
  \includegraphics[width=0.98\linewidth]{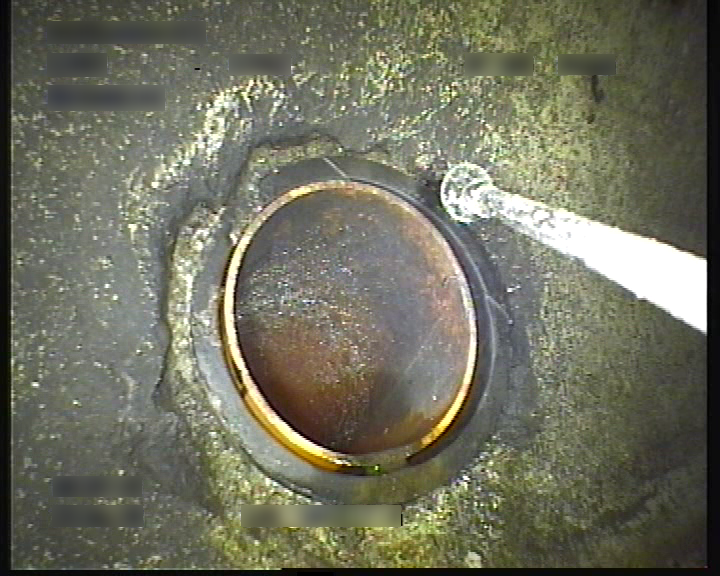}
  \caption{PB, RB, OB, FS}
\end{subfigure}%
\begin{subfigure}[b]{0.3\linewidth}
  \centering
  \includegraphics[width=0.98\linewidth]{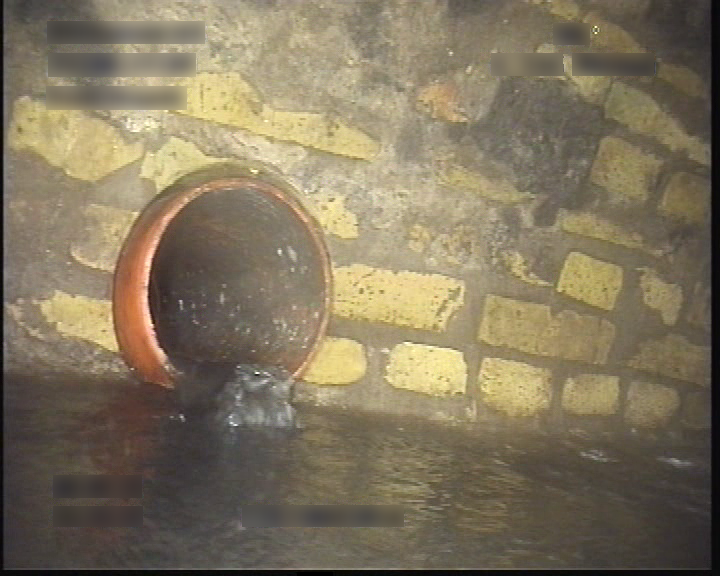}
  \caption{PH, RO, OB}
\end{subfigure}

\begin{subfigure}[b]{0.3\linewidth}
  \centering
  \includegraphics[width=0.98\linewidth]{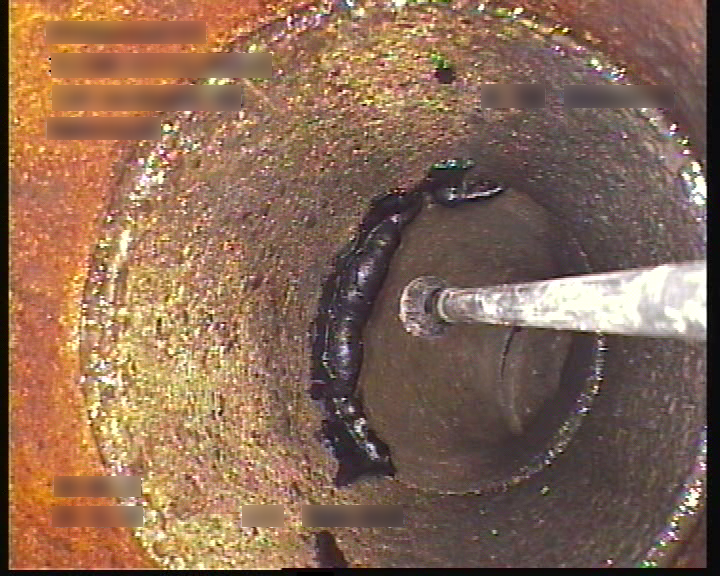}
  \caption{IS, OK}
\end{subfigure}%
\begin{subfigure}[b]{0.3\linewidth}
  \centering
  \includegraphics[width=0.98\linewidth]{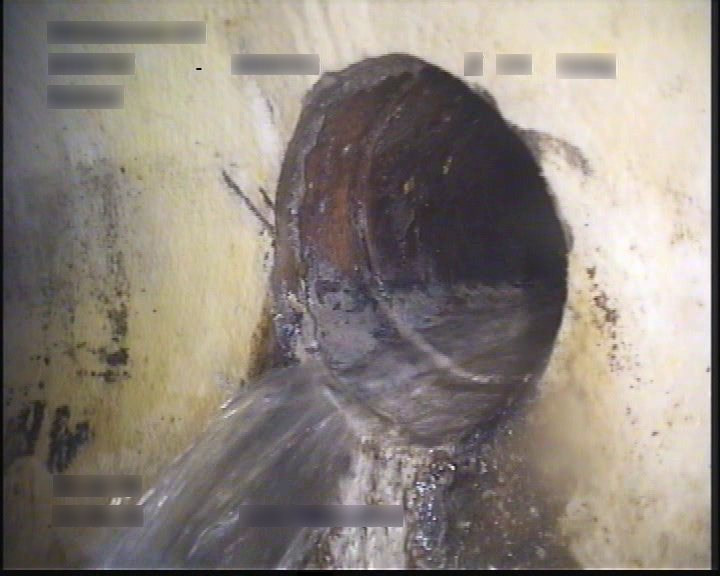}
  \caption{OS, PF}
\end{subfigure}%
\begin{subfigure}[b]{0.3\linewidth}
  \centering
  \includegraphics[width=0.98\linewidth]{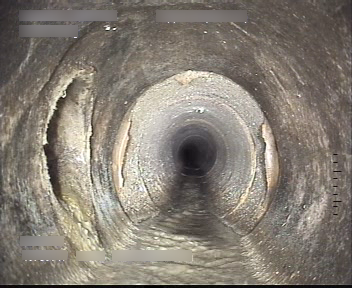}
  \caption{OP, OK, FS, OB}
\end{subfigure}%
\caption{\textbf{Sewer-ML data examples with co-occurring classes.} A subset of the images in the Sewer-ML showcasing images with multiple classes co-occurring and all annotated classes represented. The class codes are described in Table~\ref{tab:ClassOverview}.}
\label{fig:MultiExamples}
\end{figure*}

In this section we present more examples of the images in the Sewer-ML dataset. All images are annotated using the Danish inspection standard containing 18 classes \cite{Standard2010}, listed in Table~\ref{tab:ClassOverviewSup}. 
In Figure~\ref{fig:MultiExamples} we present examples of different cases with several co-occurring classes.
In Figure~\ref{tab:perClassExample} we present five examples of each class, where only the mentioned class is present.

\begin{table}[!t]
\centering
\caption{\textbf{Sewer inspection classes.} Overview and short description of each annotation class \cite{Standard2010}.}
\label{tab:ClassOverviewSup}
\begin{tabular}{l|l}\hline
\textbf{Code} & \textbf{Description}                         \\ \hline
VA     & Water Level (in percentages)     \\
RB     & Cracks, breaks, and collapses      \\
OB     & Surface damage                \\
PF     & Production error                    \\
DE     & Deformation                   \\
FS     & Displaced joint                 \\
IS     & Intruding sealing material     \\
RO     & Roots                        \\
IN     & Infiltration                      \\
AF     & Settled deposits                   \\
BE     & Attached deposits               \\
FO     & Obstacle                          \\
GR     & Branch pipe                      \\
PH     & Chiseled connection        \\
PB     & Drilled connection                 \\
OS     & Lateral reinstatement cuts     \\
OP     & Connection with transition profile    \\
OK     & Connection with construction changes \\ \hline
\end{tabular}
\end{table}

\begin{table*}[!htb]
\centering
\caption{\textbf{Class occurrences per split.} The number of occurrences for each class per dataset split.}
\label{tab:classCount2}
\resizebox{\textwidth}{!}{%
\begin{tabular}{l|rrrrrrrrrrrrrrrrrr}
\textbf{Split}      & \textbf{RB}    & \textbf{OB}     & \textbf{PF}    & \textbf{DE}    & \textbf{FS}     & \textbf{IS}   & \textbf{RO}    & \textbf{IN}    & \textbf{AF}    & \textbf{BE}    & \textbf{FO}   & \textbf{GR}    & \textbf{PH}    & \textbf{PB}   & \textbf{OS}   & \textbf{OP}   & \textbf{OK}     & \textbf{Normal} \\ \hline
Training & 45,821 & 184,379 & 16,254 & 19,084 & 283,983 & 6,271 & 22,637 & 23,782 & 74,856 & 66,499 & 5,010 & 53,986 & 23,685 & 6,746 & 4,625 & 5,325 & 154,624 & 552,820 \\
Validation   & 5,538  & 23,624  & 2,021  & 2,038  & 36,218  & 881  & 2,917 & 2,812  & 9,059  & 7,929  & 597  & 6,889  & 3,432  & 765  & 457  & 612  & 19,655  & 68,681  \\
Test  & 5,501  & 23,264  & 1,949  & 2,307  & 35,781  & 924  & 2,684  & 3,235  & 9,182  & 8,720  & 649  & 6,726  & 2,962  & 833  & 530  & 533  & 19,420  & 69,221 \\ \hline
Total & 56,860 & 231,267 & 20,224 & 23,429 & 355,982 & 8,076 & 28,238 & 29,829 & 93,097 & 83,148 & 6,256 & 67,601 & 30,079 & 8,344 & 5,612 & 6,470 & 193,699 & 690,722
\end{tabular}%
}
\end{table*}
\section{Sewer-ML Dataset Insights}\label{sec:dataInsights}

In this section, we describe the available information in the Sewer-ML dataset in more detail.
First, we report the number of occurrences for each class in the dataset splits, see Table~\ref{tab:classCount2}, where it is observed that the distribution of the classes is similar across the different splits.

Moreover, we look into the pipe properties associated with each image. Each image contains information on the pipe shape, material, dimension, and water level. 

\begin{figure}[!t]
    \centering
    \includegraphics[width=\linewidth]{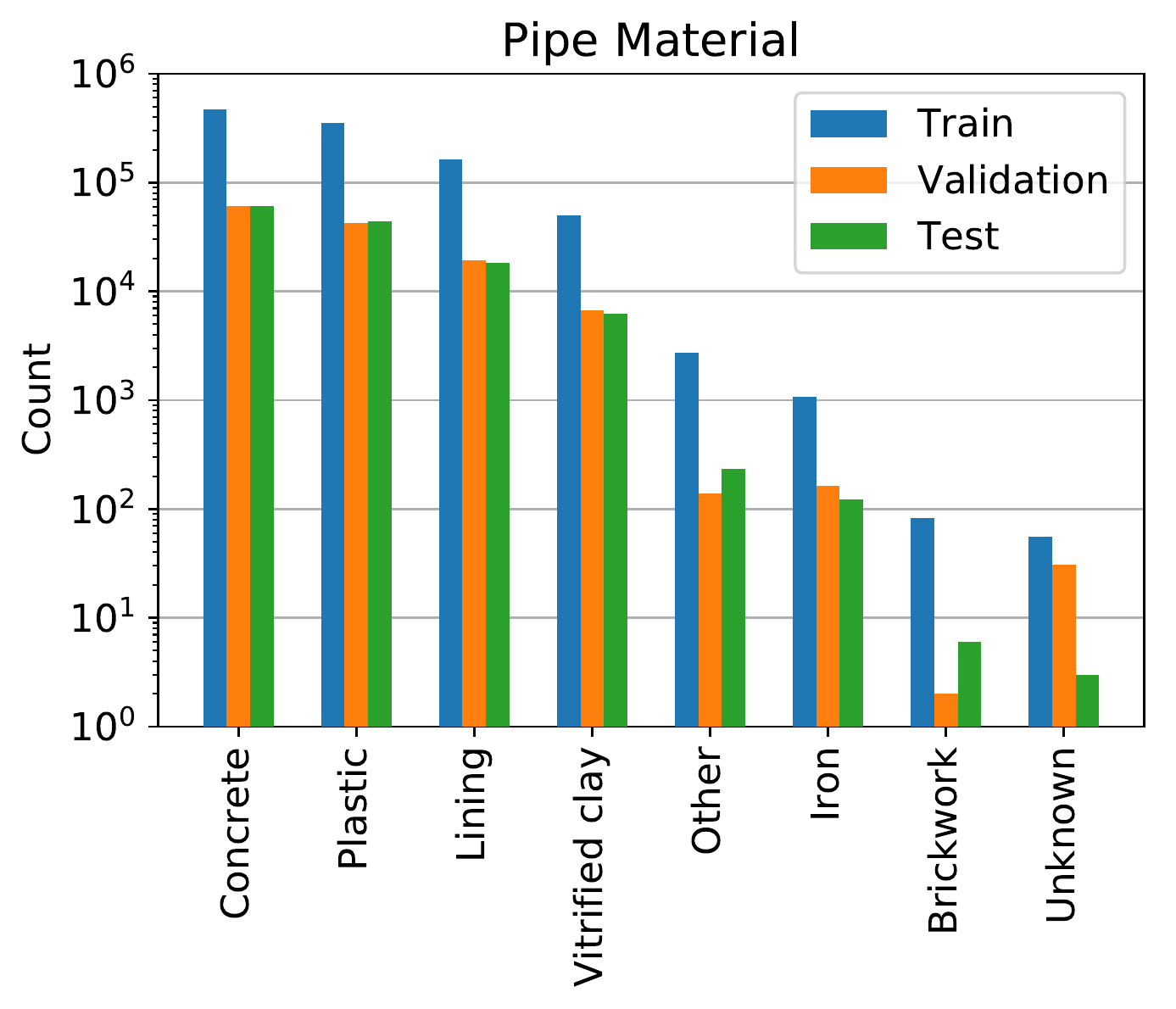}
    \caption{\textbf{Distribution of the pipe materials}. We plot the occurrence frequencies for each of the eight pipe materials in the dataset, for each dataset split. Note that the y-axis is log-scaled.}
    \label{fig:pipeMaterial}
\end{figure}

In Figure~\ref{fig:pipeMaterial} we plot the distribution of the eight different pipe material types for the images in each split. We find that the concrete, vitrified clay, plastic, and lining materials are the most common materials in the Sewer-ML dataset. We also observe that all material types are equally represented across the splits, except for the ``Brickwork'' and ``Unknown'' material types. The reason these material types are skewed for the validation and test sets, is due to these materials being rarely used anymore, and therefore rarely occur in the sewer inspection videos. Therefore, the images containing these material types are from a small subset of pipes, which were not evenly spread out across the splits.

\begin{figure}[!t]
    \centering
    \includegraphics[width=\linewidth]{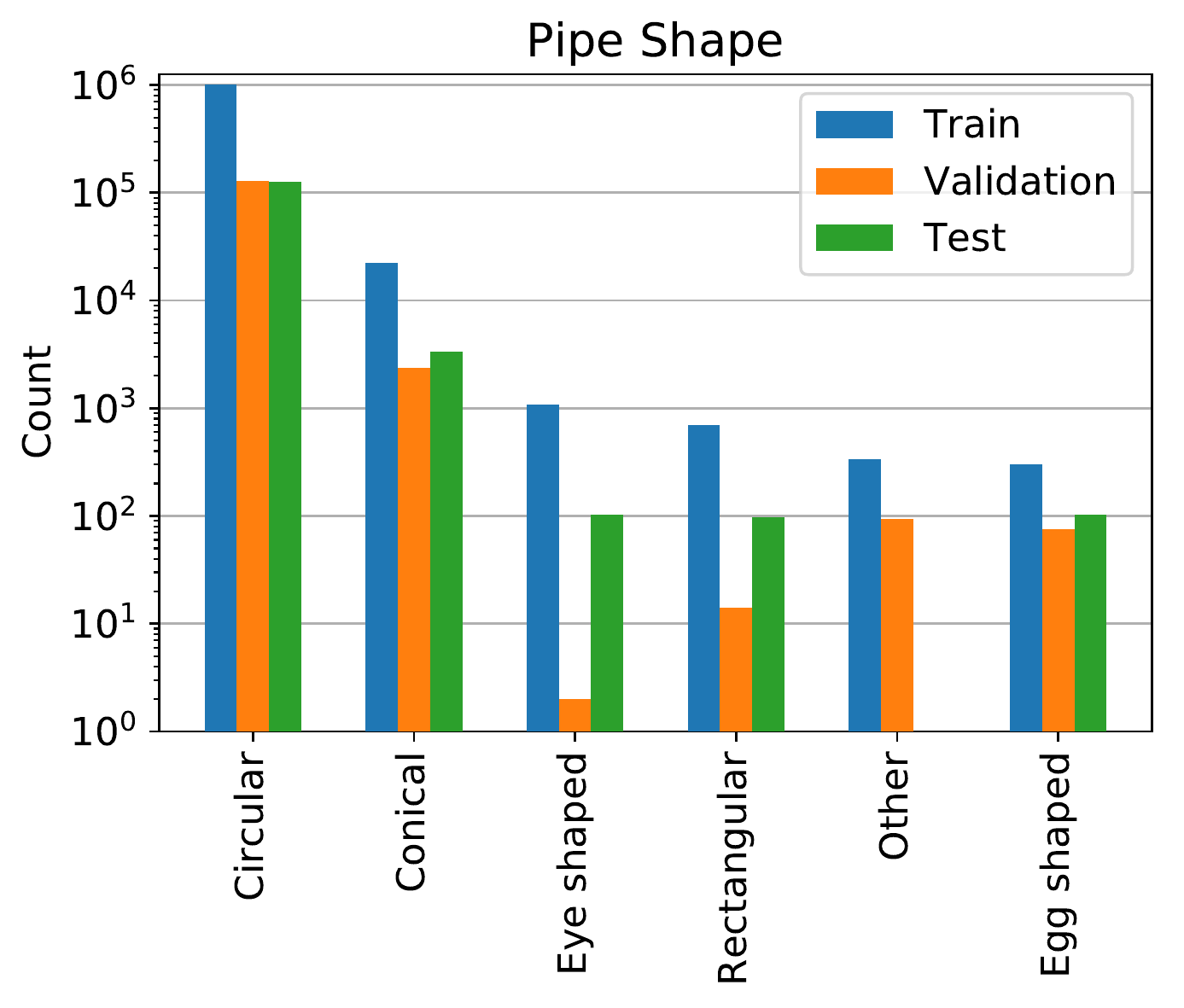}
    \caption{\textbf{Distribution of the pipe shapes.} We plot the occurrence frequencies for each of the six pipe shapes in the dataset, for each dataset split. Note that the y-axis is log-scaled.}
    \label{fig:pipeShape}
\end{figure}

In Figure~\ref{fig:pipeShape} we plot the distribution of the six different pipe shapes for the images in each of the dataset splits. We find that the circular type is by far the most common pipe shape, followed secondly by conical pipes, whereas the remaining pipe shapes only appear a few thousand times each. As with the pipe material, we see that distribution of pipe shapes are similar between dataset splits, except for the ``Eye shaped'', ``Rectangular'', and ``Other'' pipe shapes. This is again due to these pipe shapes occurring in a limited set of sewer inspections, and have therefore not been evenly divided across the splits.

\begin{figure}[!t]
    \centering
    \includegraphics[width=\linewidth]{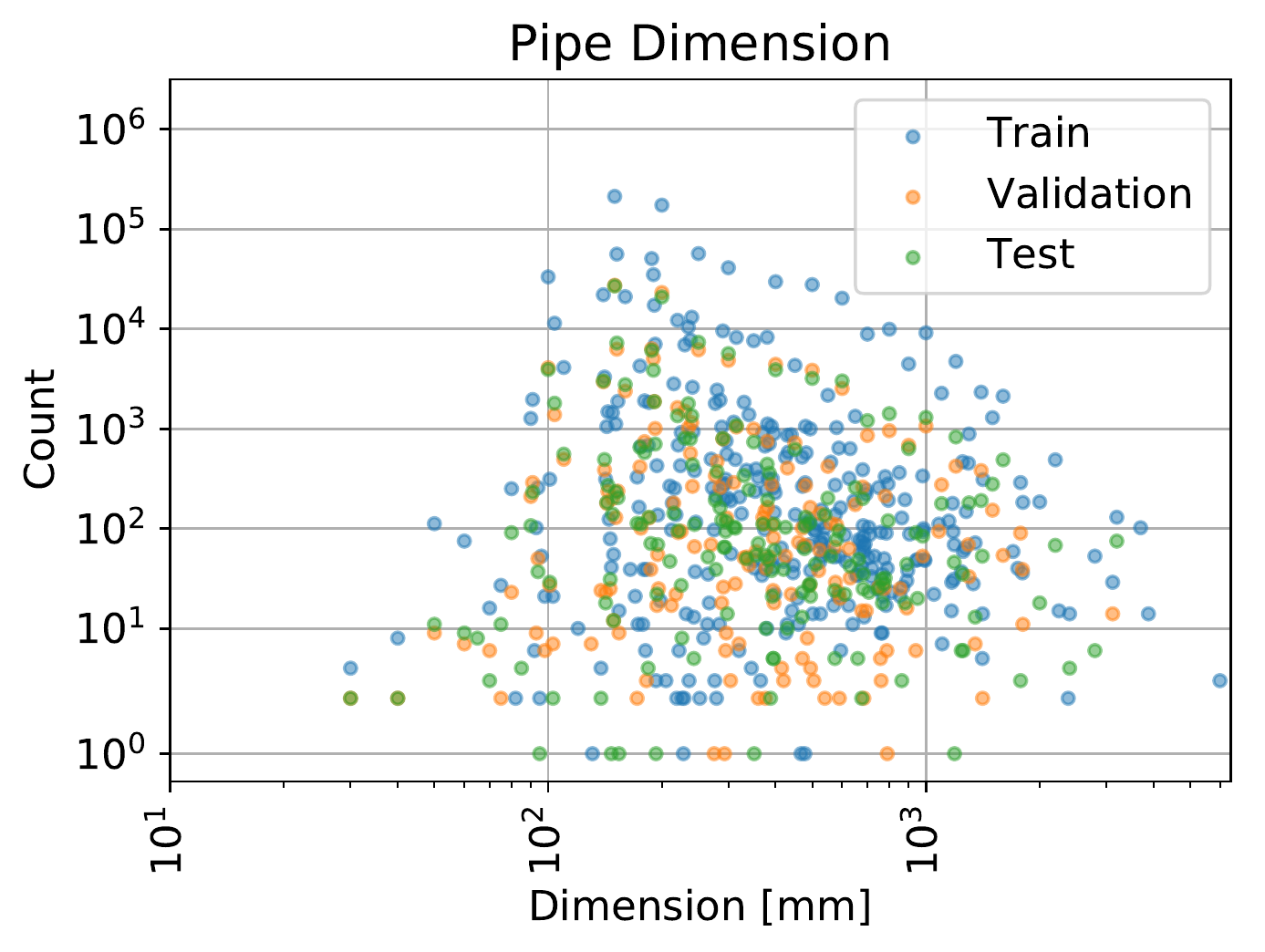}
    \caption{\textbf{Distribution of the pipe dimensions.} Plots of the occurrence frequencies of each pipe dimension, for each dataset split. Note that both axes are log-scaled.}
    \label{fig:pipeDimension}
\end{figure}

\vspace{0.5cm}
In Figure~\ref{fig:pipeDimension} we plot the occurrences of the pipe dimensions associated with each image. The dimension is denoted in millimeters, as per the industry standard. We see that the majority of images are from pipes with a diameter of 100--1,000 millimeters, with a skew towards 100 millimeters. We observe that the distribution of the pipe dimension for the training, validation, and test splits appears to be similar in shape, as expected.

\begin{figure}
    \centering
    \includegraphics[width=\linewidth]{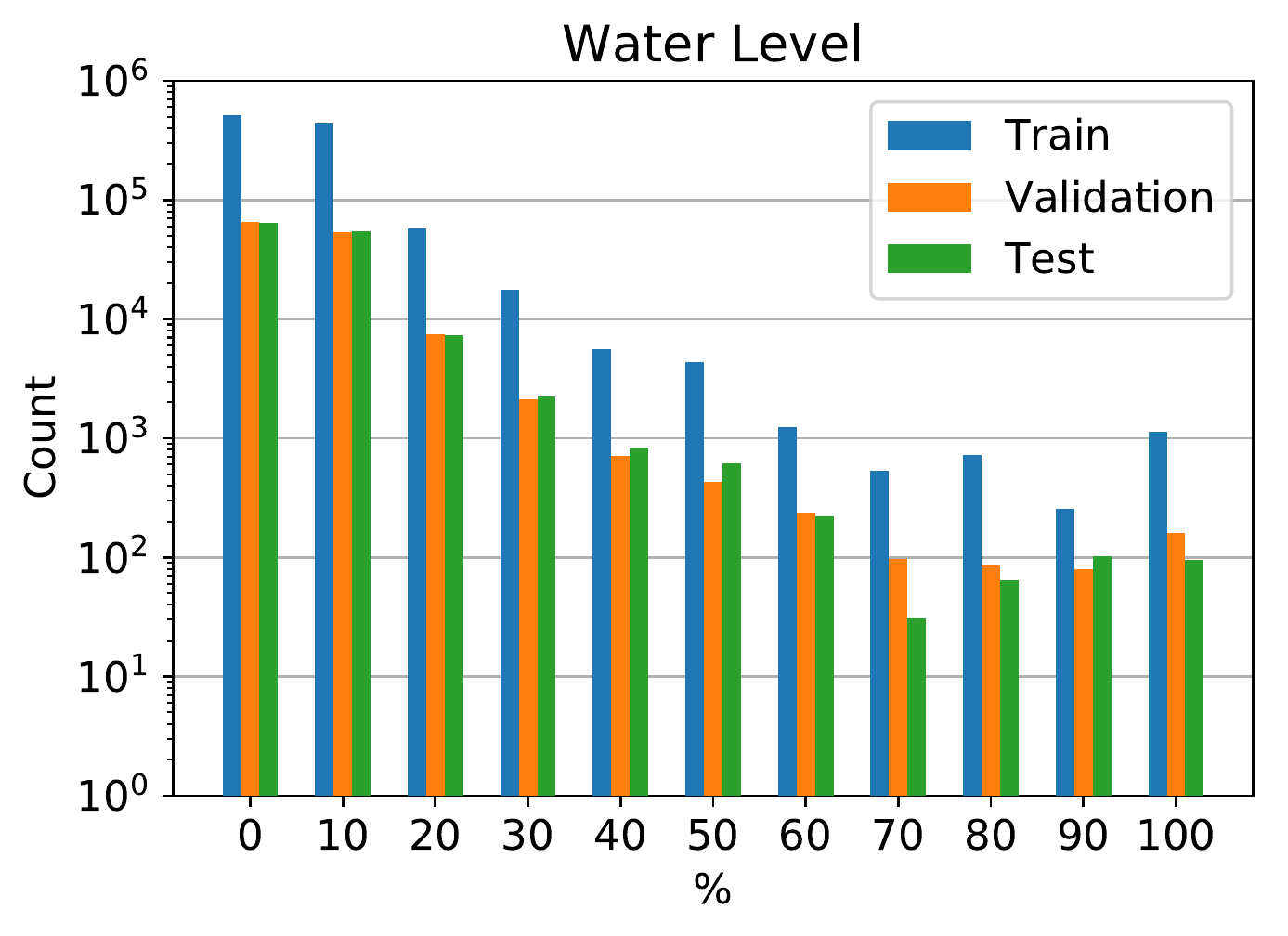}
    \caption{\textbf{Distribution of the water level.} We plot the occurrence frequencies for each of the water level classes, for each dataset split. Note that the y-axis is log-scaled.}
    \label{fig:waterLevel}
\end{figure}

In Figure~\ref{fig:waterLevel} we plot the distribution of the different water level classes for each data split. We find that the distribution of the water level classes is similar across the three dataset splits We also observe that the majority of the images have an associated water level in the range 0--30 \%, while the remaining classes occur less often, and not as evenly split between the classes. This can be explained by the fact that when the majority of a pipe is filled with water, the inspections may at times be postponed for a later time and it becomes difficult to accurately access how much water it actually contains. Furthermore, the inspection vehicle will at times be partially or fully submerged in the water, resulting in the inspector losing key reference points used for estimating the water level, such as the pipe wall.

\begin{figure}[!t]
    \centering
    \includegraphics[width=\linewidth]{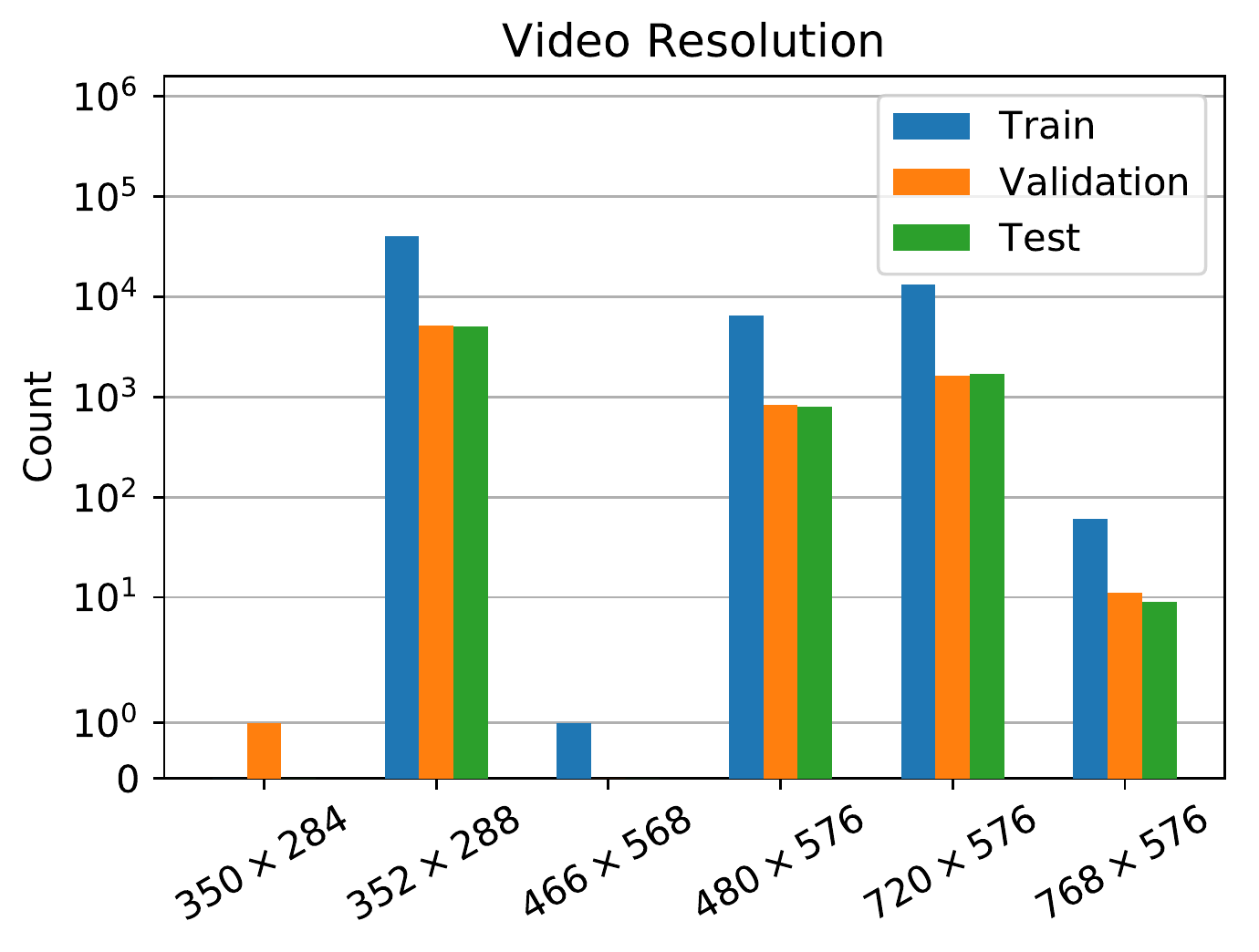}
    \caption{\textbf{Distribution of video resolution.} We present the distribution of the different resolutions for the videos in each dataset split. Note the y-axis is log-scaled.}
    \label{fig:videoResolution}
\end{figure}

\begin{figure}[!t]
    \centering
    \includegraphics[width=\linewidth]{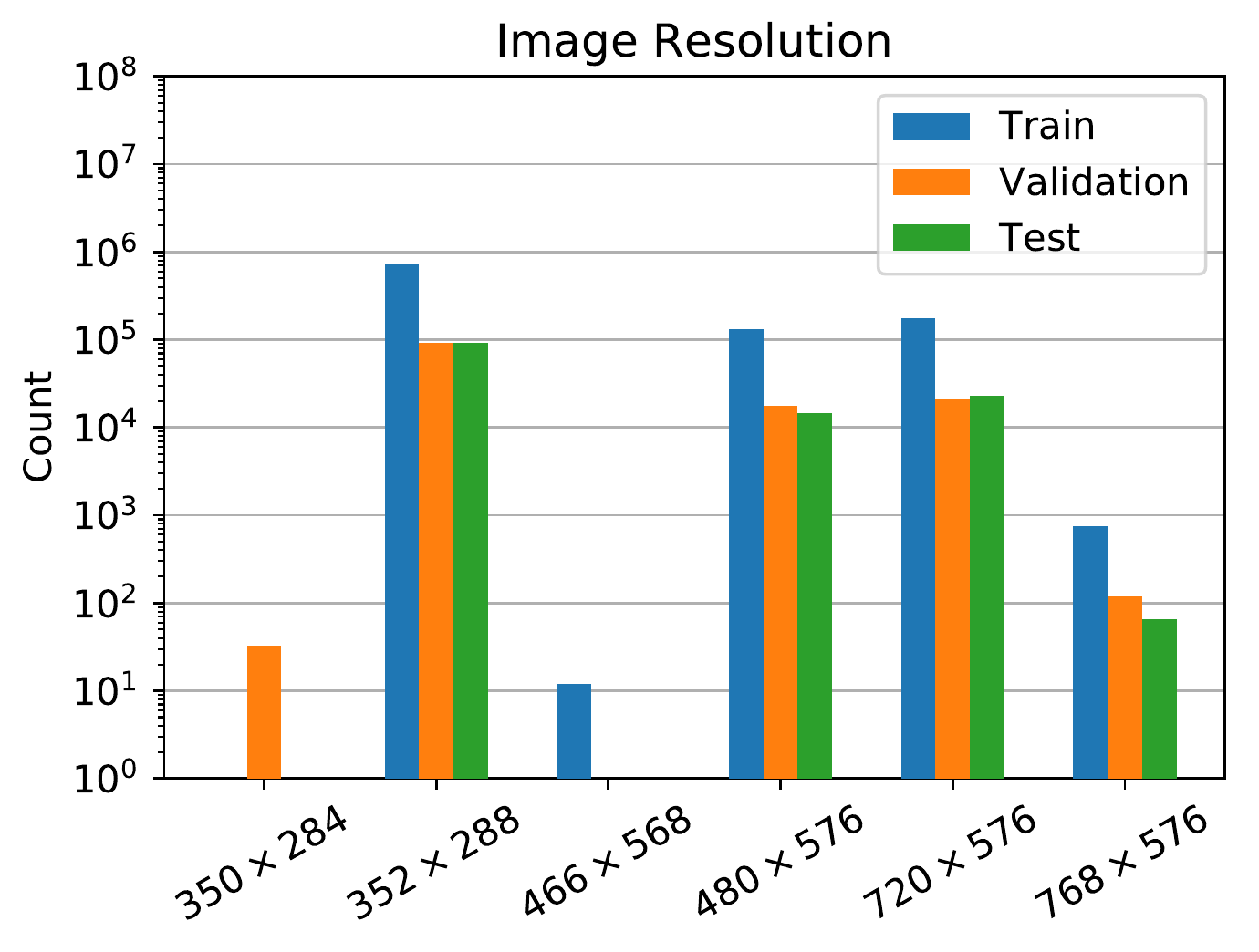}
    \caption{\textbf{Distribution of image resolution.} We present the distribution of the different resolutions for the images in each dataset split. Note the y-axis is log-scaled.}
    \label{fig:imageResolution}
\end{figure}

Lastly, in Figure~\ref{fig:videoResolution} we plot the resolution of the sewer inspection videos in each split. The resolution is denoted as width by height. It should be noted that the video resolutions reported are not the resolutions observed by the inspector. The videos are encoded in such a way that the video data is stored in the resolution reported in this work, but when presented using a media player the width is multiplied by a ``sample aspect ratio''. We decide not to apply this resizing, in order to not introduce artifacts in the image data. We find that across the videos in each dataset split, the resolutions are evenly distributed. This is also true when looking at the resolution for all the images in the dataset splits, see Figure~\ref{fig:imageResolution}.

\section{Faster-RCNN Training and Metric Details}\label{sec:textDetector}
In this section we detail the hyperparameters and training settings for the Faster-RCNN \cite{FRCNN} model we use to redact overlaid text information on the images. We also present the full COCO \cite{COCO14} metric suite performance, to show how well the network performs.  A training split of 20,739 images and a validation split of 2,305 images are used, wherein all text information is manually annotated with bounding boxes.

\begin{figure}
    \centering
    \includegraphics[width=0.95\linewidth]{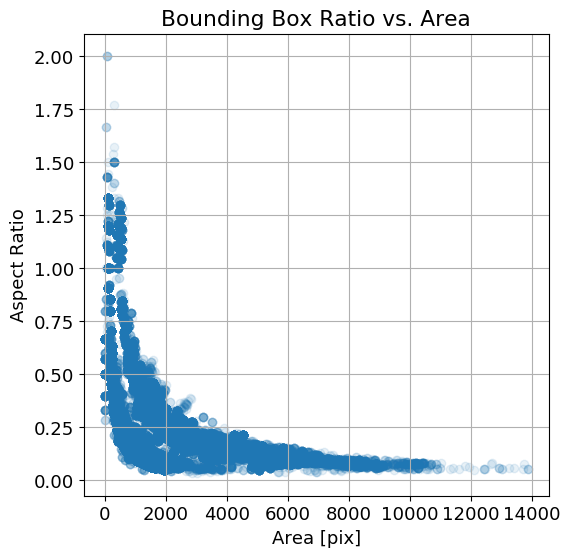}
    \caption{\textbf{Training split bounding box information.} The training split bounding box annotations are plotted with the bounding box area against the bounding box ratio.}
    \label{fig:bbInfo}
\end{figure}
\begin{figure}[!t]
    \centering
    \includegraphics[width=0.7\linewidth]{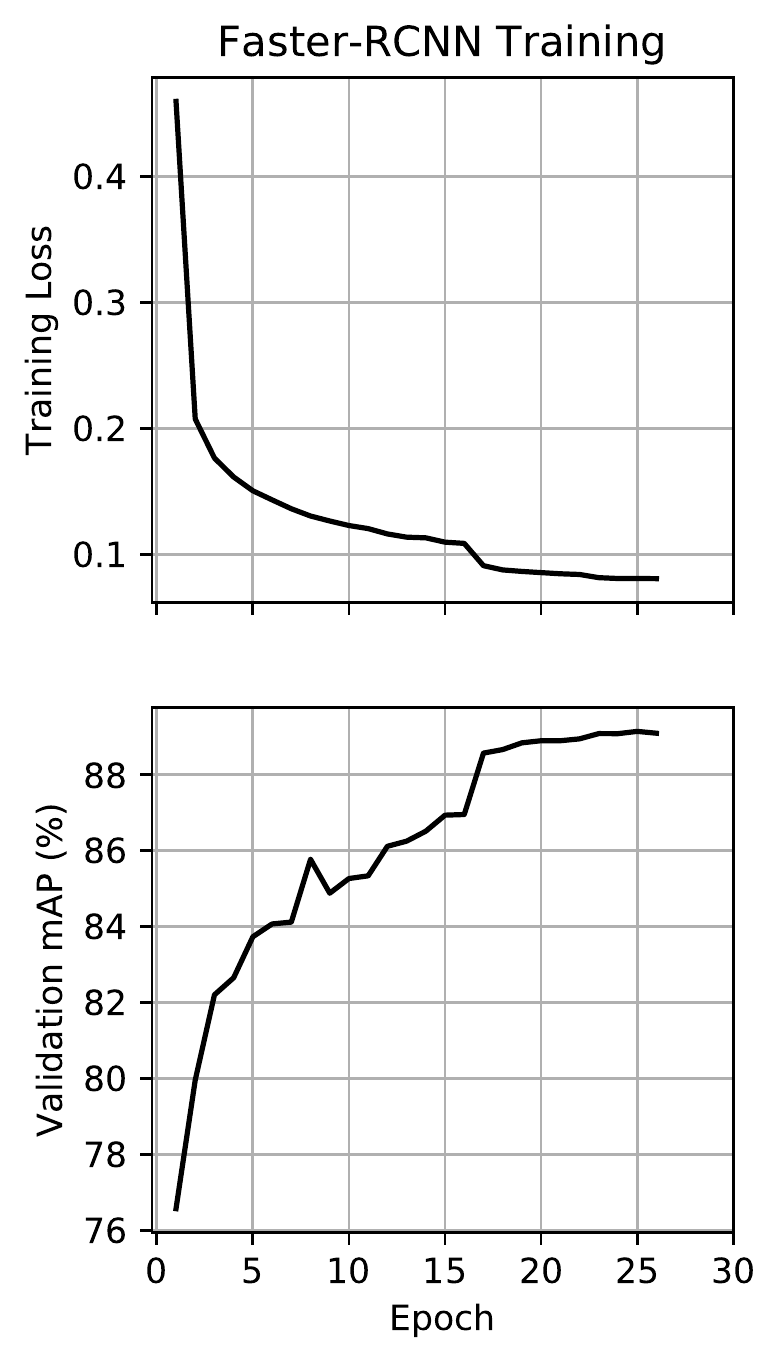}
    \caption{\textbf{Faster-RCNN loss and metric curves.} The training loss and validation metrics for the trained Faster-RCNN model. mAP@[0.5:0.95] is denoted as mAP.}
    \label{fig:frcnnLoss}
\end{figure}

\textbf{Hyperarameters.} The Faster-RCNN model is trained for 26 epochs with a batch size of 16 batches. An SGD optimizer with momentum is used, with a learning rate of 0.02, momentum of 0.9 and weight decay of 0.0001. The learning rate is multiplied by 0.1 at epoch 16 and 22, respectively. We employ linear warm up of the learning rate during the first 1,000 mini batches of the first epoch, increasing the learning rate from $10^{-3}$ to 0.02. The backbone is a ResNet-50 FPN \cite{ResNet,FPN} pre-trained on ImageNet \cite{ILSVRC15}, of which we fine-tune the last three residual blocks. Custom anchor boxes are used, with a bounding box ratios (height over width) of 1:8, 1:4 and 1:2, and bounding box scales with areas of $32^2$, $64^2$, $128^2$, $256^2$, and $512^2$. These values are determined based on the bounding box information in the training split, see Figure~\ref{fig:bbInfo}. All images are normalized using the ImageNet per channel mean and standard deviation, and horizontal flipping with a 50\% chance is used during training. The images are rescaled such that the shortest side is 800 pixels, while enforcing that the largest side is no larger than 1,333 pixels. The training loss and mAP[0.5:0.95] on the validation set are plotted in Figure~\ref{fig:frcnnLoss}.

\begin{table*}[!t]
\centering
\caption{\textbf{Full COCO metric suite.} The performance of the trained Faster-RCNN model on the validation set, for different Average Precision (AP) and Average Recall (AR) settings.} 
\label{tab:cocoMetrics}
\begin{tabular}{ccc|ccc|ccc|ccc} \hline
\multicolumn{3}{c|}{\textbf{AP, IoU:}} & \multicolumn{3}{c|}{\textbf{AP@[0.5:0.95], Area:}} & \multicolumn{3}{c|}{\textbf{AR@[0.5:0.95], \#Dets:}} & \multicolumn{3}{c}{\textbf{AR@[0.5:0.95], Area:}} \\
0.5:0.95 & 0.5 & 0.75 & S & M & L & 1 & 10 & 100 & S & M & L \\ \hline
     89.10       & 98.89        & 96.39        & 88.08  &   89.96    & 95.63         & 10.06          & 88.31         & 92.25          & 91.72          & 92.71         & 96.28          \\ \hline
\end{tabular}
\end{table*}

\textbf{Metrics.} In order to determine the effect of the Faster-RCNN model, we compute the full COCO metrics suite on the validation set, as shown in Table~\ref{tab:cocoMetrics}. As shown in the metrics, we have a high precision and recall, though the recall indicates that not all of the text objects have been detected. This is partially due to some text information being annotated with a single bounding box but detected as several boxes, and vice versa. To verify the annotations we manually inspect a set of randomly selected samples.

\section{Extra Trees Hyperparameter Grid Search}\label{sec:ETSearch}
\begin{figure*}[!t]
    \centering
    \includegraphics[width=0.95\linewidth]{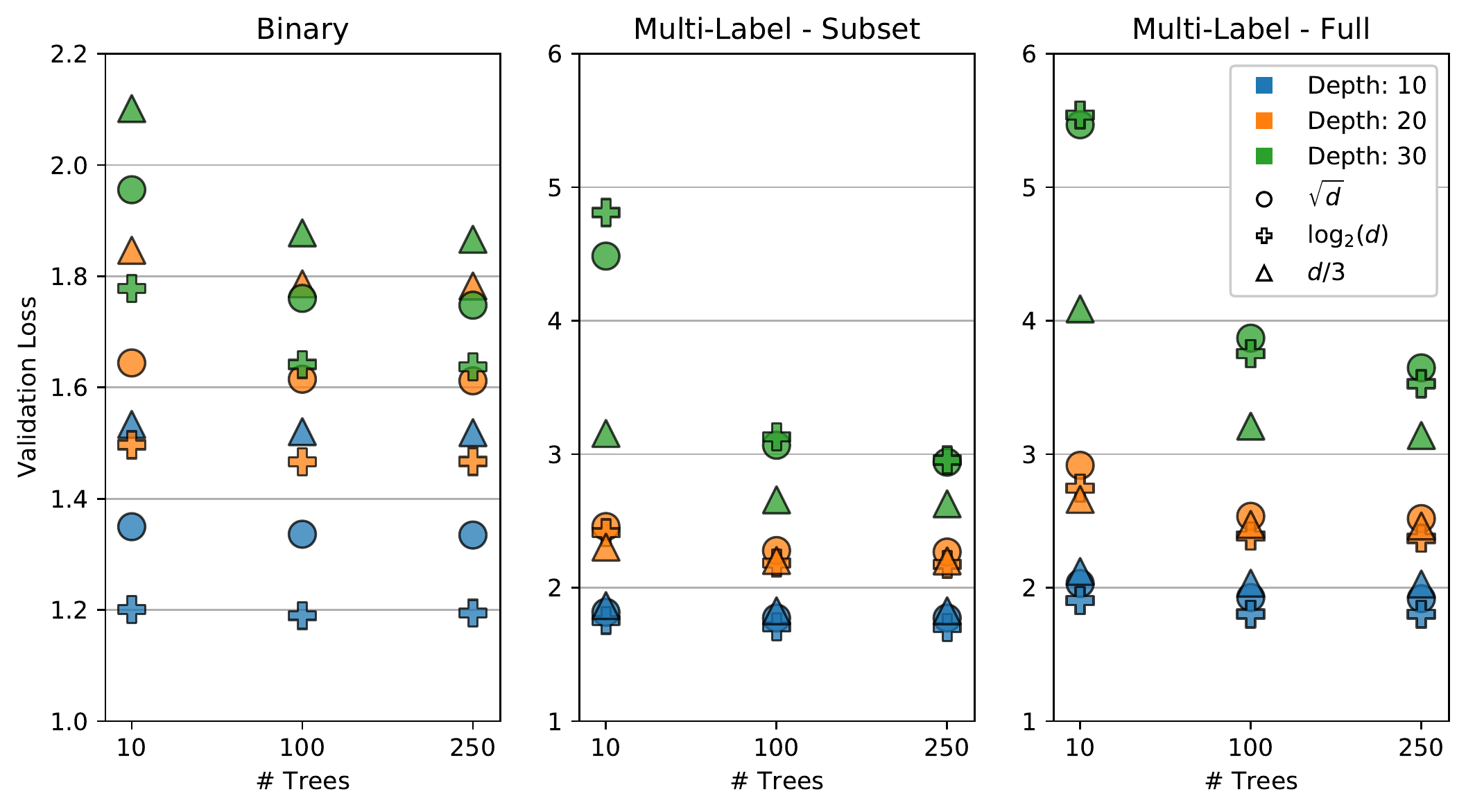}
    \caption{\textbf{Extra Trees grid search results.} Results of the grid search of the Extra Trees classifiers for: Binary classifier trained on full dataset, multi-label classifier trained on a subset of the dataset, and multi-label classifier trained on the full dataset.}
    \label{fig:myransLoss}
\end{figure*}

For the system proposed by Myrans \etal \cite{Myrans2018MultiClass, Myrans2018Binary}, two Extra Trees classifiers are used in sequence. However, the hyperparameters of the trees are not specified. Therefore, we conduct a small grid search across three hyperparameters: The amount of trees in the ensemble, the maximum depth of the trees, and the maximum amount of features used when splitting an internal node. The investigated parameters are reported in Table~\ref{tab:TreeHyperparameterSearch}. We train the Extra Trees classifier in three settings. First, we train under a binary setting determining whether there is \textit{any} class in the image. Thereafter, we train a multi-label setting, first on a subset of the dataset only containing images with annotated classes, and secondly on the full dataset. The resulting validation losses of the hyperparameter search is shown in Figure~\ref{fig:myransLoss}. From this we conclude that for the binary Extra Trees classifier 100 trees, with a maximum depth of 10 and using $\log_2(d)$ features when splitting, should be used. Similarly, we find that for the multi-label Extra Trees classifiers 250 trees, with a maximum depth of 10 and using $\log_2(d)$ features when splitting, should be used.

\begin{table}[!t]
\centering
\caption{\textbf{Extra Trees grid search intervals.} Hyperparameter search intervals for the Extra Trees classifiers. \textit{d} denotes the dimensionality of the GIST descriptor.}
\label{tab:TreeHyperparameterSearch}
\begin{tabular}{lc}\hline
\textbf{Parameter}       & \textbf{Values}                         \\ \hline
Number of Trees & {[}10, 100, 250{]}             \\
Max Depth       & {[}10, 20, 30{]}               \\
Max Features    & [$\sqrt{d}$, $\log_2(d)$, $d / 3$] \\ \hline
\end{tabular}
\end{table}

\section{CNN Loss Curves}\label{sec:lossCurves}
\begin{figure*}[!t]
    \centering
    \includegraphics[width=0.95\linewidth]{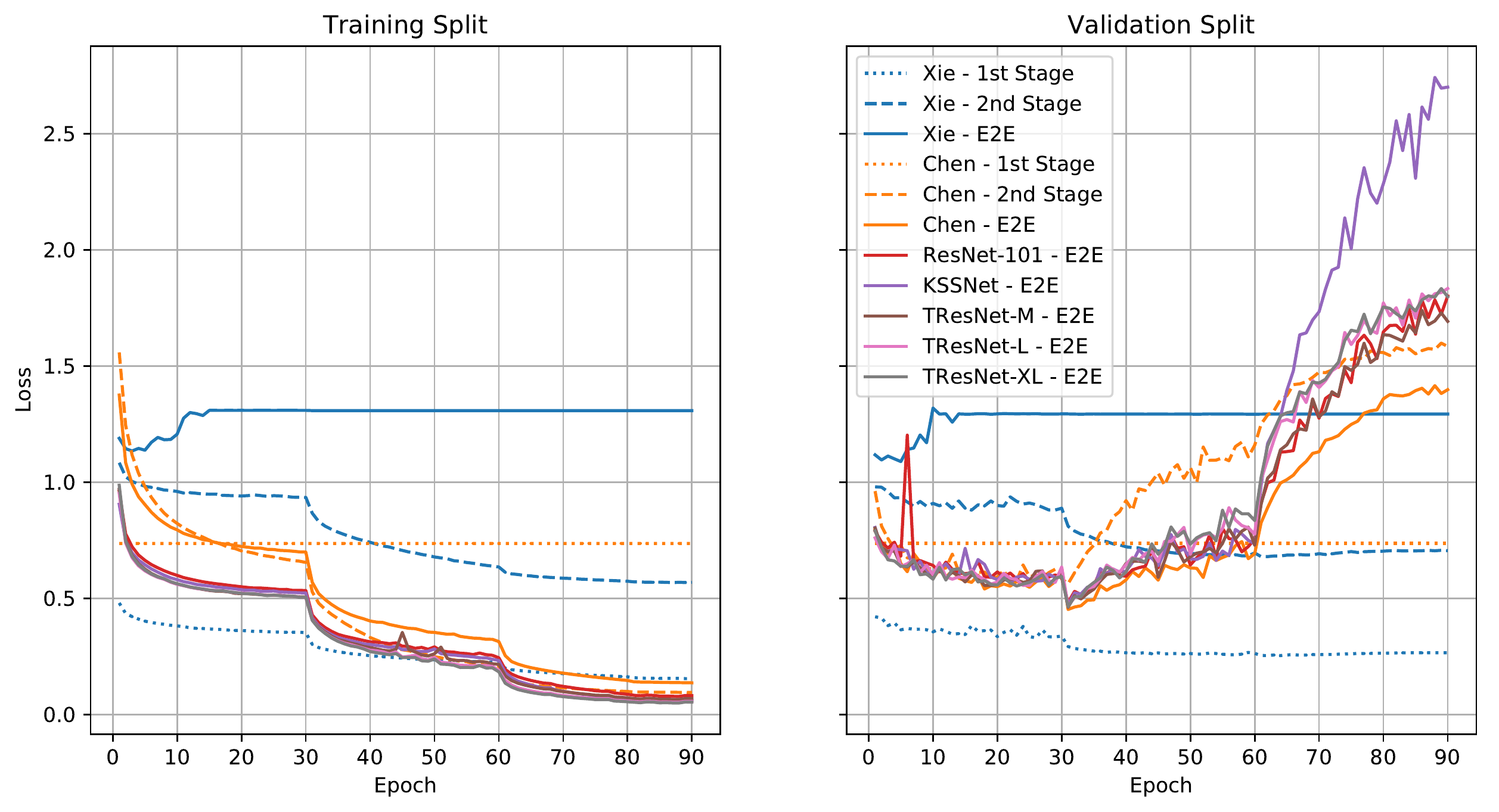}
    \caption{\textbf{Multi-label CNN loss curves.} The training and validation loss curves for all tested networks. ``1st Stage'' indicates a binary classifier, ``2nd Stage'' indicates a multi-label classifier trained on a subset of the dataset, and ``E2E'' indicates a multi-label classifier trained in an end-to-end manner with the full dataset.}
    \label{fig:netowrkLoss}
\end{figure*}

We present the loss curves for all the tested convolutional neural networks (CNNs) tested, see Figure~\ref{fig:netowrkLoss}. All networks are trained using the weighted binary cross-entropy loss, and using hyperparameters set based on the guidelines from Goyal \etal \cite{Goyal17}. Further training details are presented in the main manuscript.

From the loss plots we observe that the validation loss of the majority of the tested networks start diverging after approximately 30-40 epochs, a clear sign of overfitting. The method by Xie \etal \cite{Xie2019HierCNN} is an exception, with the first and second stage methods stagnating after 60--70 epochs. We also observe that the first stage of Chen \etal \cite{Chen2018}, the SqueezeNet \cite{SqueezeNet}, has a constant loss value for both the training and validation loss. Similarly, the second stage of Xie \etal settles on a constant loss after the initial 10 epochs when trained on the full dataset.

\begin{table}[!t]
    \centering
    \caption{\textbf{Effect of binary stage in two-stage classifiers.} We present the metric performance for the two-stage methods, comparing the effect of the full pipeline and using only the multi-label classifier. TS denotes that both stages are used, otherwise only the second stage is used.}
    \resizebox{\linewidth}{!}{%
    \begin{tabular}{c|c|cc|cc} \hline
    \multicolumn{1}{c|}{\multirow{2}{*}{\textbf{Model}}} & \multirow{2}{*}{\textbf{TS}} & \multicolumn{2}{c|}{\textbf{Validation}} & \multicolumn{2}{c}{\textbf{Test}} \\ \cline{3-6} 
\multicolumn{1}{c|}{}                        &                        & \textbf{$\text{F}2_{\text{CIW}}$} $\uparrow$        & \textbf{$\text{F}1_{\text{Normal}}$} $\uparrow$        & \textbf{$\text{F}2_{\text{CIW}}$} $\uparrow$     & \textbf{$\text{F}1_{\text{Normal}}$} $\uparrow$   \\ \hline

    \multicolumn{1}{c|}{\multirow{2}{*}{Xie \cite{Xie2019HierCNN}}}     &\checkmark &\textbf{ 48.57}        & \textbf{91.08}            & \textbf{48.34}     & \textbf{90.62} \\
         & & 37.65 & 0.52 & 37.83 & 0.68  \\ \hline
    \multicolumn{1}{c|}{\multirow{2}{*}{Chen \cite{Chen2018}}}     &\checkmark & 42.03        & 3.96             & 41.74     & 3.59      \\
         & & 42.03        & 3.96             & 41.74     & 3.59     \\ \hline
    \multicolumn{1}{c|}{\multirow{2}{*}{Myrans \cite{Myrans2018MultiClass}}}     &\checkmark &    4.01 & 26.03           & 4.11 & 27.48    \\
         &  & 19.25 & 0.00  & 19.19 & 0.00 \\ \hline
    \end{tabular}%
    }
    \label{tab:twoStageRes}
\end{table}

\begin{table}[!t]
    \centering
    \caption{\textbf{Effect of training second stage on full dataset.} The metric performance for the two-stage methods, when training both stages on the full dataset. TS denotes that both stages are used, otherwise only the second stage is used. }
    \resizebox{\linewidth}{!}{%
    \begin{tabular}{c|c|cc|cc} \hline
    \multicolumn{1}{c|}{\multirow{2}{*}{\textbf{Model}}} & \multirow{2}{*}{\textbf{TS}} & \multicolumn{2}{c|}{\textbf{Validation}} & \multicolumn{2}{c}{\textbf{Test}} \\ \cline{3-6} 
\multicolumn{1}{c|}{}                        &                        & \textbf{$\text{F}2_{\text{CIW}}$} $\uparrow$         & \textbf{$\text{F}1_{\text{Normal}}$ } $\uparrow$       & \textbf{$\text{F}2_{\text{CIW}}$ } $\uparrow$     & \textbf{$\text{F}1_{\text{Normal}}$} $\uparrow$   \\ \hline
    \multicolumn{1}{c|}{\multirow{2}{*}{Xie \cite{Xie2019HierCNN}}}     &\checkmark & 31.98 & \textbf{88.23}  &  31.82 & \textbf{87.95}        \\
        &  & 28.12 & 59.98  & 27.96 & 59.99       \\ \hline
         \multicolumn{1}{c|}{\multirow{2}{*}{Chen \cite{Chen2018}}}     &\checkmark & \textbf{43.45} & 76.73 & \textbf{43.14} & 75.68      \\
         & & \textbf{43.45} & 76.73 & \textbf{43.14} & 75.68      \\ \hline
    \multicolumn{1}{c|}{\multirow{2}{*}{Myrans \cite{Myrans2018MultiClass}}}     &\checkmark & 2.58 & 25.98 & 2.61 & 27.48  \\
         &  & 7.48 & 0.00  & 7.37 & 0.00 \\ \hline
    \end{tabular}%
    }
    \label{tab:twoStageData}
\end{table}

\section{Two-Stage Ablation Study}\label{sec:ablationSup}
We conduct two ablation studies on the two-stage classifiers, to determine the effect of the different stages and training methodology.\\

\textbf{What is the effect of the binary classifier?} We compare the effect on performance of using both stages or only the second stage. These results are presented in Table~\ref{tab:twoStageRes}, and indicate that the first stage is crucial. Performance for Xie \etal \cite{Xie2019HierCNN} degrades for both metrics when the first stage is missing, whereas for Chen \etal \cite{Chen2018} there is no difference as the first stage never predicts a normal pipe. For Myrans \etal \cite{Myrans2018MultiClass} the first stage inaccurately classifies images with classes as normal pipes, causing a lower $\text{F}2_{\text{CIW}}$ score. This is improved when using only the second stage, but at the cost of an inability to recognize any normal pipes.

\textbf{Training the second stage on the full dataset.} Classically within the sewer classification domain, the second stage is only trained on data which contains some kind of class. We investigate whether performance improves by training on the full dataset, such that the second stage also sees normal pipes. The results are shown in Table~\ref{tab:twoStageData}. For Myrans \etal the performance is reduced substantially in both tested settings, and the second stage is still unable to classify normal pipes. For Xie \etal both metrics are lower when comparing to Table~\ref{tab:twoStageRes}, except for the large increase in $\text{F}_{\text{Normal}}$ score when only using the second stage. The only performance improvement is achieved by Chen \etal through the use of the deeper InceptionV3 network.

\section{Multi-Label Metrics and Results}\label{sec:metricsSup}
When evaluating multi-label tasks, a large suite of metrics are commonly used, in order to uncover different aspects of the tested methods. Commonly, the F1-score is used in different variations, depending on how the F1-score is calculated or averaged. An overview of the different metrics is provided in Table~\ref{tab:metricOverview}. Each of the metrics are in the range $[0, 1]$, and for all a high score is better. As a reference on how to compute the metrics, we refer to the supplementary materials of the work by Durand \etal \cite{PartialLabel20}. We present the classic performance metrics for each of the tested methods on both the validation and test splits, as well as the per-class F1, F2, Recall, Precision, and Average Precision (AP). It should be noted, that AP cannot be calculated for the normal class. This is due to the normal class being an implicit class, and therefore not possible to rank as it does not have a single associated probability. The Kumar \etal \cite{Kumar2018EnsembleCNN}, Meijer \etal \cite{Meijer2019MultiLabel} and ML-GCN \cite{MLGCN19} methods are not shown in the metric tables as the models diverged during training. The benchmark algorithm consisting of the first stage from Xie \etal \cite{Xie2019HierCNN} and the TResNet-L multi-label classifier \cite{TResNet20} is reported as ``Benchmark''. The metrics for the validation split are presented in Table~\ref{tab:mainResVal}-\ref{tab:classAPVal} and the metrics for the test split are presented in Table~\ref{tab:mainResTest}-\ref{tab:classAPTest}.

\clearpage
\onecolumn
\def\tableTemp{\thetable}
\renewcommand\tablename{Figure}
\setcounter{table}{\thefigure}
\begin{longtable}[c]{m{0.9cm}m{2.69cm}m{2.69cm}m{2.69cm}m{2.69cm}m{2.69cm}}
\caption{\textbf{Sewer-ML data examples.} A subset of the images in the Sewer-ML showcasing five images from each of the annotated classes as well as normal pipes in each row. The class codes are described in Table~\ref{tab:ClassOverview}.}\\
\endfirsthead
\multicolumn{6}{c}%
        {{ Figure \thetable: \textbf{Continued from previous page}}} \\
\endhead
\multicolumn{6}{r}{{Continued on next page}} 
\endfoot
\endlastfoot
\centering
\textbf{RB}     & \includegraphics[width=1\linewidth]{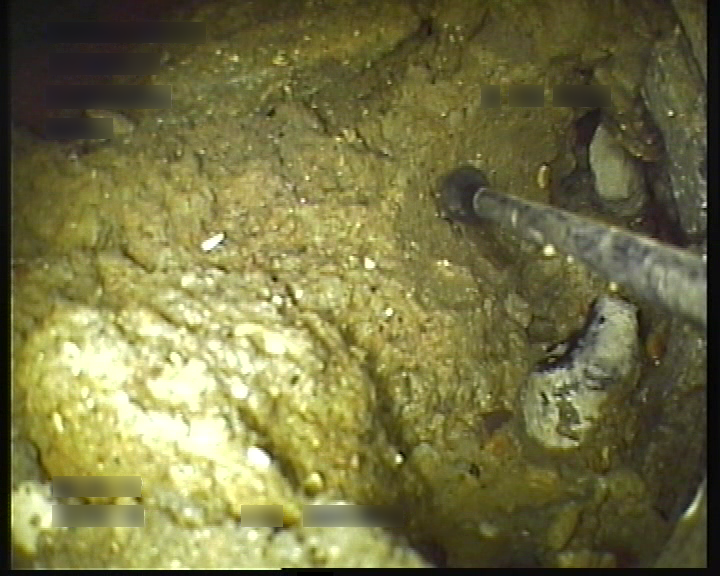} & \includegraphics[width=1\linewidth]{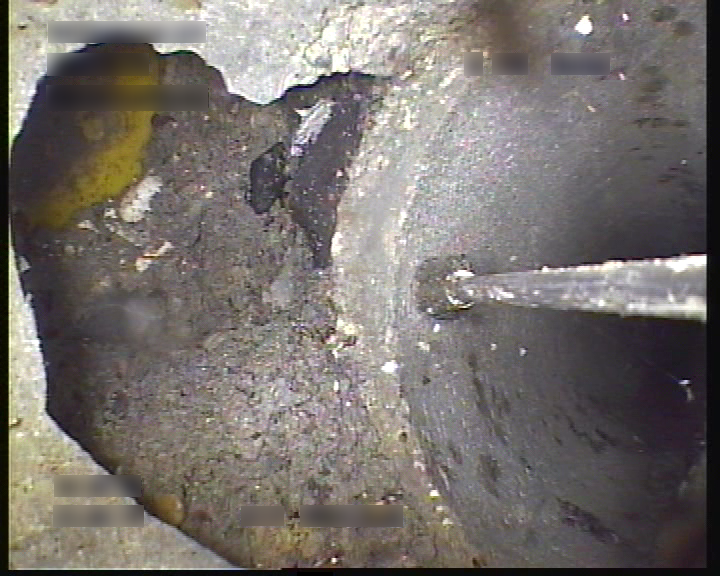} & \includegraphics[width=1\linewidth]{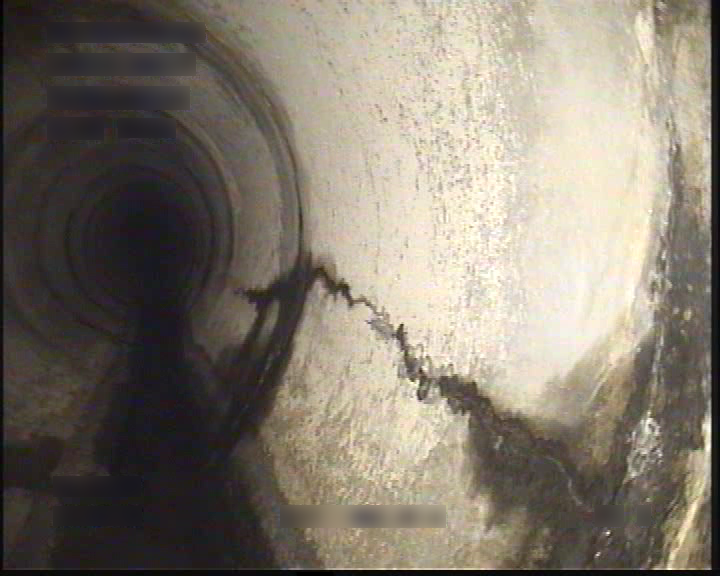} & \includegraphics[width=1\linewidth]{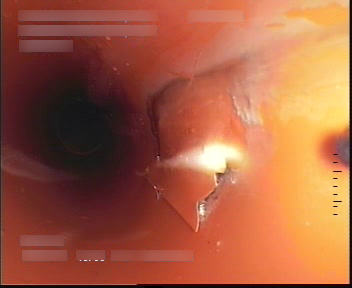} & \includegraphics[width=1\linewidth]{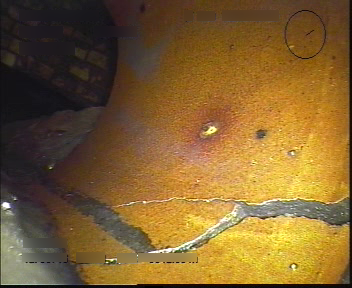}  \\
\textbf{OB}     & \includegraphics[width=1\linewidth]{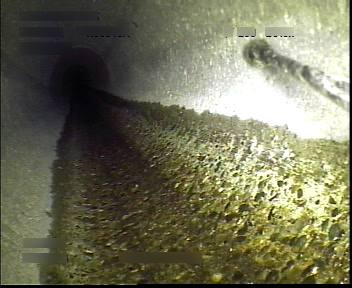} & \includegraphics[width=1\linewidth]{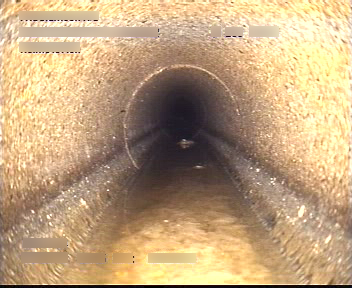} & \includegraphics[width=1\linewidth]{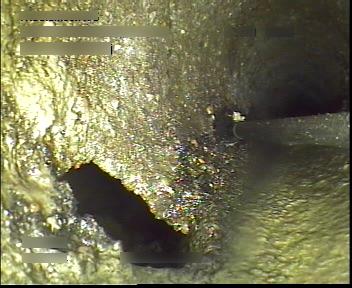} & \includegraphics[width=1\linewidth]{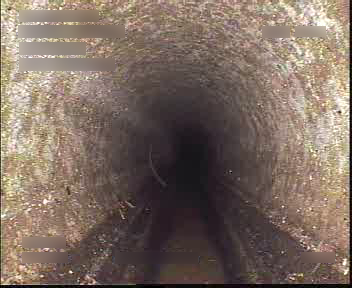} & \includegraphics[width=1\linewidth]{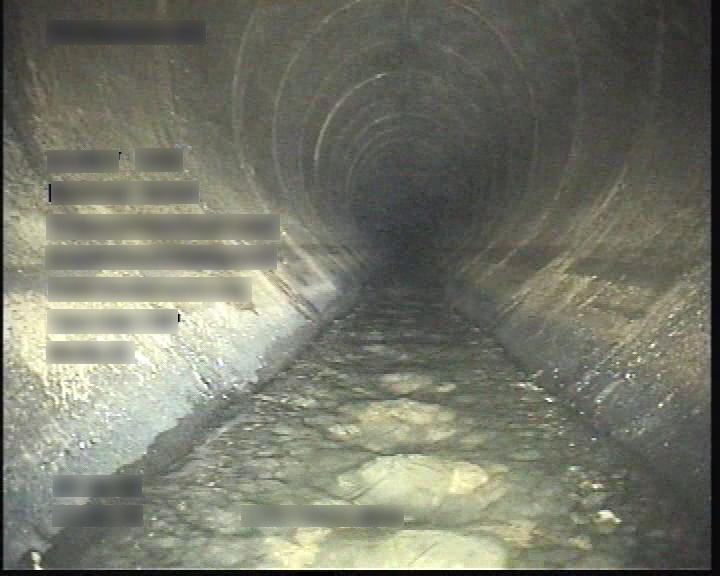}  \\
\textbf{PF}     & \includegraphics[width=1\linewidth]{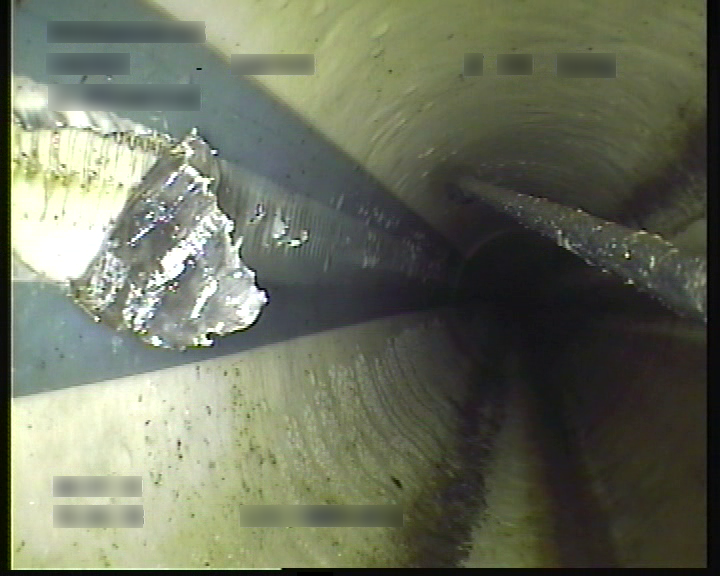} & \includegraphics[width=1\linewidth]{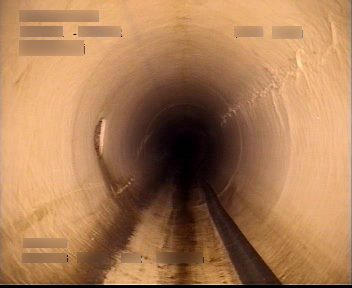} & \includegraphics[width=1\linewidth]{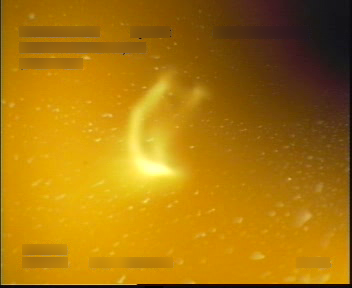} & \includegraphics[width=1\linewidth]{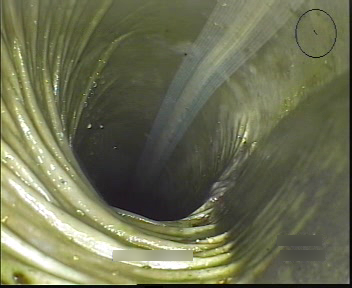} & \includegraphics[width=1\linewidth]{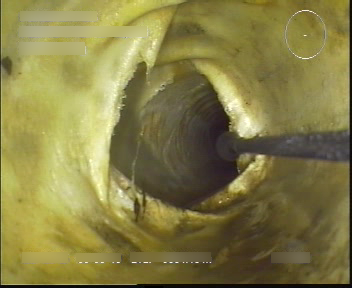}  \\
\textbf{DE}     & \includegraphics[width=1\linewidth]{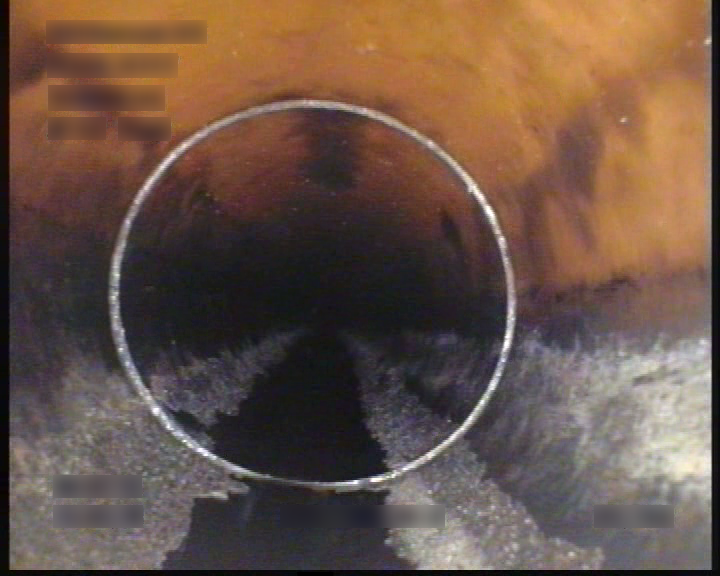} & \includegraphics[width=1\linewidth]{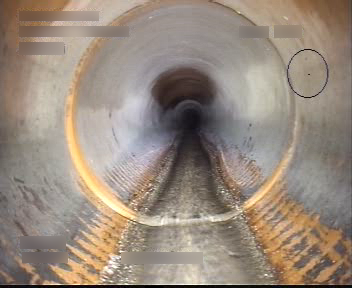} & \includegraphics[width=1\linewidth]{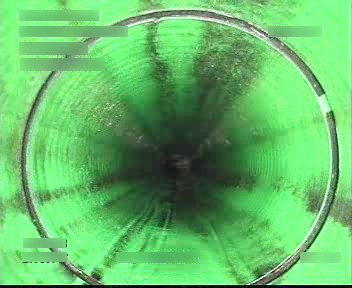} & \includegraphics[width=1\linewidth]{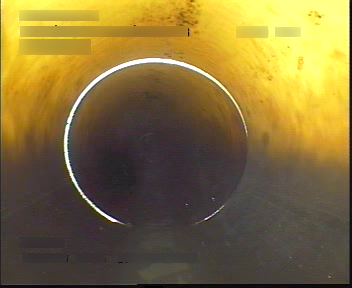} & \includegraphics[width=1\linewidth]{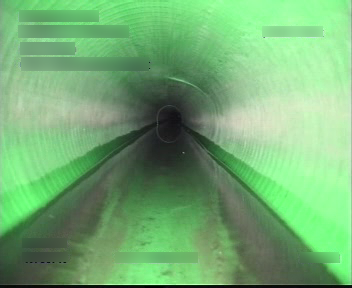}  \\
\textbf{FS}     & \includegraphics[width=1\linewidth]{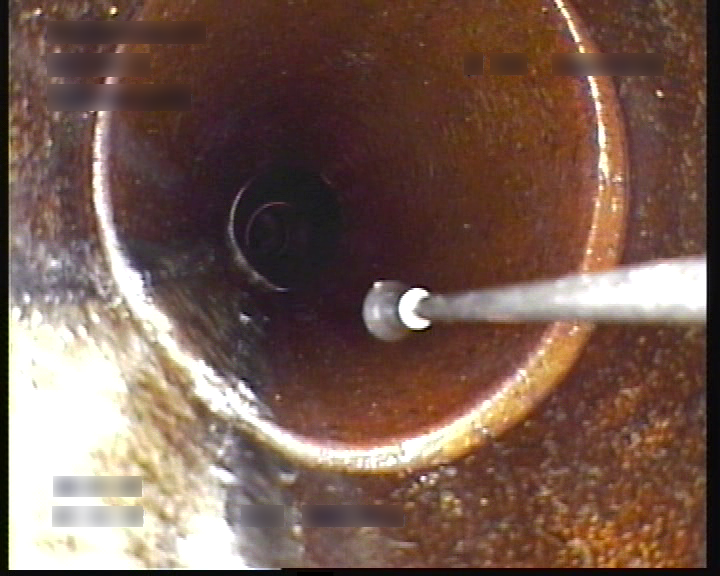} & \includegraphics[width=1\linewidth]{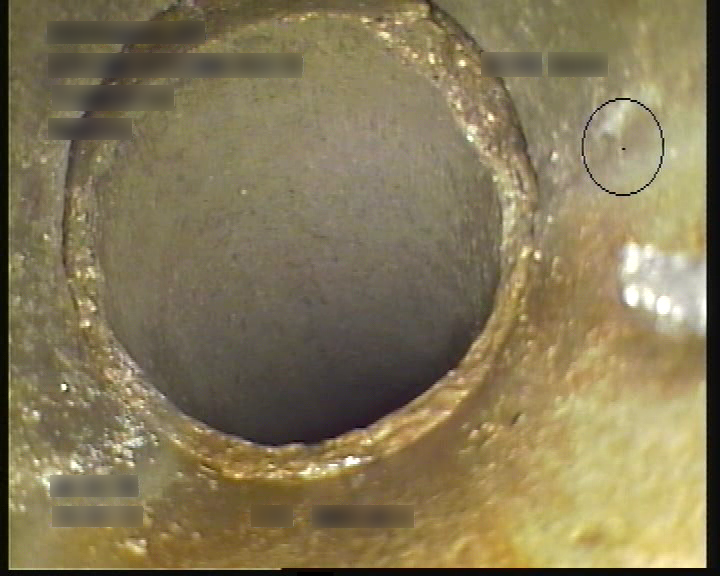} & \includegraphics[width=1\linewidth]{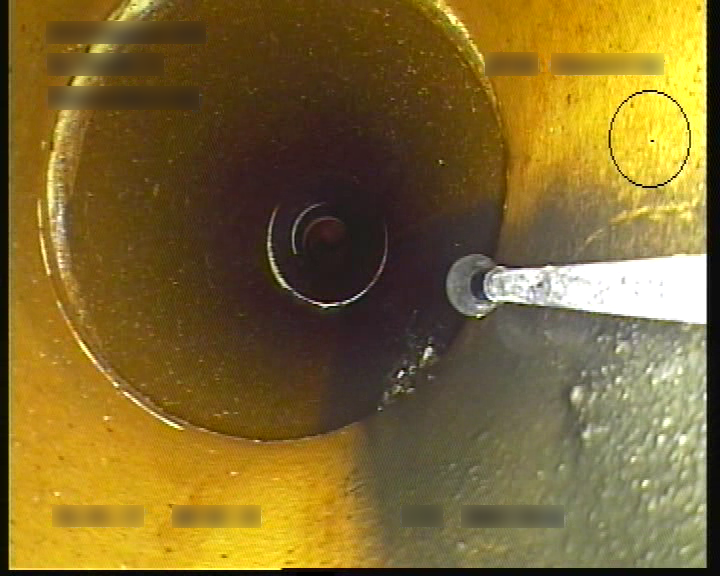} & \includegraphics[width=1\linewidth]{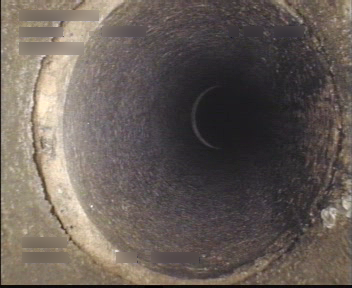} & \includegraphics[width=1\linewidth]{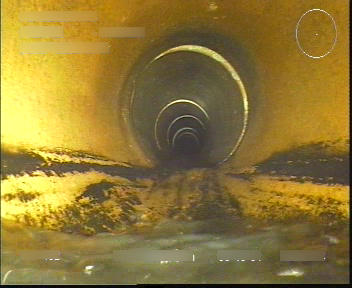}  \\
\textbf{IS}     & \includegraphics[width=1\linewidth]{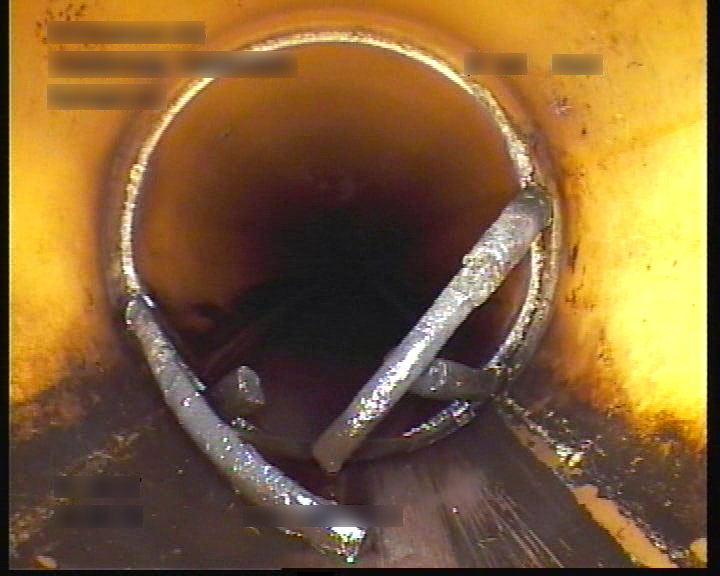} & \includegraphics[width=1\linewidth]{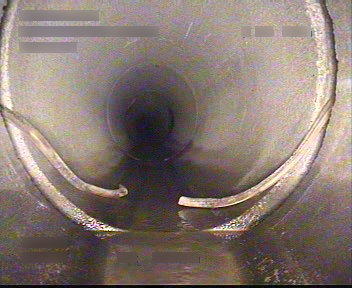} & \includegraphics[width=1\linewidth]{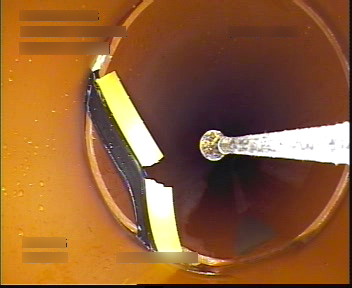} & \includegraphics[width=1\linewidth]{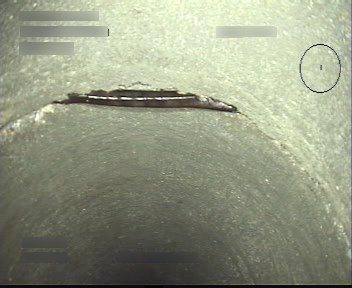} & \includegraphics[width=1\linewidth]{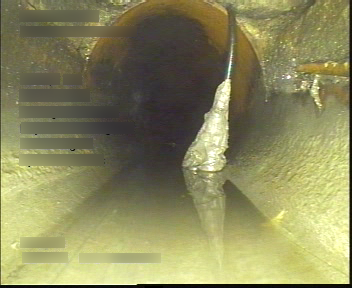}  \\
\textbf{RO}     & \includegraphics[width=1\linewidth]{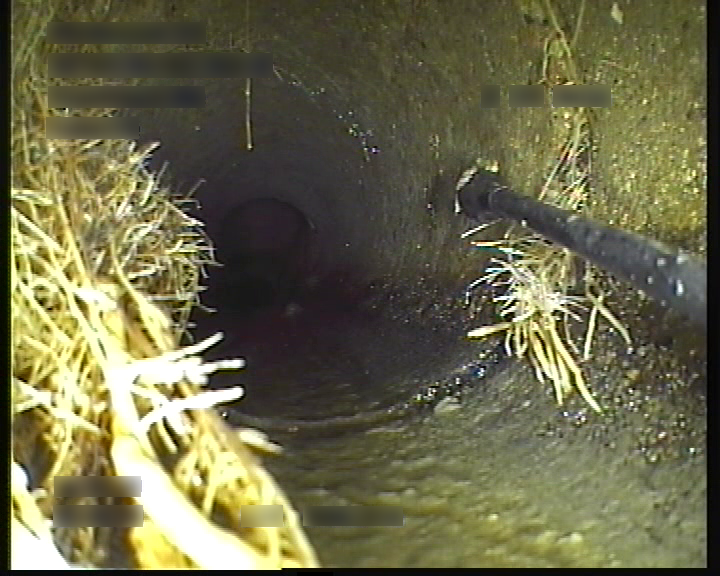} & \includegraphics[width=1\linewidth]{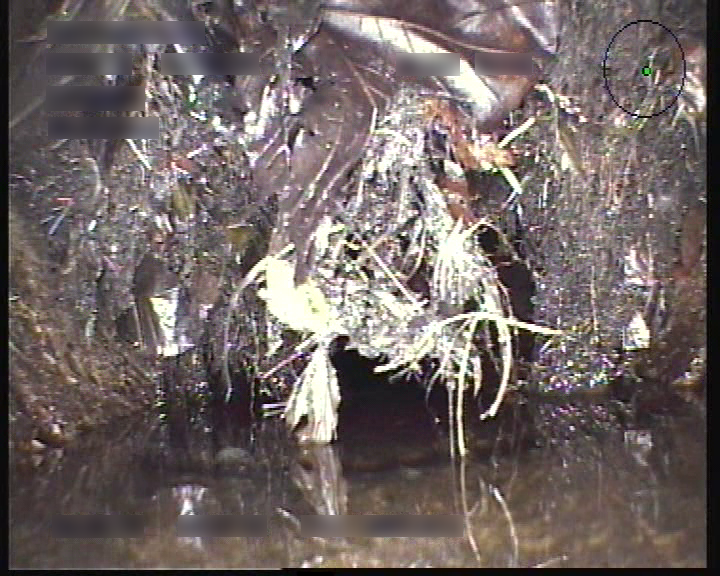} & \includegraphics[width=1\linewidth]{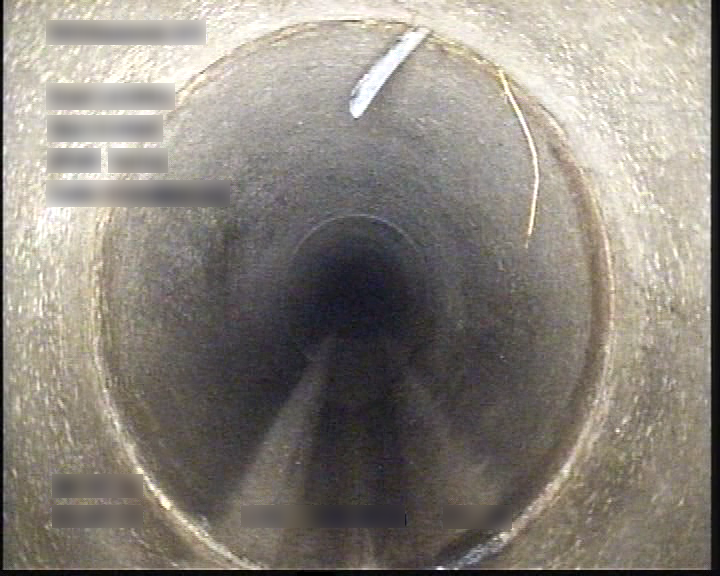} & \includegraphics[width=1\linewidth]{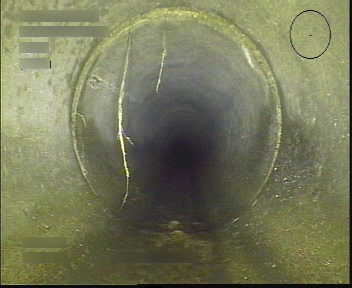} & \includegraphics[width=1\linewidth]{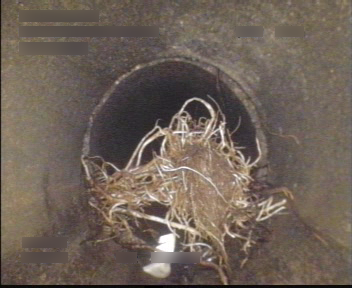}  \\
\textbf{IN}     & \includegraphics[width=1\linewidth]{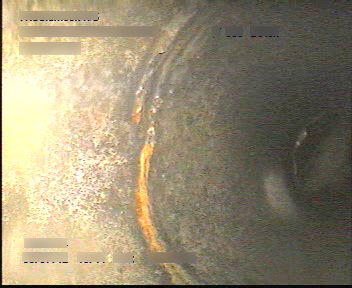} & \includegraphics[width=1\linewidth]{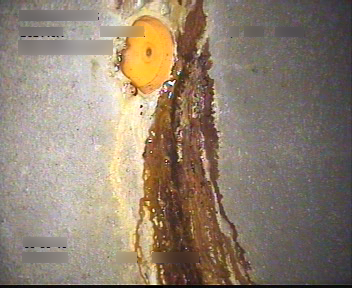} & \includegraphics[width=1\linewidth]{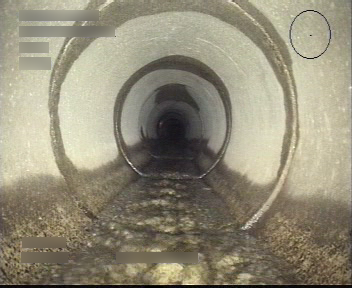} & \includegraphics[width=1\linewidth]{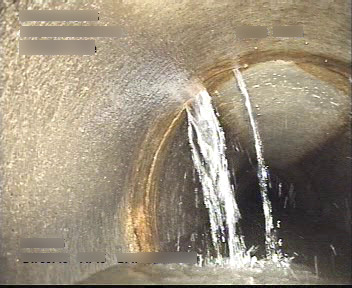} & \includegraphics[width=1\linewidth]{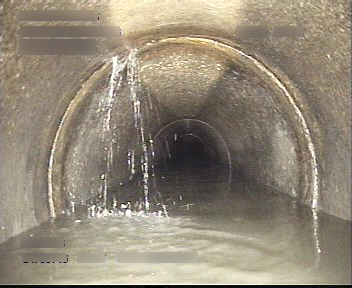}  \\
\textbf{AF}     & \includegraphics[width=1\linewidth]{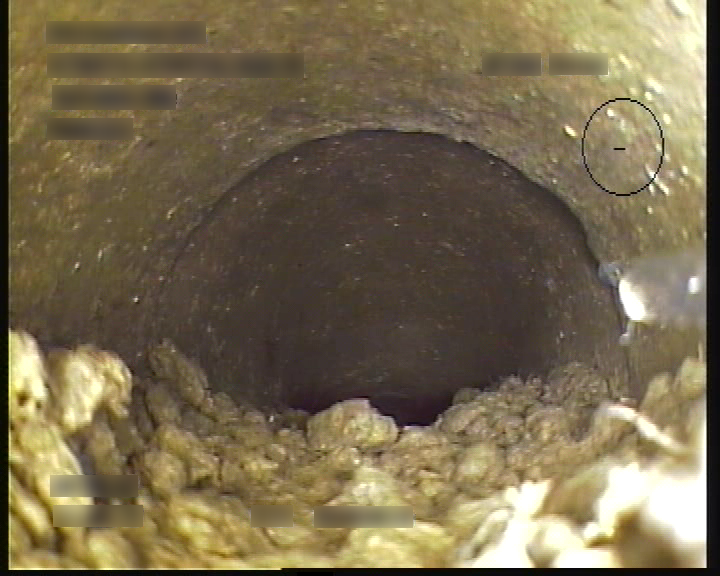} & \includegraphics[width=1\linewidth]{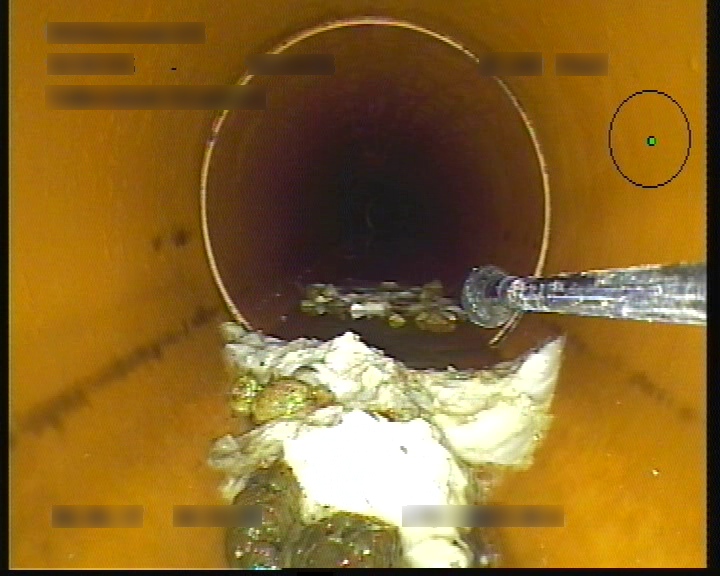} & \includegraphics[width=1\linewidth]{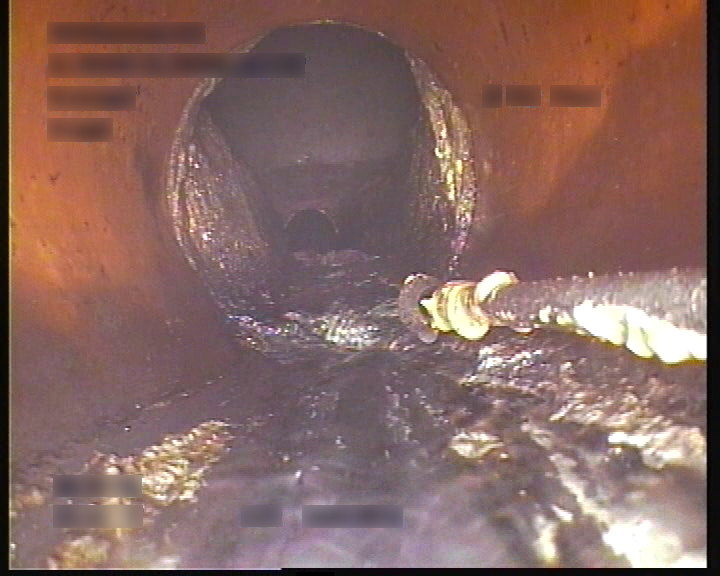} & \includegraphics[width=1\linewidth]{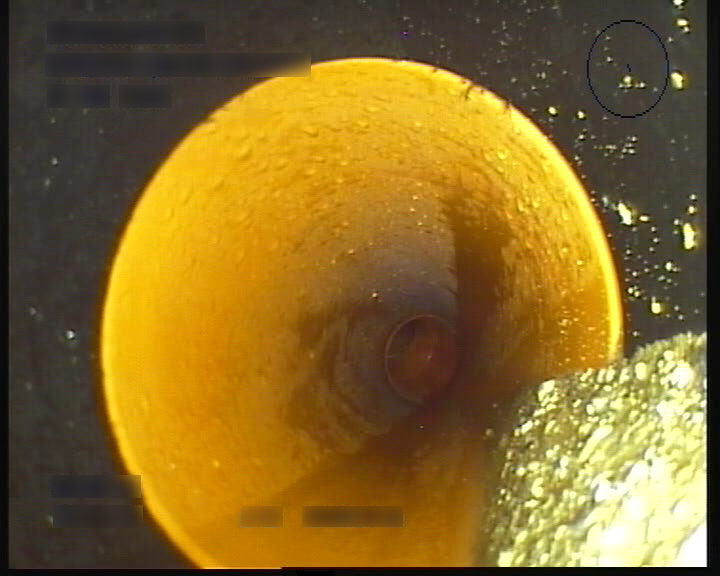} & \includegraphics[width=1\linewidth]{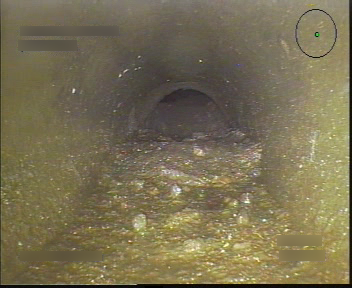}  \\
\textbf{BE}     & \includegraphics[width=1\linewidth]{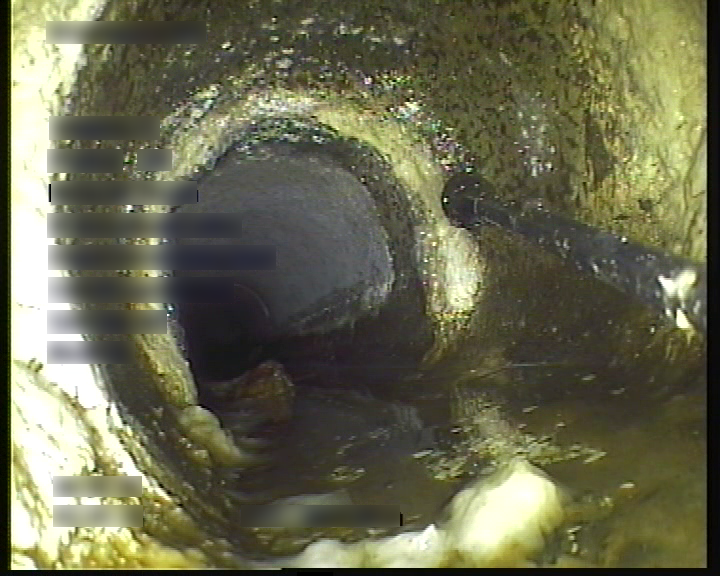} & \includegraphics[width=1\linewidth]{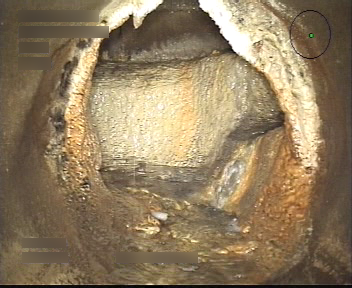} & \includegraphics[width=1\linewidth]{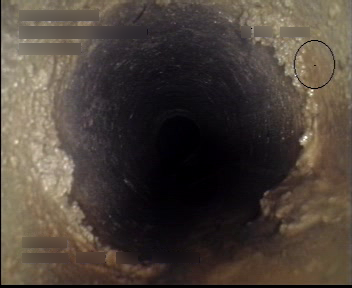} & \includegraphics[width=1\linewidth]{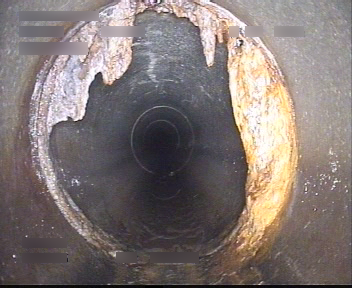} & \includegraphics[width=1\linewidth]{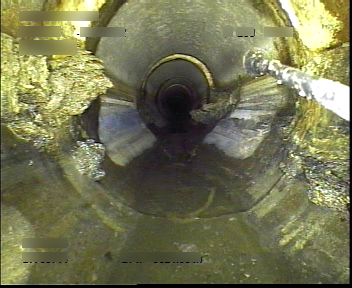}  \\
\textbf{FO}     & \includegraphics[width=1\linewidth]{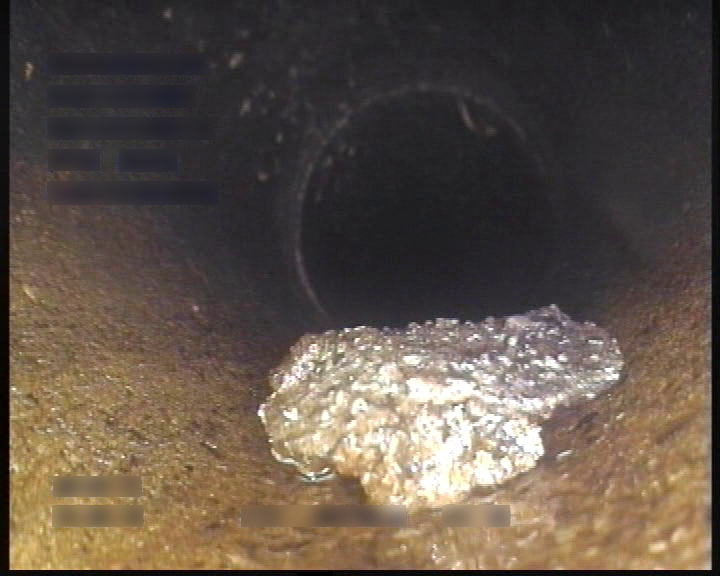} & \includegraphics[width=1\linewidth]{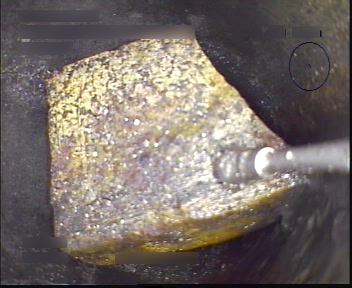} & \includegraphics[width=1\linewidth]{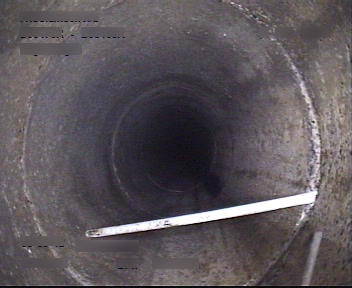} & \includegraphics[width=1\linewidth]{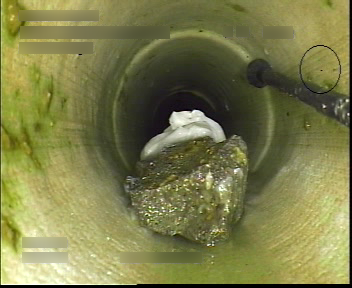} & \includegraphics[width=1\linewidth]{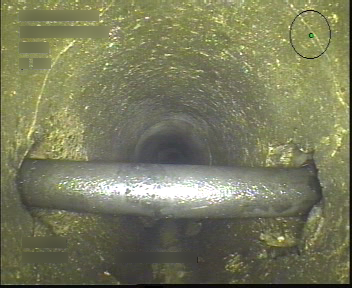}  \\
\textbf{GR}     & \includegraphics[width=1\linewidth]{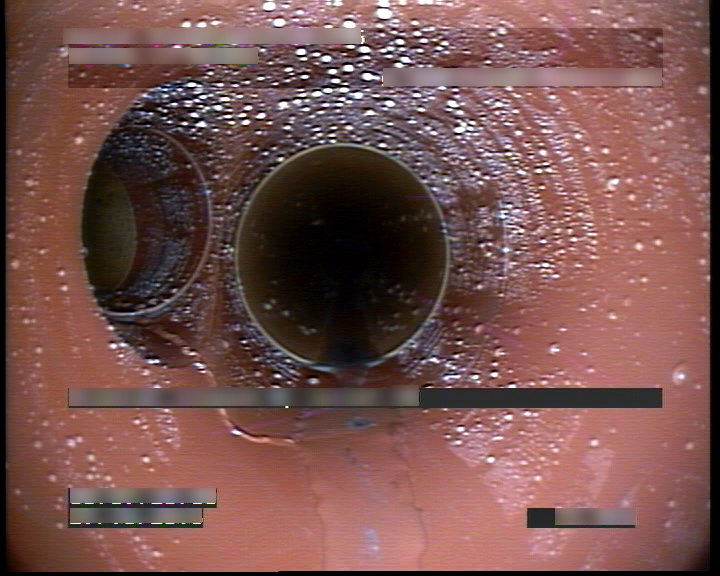} & \includegraphics[width=1\linewidth]{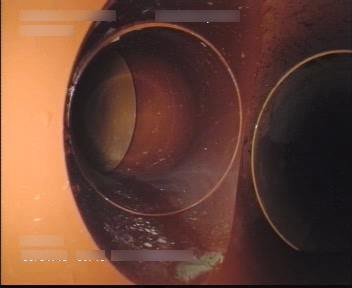} & \includegraphics[width=1\linewidth]{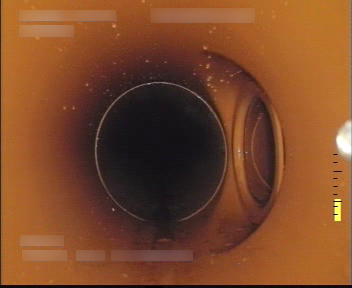} & \includegraphics[width=1\linewidth]{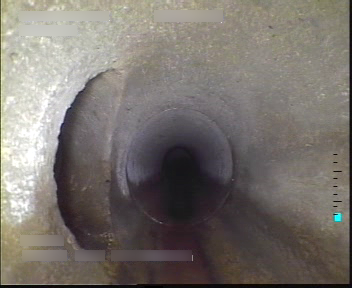} & \includegraphics[width=1\linewidth]{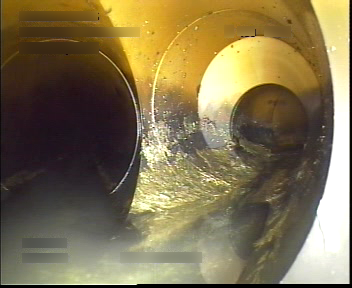}  \\
\textbf{PH}     & \includegraphics[width=1\linewidth]{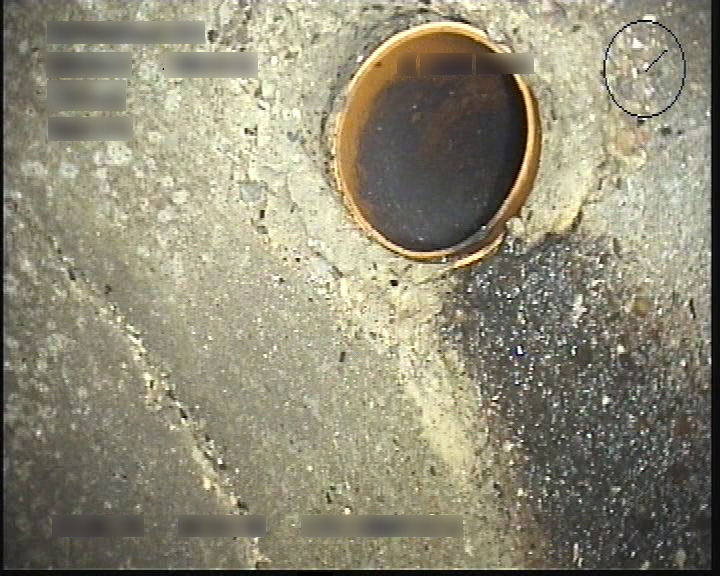} & \includegraphics[width=1\linewidth]{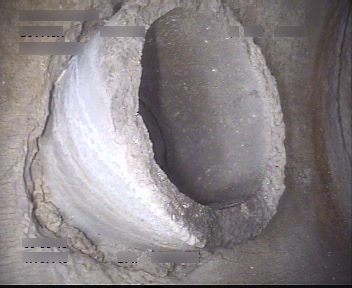} & \includegraphics[width=1\linewidth]{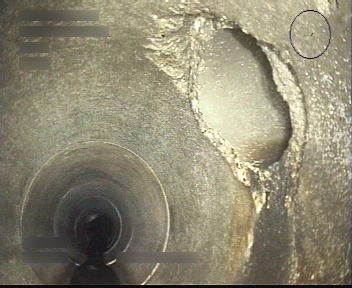} & \includegraphics[width=1\linewidth]{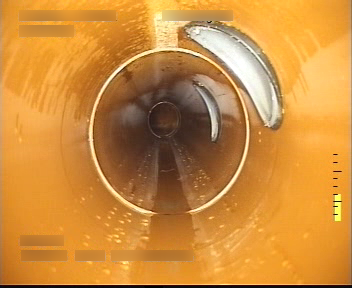} & \includegraphics[width=1\linewidth]{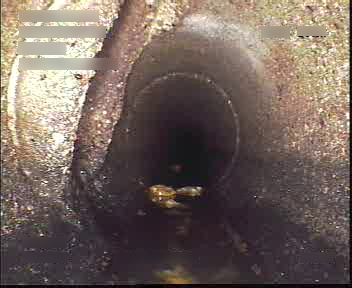}  \\
\textbf{PB}     & \includegraphics[width=1\linewidth]{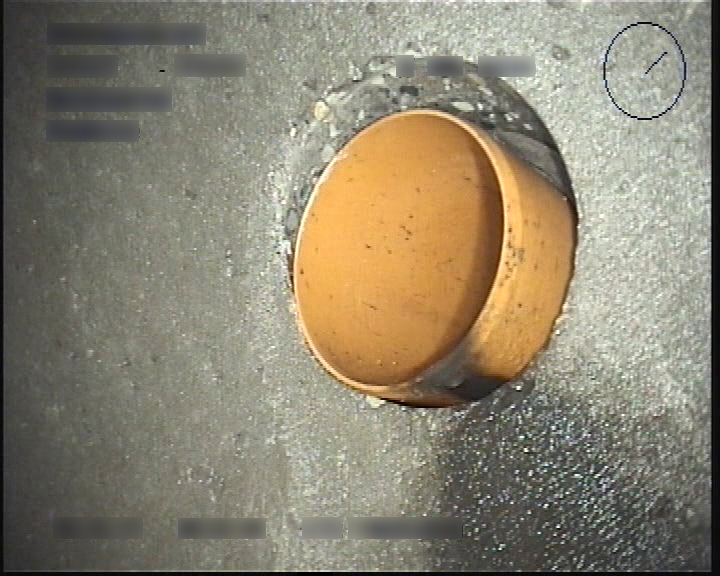} & \includegraphics[width=1\linewidth]{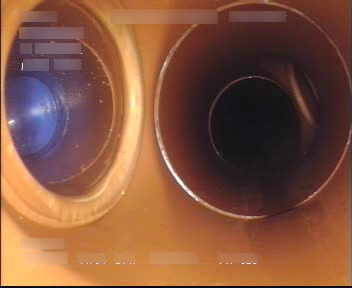} & \includegraphics[width=1\linewidth]{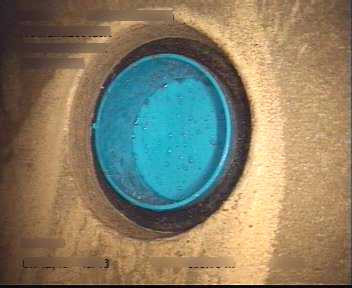} & \includegraphics[width=1\linewidth]{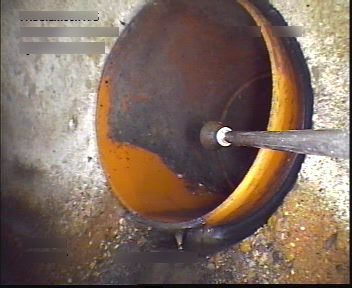} & \includegraphics[width=1\linewidth]{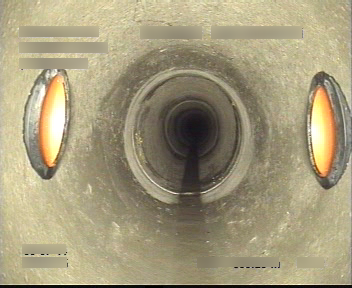}  \\
\textbf{OS}     & \includegraphics[width=1\linewidth]{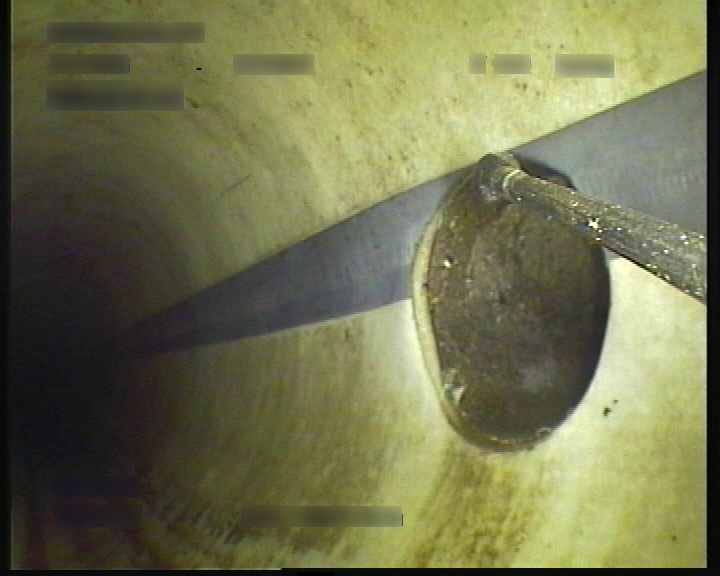} & \includegraphics[width=1\linewidth]{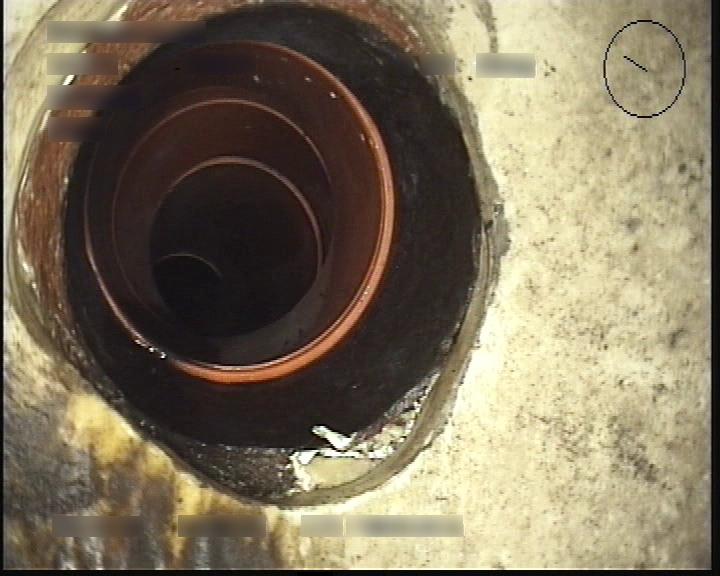} & \includegraphics[width=1\linewidth]{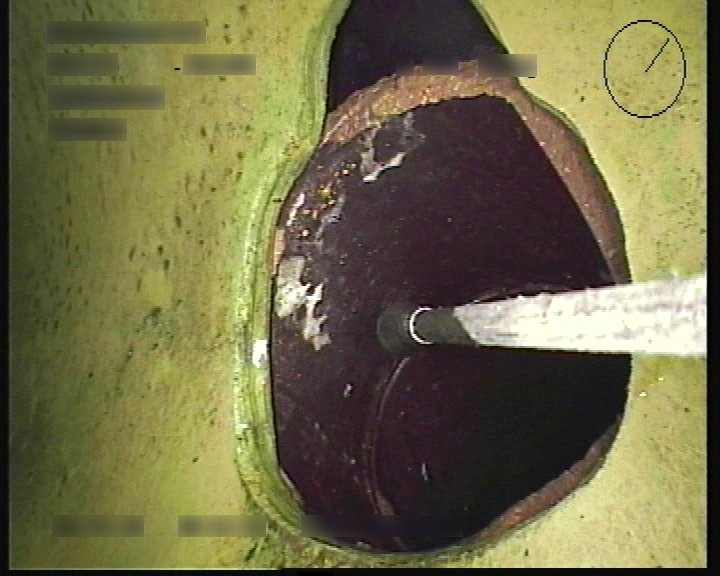} & \includegraphics[width=1\linewidth]{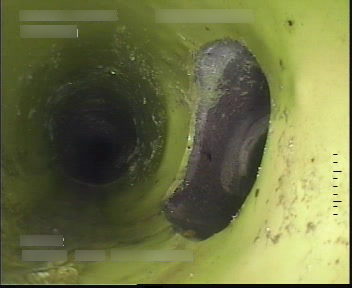} & \includegraphics[width=1\linewidth]{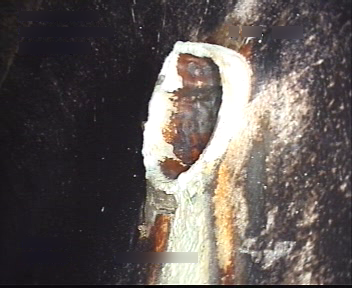}  \\
\textbf{OP}     & \includegraphics[width=1\linewidth]{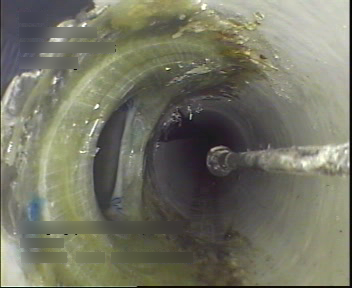} & \includegraphics[width=1\linewidth]{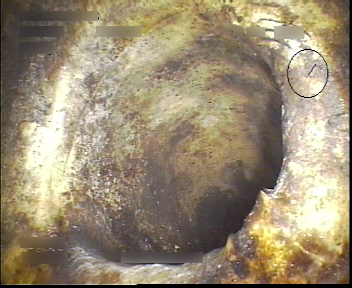} & \includegraphics[width=1\linewidth]{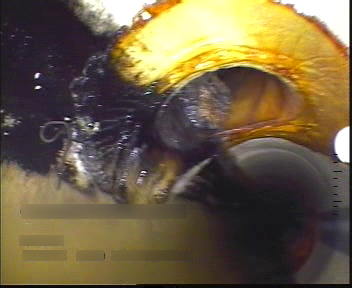} & \includegraphics[width=1\linewidth]{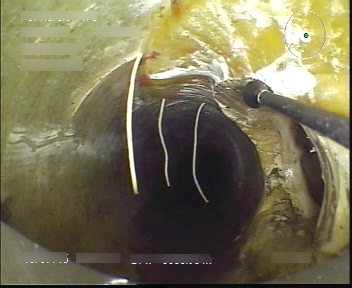} & \includegraphics[width=1\linewidth]{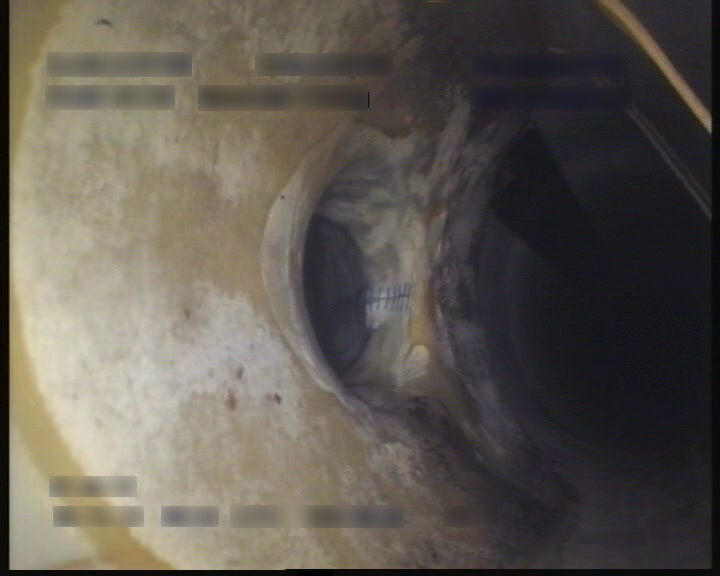}  \\
\textbf{OK}     & \includegraphics[width=1\linewidth]{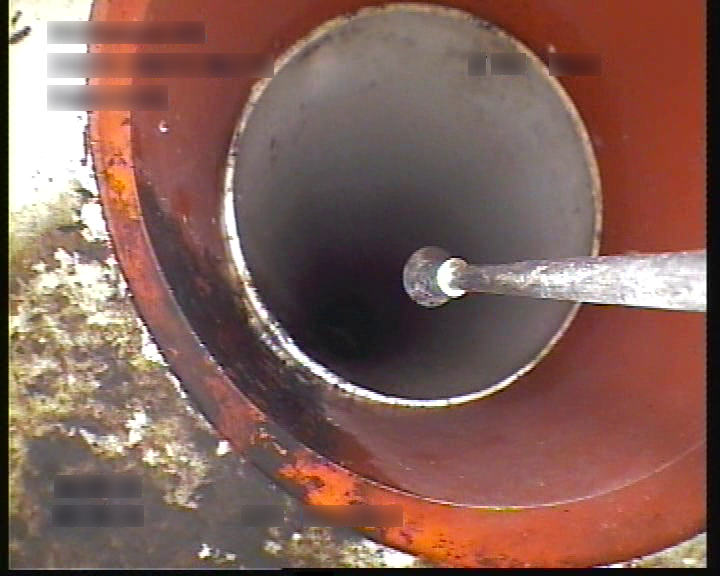} & \includegraphics[width=1\linewidth]{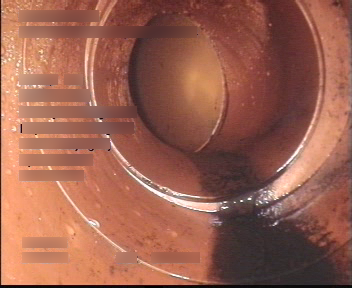} & \includegraphics[width=1\linewidth]{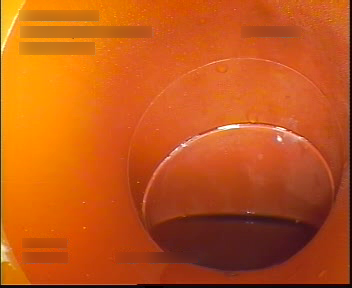} & \includegraphics[width=1\linewidth]{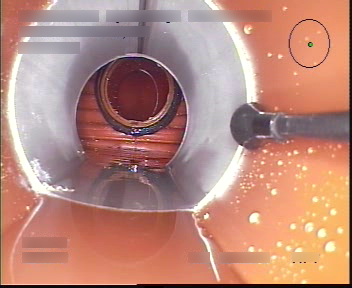} & \includegraphics[width=1\linewidth]{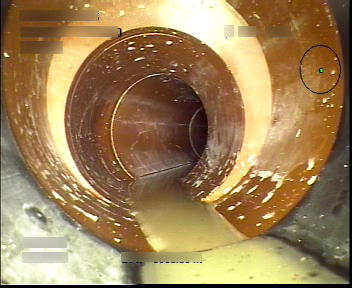}  \\
\textbf{Normal}     & \includegraphics[width=1\linewidth]{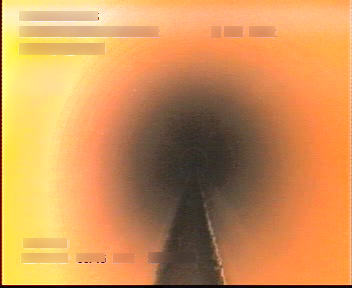} & \includegraphics[width=1\linewidth]{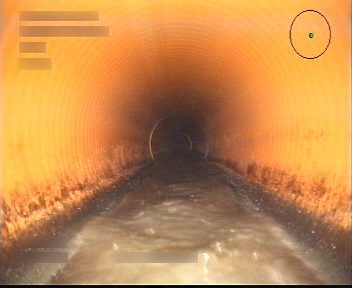} & \includegraphics[width=1\linewidth]{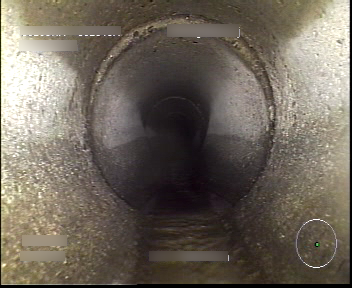} & \includegraphics[width=1\linewidth]{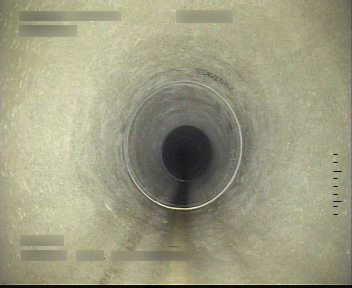} & \includegraphics[width=1\linewidth]{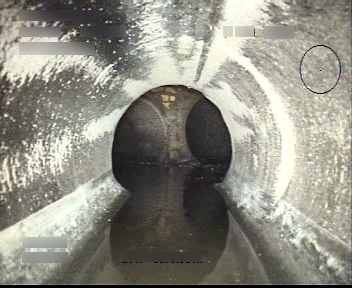}  \\
\label{tab:perClassExample}
\end{longtable}
\twocolumn
\renewcommand\tablename{Table}
\setcounter{table}{11}

\begin{table*}[!t]
\centering
\caption{\textbf{Multi-label classification metrics.} A short description of the commonly used multi-label classification metrics. For details on how the metrics are computed, we refer to Durand \etal \cite{PartialLabel20}.}
\label{tab:metricOverview}
\begin{tabular}{ll}
\hline
\textbf{Metric}                              & \textbf{Description}                                                        \\ \hline
Macro-F1 (M-F1)                     & Average F1-score across all classes.                               \\
Micro-F1 (m-F1)                     & F1 score calculated over all samples.                              \\
Overall Precision (OV-P)            & Precision metric calculated over all samples, regardless of class. \\
Overall Recall (OV-R)               & Recall metric calculated over all samples, regardless of class.    \\
Overall F1 (OV-F1)                  & F1 score calculated using OV-P and OV-R.                           \\
Per-class Precision (PC-P)          & Average precision metric across all classes.                       \\
Per-class Recall (PC-R)             & Average recall metric across all classes.                          \\
Per-class F1 (PC-F1)                & F1-score calculated using PC-P and PC-R.                            \\
Zero-one Exact Match Accuracy (0-1) & Ratio of samples with all labels correctly predicted.                \\
mean Average Precision (mAP)        & Average of the Average Precision of all annotated classes             \\ \hline
\end{tabular}
\end{table*}

\begin{table*}[!t]
    \centering
    \caption{\textbf{Performance metrics for each method - Validation Split.} The metrics are presented as percentages, and the highest score in each column is denoted in bold.}
    \begin{tabular}{clrrrrrrrrrr} \hline
    & \textbf{Model} & \textbf{m-F1} & \textbf{M-F1} & \textbf{OV-F1} & \textbf{OV-P} & \textbf{OV-R} & \textbf{PC-F1} & \textbf{PC-P} & \textbf{PC-R}& \textbf{0-1} & \textbf{mAP} \\ \hline
    \parbox[t]{2mm}{\multirow{4}{*}{\rotatebox[origin=c]{90}{Sewer}}}                                             & Xie \etal \cite{Xie2019HierCNN} & 59.33 & 38.10 & 59.33 & 46.31 & 82.52 & 42.43 & 29.61 & 74.79 & 51.64 & 66.40 \\ 
    & Chen \etal \cite{Chen2018} & 33.94 & 26.62 & 33.94 & 26.38 & 47.60 & 35.09 & 23.97 & 65.40 & 7.96 & 62.06 \\ 
    & Hassan \etal \cite{Hassan2019AlexNet} & 12.76 & 6.36 & 12.76 & 7.44 & 44.86 & 6.92 & 3.72 & 50.00 & 0.00 & 8.89 \\ 
    & Myrans \etal \cite{Myrans2018MultiClass} & 5.39 & 3.69 & 5.39 & 3.19 & 17.27 & 4.80 & 2.90 & 14.06 & 13.66 & 0.54 \\ \hline
    \parbox[t]{2mm}{\multirow{5}{*}{\rotatebox[origin=c]{90}{General}}}                                             & ResNet-101 \cite{ResNet} & 54.47 & 38.08 & 54.47 & 40.63 & 82.62 & 43.58 & 28.98 & 87.83 & 39.96 & 76.27 \\ 
    & KSSNet \cite{KSSNet20} & 56.18 & 39.37 & 56.18 & 42.52 & 82.77 & 44.82 & 30.12 & 87.56 & 41.28 & 77.63 \\ 
    & TResNet-M \cite{TResNet20} & 55.27 & 38.69 & 55.27 & 41.22 & 83.88 & 44.14 & 29.35 & \textbf{88.93} & 41.07 & 78.29 \\ 
    & TResNet-L \cite{TResNet20} & 56.01 & 39.63 & 56.01 & 42.09 & 83.69 & 44.90 & 30.10 & 88.32 & 41.22 & 78.75 \\ 
    & TResNet-XL \cite{TResNet20} & 55.83 & 39.30 & 55.83 & 41.82 & 83.98 & 44.66 & 29.85 & 88.67 & 41.68 & 78.32 \\ \hline
    & \textit{Benchmark} & \textbf{61.45} & \textbf{42.39} & \textbf{61.45} & \textbf{47.02} & \textbf{88.67} & \textbf{46.38} & \textbf{32.25} & 82.55 & \textbf{51.65} & \textbf{79.79} \\ 
    \hline
    \end{tabular}
    \label{tab:mainResVal}
\end{table*}

\begin{table*}[!t]
    \centering
    \caption{\textbf{Per-class F1 score - Validation Split}. The metrics are presented as percentages, and the highest score in each column is denoted in bold.}
    
    \resizebox{\textwidth}{!}{%
    \begin{tabular}{clrrrrrrrrrrrrrrrrrr}\hline
    & \textbf{Model} & \textbf{RB} & \textbf{OB} & \textbf{PF} & \textbf{DE} & \textbf{FS} & \textbf{IS} & \textbf{RO} & \textbf{IN} & \textbf{AF} & \textbf{BE} & \textbf{FO} & \textbf{GR} & \textbf{PH} & \textbf{PB} & \textbf{OS} & \textbf{OP} & \textbf{OK}& \textbf{Normal} \\ \hline
\parbox[t]{2mm}{\multirow{4}{*}{\rotatebox[origin=c]{90}{Sewer}}}                                             & Xie \etal \cite{Xie2019HierCNN} & 22.97 & 71.40 & 35.83 & 22.84 & 77.12 & 9.77 & 17.94 & 28.39 & 40.27 & 38.09 & \textbf{5.84} & 49.03 & 37.86 & 24.21 & 13.80 & 29.03 & 70.29 & 91.08 \\ 
& Chen \etal \cite{Chen2018} & 24.60 & 57.19 & 17.17 & 10.08 & 68.37 & 6.03 & \textbf{31.78} & 21.05 & 26.71 & 39.01 & 5.33 & 25.26 & 34.27 & 8.79 & 10.10 & 32.34 & 57.18 & 3.96 \\ 
& Hassan \etal \cite{Hassan2019AlexNet} & 0.00 & 30.75 & 3.06 & 3.09 & 43.57 & 1.35 & 4.39 & 0.00 & 13.02 & 0.00 & 0.00 & 10.06 & 5.14 & 0.00 & 0.00 & 0.00 & 0.00 & 0.00 \\ 
& Myrans \etal \cite{Myrans2018MultiClass} & 1.61 & 5.70 & 1.79 & 1.07 & 8.07 & 0.28 & 0.53 & 0.84 & 3.29 & 2.56 & 0.20 & 3.07 & 0.94 & 0.94 & 0.28 & 0.14 & 9.16 & 26.03 \\ \hline
\parbox[t]{2mm}{\multirow{5}{*}{\rotatebox[origin=c]{90}{General}}}                                             & ResNet-101 \cite{ResNet} & 24.08 & 73.10 & 30.46 & 18.47 & 79.44 & 9.48 & 20.43 & 27.80 & 39.92 & 41.07 & 4.50 & 47.41 & 40.62 & 24.66 & 18.08 & 32.83 & 73.52 & 79.55 \\ 
& KSSNet \cite{KSSNet20} & \textbf{25.40} & 73.64 & 31.52 & 18.84 & 80.59 & 10.56 & 21.06 & 28.88 & 41.11 & 42.15 & 4.82 & 51.24 & 43.69 & 25.23 & 19.28 & 35.55 & 74.58 & 80.60 \\ 
& TResNet-M \cite{TResNet20} & 24.87 & 72.90 & 30.24 & 19.72 & 80.13 & 10.47 & 20.16 & 28.31 & 40.61 & 40.32 & 4.54 & 48.04 & 44.08 & 24.17 & 17.26 & 35.64 & 73.80 & 81.23 \\ 
& TResNet-L \cite{TResNet20} & 24.51 & 73.14 & 30.87 & 19.74 & 79.94 & 11.57 & 19.76 & 29.49 & 41.24 & 41.49 & 4.74 & 50.31 & 47.15 & 28.00 & 17.88 & 38.95 & 73.34 & 81.22 \\ 
& TResNet-XL \cite{TResNet20} & 24.75 & 73.15 & 32.20 & 20.58 & 79.89 & 10.21 & 19.76 & 29.09 & 40.30 & 41.15 & 4.69 & 48.35 & 45.22 & 26.45 & 18.23 & 37.08 & 74.57 & 81.81 \\ \hline
& \textit{Benchmark} & 24.81 & \textbf{74.50} & \textbf{36.39} & \textbf{23.91} & \textbf{80.69} & \textbf{11.87} & 20.05 & \textbf{31.19} & \textbf{42.39} & \textbf{42.63} & 4.94 & \textbf{51.40} & \textbf{48.53} & \textbf{33.09} & \textbf{21.58} & \textbf{48.75} & \textbf{75.00} & \textbf{91.32} \\ 
\hline

    \end{tabular}%
    }
    \label{tab:classF1Val}
\end{table*}

\begin{table*}[!t]
    \centering
    \caption{\textbf{Per-class F2 score - Validation Split}. The metrics are presented as percentages, and the highest score in each column is denoted in bold.}
    
    \resizebox{\textwidth}{!}{%
    \begin{tabular}{clrrrrrrrrrrrrrrrrrr}\hline
    & \textbf{Model} & \textbf{RB} & \textbf{OB} & \textbf{PF} & \textbf{DE} & \textbf{FS} & \textbf{IS} & \textbf{RO} & \textbf{IN} & \textbf{AF} & \textbf{BE} & \textbf{FO} & \textbf{GR} & \textbf{PH} & \textbf{PB} & \textbf{OS} & \textbf{OP} & \textbf{OK}& \textbf{Normal} \\ \hline
\parbox[t]{2mm}{\multirow{4}{*}{\rotatebox[origin=c]{90}{Sewer}}}                                             & Xie \etal \cite{Xie2019HierCNN} & 38.87 & 77.86 & 52.92 & 37.48 & 79.78 & 19.96 & 32.90 & 46.59 & 53.52 & 53.48 & \textbf{12.66} & 58.74 & 56.49 & 40.44 & 26.37 & 47.51 & 76.98 & 90.32 \\ 
& Chen \etal \cite{Chen2018} & 38.82 & 66.38 & 31.00 & 20.81 & 73.90 & 12.77 & \textbf{43.33} & 33.70 & 42.02 & 45.98 & 11.64 & 43.15 & 52.73 & 19.18 & 20.63 & 51.75 & 66.21 & 2.52 \\ 
& Hassan \etal \cite{Hassan2019AlexNet} & 0.00 & 52.60 & 7.32 & 7.37 & 65.87 & 3.30 & 10.29 & 0.00 & 27.24 & 0.00 & 0.00 & 21.86 & 11.94 & 0.00 & 0.00 & 0.00 & 0.00 & 0.0 \\ 
& Myrans \etal \cite{Myrans2018MultiClass} & 3.23 & 7.81 & 4.06 & 1.84 & 9.59 & 0.66 & 1.16 & 1.85 & 5.95 & 4.78 & 0.49 & 5.91 & 2.03 & 2.26 & 0.69 & 0.33 & 13.32 & 25.93 \\ \hline
\parbox[t]{2mm}{\multirow{5}{*}{\rotatebox[origin=c]{90}{General}}}                                             & ResNet-101 \cite{ResNet} & 42.45 & 84.34 & 51.08 & 35.34 & 87.49 & 19.98 & 37.81 & 47.47 & 59.18 & 59.87 & 10.39 & 64.78 & 61.24 & 44.03 & 34.81 & 54.23 & 82.99 & 71.60 \\ 
& KSSNet \cite{KSSNet20} & \textbf{43.74} & \textbf{84.64} & 52.24 & 35.92 & 87.45 & 21.76 & 38.67 & 48.54 & 59.89 & \textbf{60.81} & 11.08 & \textbf{67.40} & 63.94 & 44.73 & 36.60 & 57.05 & 83.30 & 72.95 \\ 
& TResNet-M \cite{TResNet20} & 43.55 & 84.49 & 51.02 & 37.23 & 87.79 & 21.75 & 37.57 & 47.93 & 60.01 & 59.99 & 10.51 & 65.61 & 64.43 & 43.62 & 33.70 & 57.04 & \textbf{83.71} & 73.71 \\ 
& TResNet-L \cite{TResNet20} & 43.08 & 84.39 & 51.75 & 37.39 & \textbf{87.81} & 23.50 & 37.03 & 49.20 & \textbf{60.10} & 60.60 & 10.91 & 67.05 & \textbf{66.64} & 48.24 & 34.53 & 60.12 & 83.59 & 73.69 \\ 
& TResNet-XL \cite{TResNet20} & 43.34 & 84.44 & 52.99 & 38.50 & 87.69 & 21.37 & 37.01 & 48.93 & 59.79 & 60.42 & 10.79 & 65.77 & 65.07 & 46.46 & 35.12 & 58.61 & 83.63 & 74.49 \\ \hline
& \textit{Benchmark} & 42.92 & 83.56 & \textbf{54.06} & \textbf{39.16} & 86.99 & \textbf{23.89} & 37.17 & \textbf{50.41} & 59.64 & 60.12 & 11.30 & 66.39 & 66.40 & \textbf{50.16} & \textbf{37.32} & \textbf{64.86} & 82.88 & \textbf{90.79} \\ 
\hline
    \end{tabular}%
    }
    \label{tab:classF2Val}
\end{table*}

\begin{table*}[!t]
    \centering
    \caption{\textbf{Per-class Precision score - Validation Split}. The metrics are presented as percentages, and the highest score in each column is denoted in bold.}
    
    \resizebox{\textwidth}{!}{%
    \begin{tabular}{clrrrrrrrrrrrrrrrrrr}\hline
    & \textbf{Model} & \textbf{RB} & \textbf{OB} & \textbf{PF} & \textbf{DE} & \textbf{FS} & \textbf{IS} & \textbf{RO} & \textbf{IN} & \textbf{AF} & \textbf{BE} & \textbf{FO} & \textbf{GR} & \textbf{PH} & \textbf{PB} & \textbf{OS} & \textbf{OP} & \textbf{OK}& \textbf{Normal} \\ \hline
\parbox[t]{2mm}{\multirow{4}{*}{\rotatebox[origin=c]{90}{Sewer}}}                                             & Xie \etal \cite{Xie2019HierCNN} & 13.66 & 62.73 & 23.30 & 13.83 & \textbf{73.05} & 5.28 & 10.21 & 17.20 & 28.51 & 25.75 & \textbf{3.08} & \textbf{38.45} & 24.43 & 14.51 & 7.69 & 17.61 & 61.39 & 92.36 \\ 
& Chen \etal \cite{Chen2018} & \textbf{15.27} & 46.48 & 9.85 & 5.42 & 60.78 & 3.20 & \textbf{22.00} & 12.94 & 16.62 & \textbf{31.15} & 2.80 & 14.93 & 21.64 & 4.62 & 5.46 & 19.90 & 46.59 & 91.88 \\ 
& Hassan \etal \cite{Hassan2019AlexNet} & 0.00 & 18.17 & 1.55 & 1.57 & 27.85 & 0.68 & 2.24 & 0.00 & 6.97 & 0.00 & 0.00 & 5.30 & 2.64 & 0.00 & 0.00 & 0.00 & 0.00 & 0.00 \\ 
& Myrans \etal \cite{Myrans2018MultiClass} & 0.88 & 3.93 & 0.93 & 0.63 & 6.38 & 0.14 & 0.28 & 0.44 & 1.88 & 1.44 & 0.10 & 1.71 & 0.49 & 0.47 & 0.14 & 0.07 & 6.03 & 26.21 \\ \hline
\parbox[t]{2mm}{\multirow{5}{*}{\rotatebox[origin=c]{90}{General}}}                                             & ResNet-101 \cite{ResNet} & 13.99 & 59.82 & 18.21 & 10.29 & 68.89 & 5.05 & 11.56 & 16.44 & 25.88 & 26.96 & 2.32 & 32.76 & 26.02 & 14.23 & 10.04 & 19.80 & 61.78 & 97.61 \\ 
& KSSNet \cite{KSSNet20} & 14.95 & 60.52 & 18.98 & 10.51 & 71.26 & 5.69 & 11.97 & 17.24 & 26.99 & 27.89 & 2.48 & 36.61 & 28.59 & 14.61 & 10.78 & 21.84 & 63.50 & 97.68 \\ 
& TResNet-M \cite{TResNet20} & 14.50 & 59.34 & 18.01 & 11.06 & 69.95 & 5.62 & 11.37 & 16.83 & 26.39 & 26.07 & 2.33 & 33.21 & 28.88 & 13.86 & 9.52 & 21.93 & 61.64 & 97.87 \\ 
& TResNet-L \cite{TResNet20} & 14.26 & 59.84 & 18.46 & 11.05 & 69.55 & 6.27 & 11.12 & 17.69 & 27.08 & 27.19 & 2.44 & 35.53 & 31.70 & 16.48 & 9.91 & 24.54 & 60.88 & \textbf{97.89} \\ 
& TResNet-XL \cite{TResNet20} & 14.43 & 59.81 & 19.47 & 11.59 & 69.58 & 5.46 & 11.12 & 17.36 & 26.12 & 26.87 & 2.41 & 33.55 & 29.98 & 15.40 & 10.12 & 23.00 & 63.16 & 97.86 \\ \hline
& \textit{Benchmark} & 14.56 & \textbf{63.11} & \textbf{23.56} & \textbf{14.50} & 72.00 & \textbf{6.46} & 11.35 & \textbf{19.07} & \textbf{28.60} & 28.71 & 2.55 & 37.34 & \textbf{33.50} & \textbf{21.11} & \textbf{12.67} & \textbf{34.49} & \textbf{64.74} & 92.21 \\ 
\hline

    \end{tabular}%
    }
    \label{tab:classPrcVal}
\end{table*}

\begin{table*}[!t]
    \centering
    \caption{\textbf{Per-class Recall score - Validation Split}. The metrics are presented as percentages, and the highest score in each column is denoted in bold.}
    
    \resizebox{\textwidth}{!}{%
    \begin{tabular}{clrrrrrrrrrrrrrrrrrr}\hline
    & \textbf{Model} & \textbf{RB} & \textbf{OB} & \textbf{PF} & \textbf{DE} & \textbf{FS} & \textbf{IS} & \textbf{RO} & \textbf{IN} & \textbf{AF} & \textbf{BE} & \textbf{FO} & \textbf{GR} & \textbf{PH} & \textbf{PB} & \textbf{OS} & \textbf{OP} & \textbf{OK}& \textbf{Normal} \\ \hline
\parbox[t]{2mm}{\multirow{4}{*}{\rotatebox[origin=c]{90}{Sewer}}}                                             & Xie \etal \cite{Xie2019HierCNN} & 72.16 & 82.86 & 77.59 & 65.46 & 81.67 & 65.49 & 74.08 & 81.33 & 68.54 & 73.19 & 57.29 & 67.67 & 84.06 & 73.07 & 67.18 & 82.52 & 82.20 & 89.83 \\ 
& Chen \etal \cite{Chen2018} & 63.16 & 74.34 & 66.95 & 71.59 & 78.12 & 50.28 & 57.18 & 56.26 & 68.01 & 52.19 & 54.94 & 81.80 & 82.28 & 90.20 & 67.61 & 86.27 & 74.01 & 2.03 \\ 
& Hassan \etal \cite{Hassan2019AlexNet} & 0.00 & \textbf{100.00} & \textbf{100.00} & \textbf{100.00} & \textbf{100.00} & \textbf{100.00} & \textbf{100.00} & 0.00 & \textbf{100.00} & 0.00 & 0.00 & \textbf{100.00} & \textbf{100.00} & 0.00 & 0.00 & 0.00 & 0.00 & 0.00 \\ 
& Myrans \etal \cite{Myrans2018MultiClass} & 9.84 & 10.37 & 26.37 & 3.58 & 10.98 & 9.88 & 5.90 & 9.67 & 12.94 & 11.33 & 9.05 & 15.42 & 8.97 & 38.56 & 19.47 & 5.72 & 19.09 & 25.86 \\ \hline
\parbox[t]{2mm}{\multirow{5}{*}{\rotatebox[origin=c]{90}{General}}}                                             & ResNet-101 \cite{ResNet} & 86.38 & 93.96 & 93.07 & 90.28 & 93.82 & 76.50 & 87.42 & \textbf{89.83} & 87.25 & 86.16 & 80.90 & 85.73 & 92.57 & 92.42 & 90.81 & \textbf{95.92} & 90.78 & 67.13 \\ 
& KSSNet \cite{KSSNet20} & 84.36 & 94.01 & 92.97 & 90.73 & 92.72 & 74.23 & 87.38 & 88.90 & 86.12 & 86.25 & 82.41 & 85.34 & 92.54 & 92.29 & 91.25 & 95.59 & 90.34 & 68.61 \\ 
& TResNet-M \cite{TResNet20} & \textbf{87.25} & 94.51 & 94.16 & 91.22 & 93.77 & 77.07 & 88.58 & 89.12 & 88.07 & \textbf{88.93} & \textbf{85.26} & 86.76 & 93.07 & \textbf{94.12} & \textbf{92.34} & 95.10 & 91.95 & 69.42 \\ 
& TResNet-L \cite{TResNet20} & 87.04 & 94.04 & 94.26 & 92.59 & 93.98 & 75.14 & 88.79 & 88.73 & 86.46 & 87.45 & 83.08 & 86.17 & 91.99 & 93.07 & 91.03 & 94.28 & \textbf{92.19} & 69.39 \\ 
& TResNet-XL \cite{TResNet20} & 86.85 & 94.13 & 93.02 & 91.81 & 93.80 & 78.89 & 88.58 & 89.69 & 88.22 & 87.84 & 82.24 & 86.56 & 91.99 & 93.73 & 91.90 & 95.59 & 91.00 & 70.29 \\ \hline
& \textit{Benchmark} & 83.66 & 90.93 & 79.91 & 68.11 & 91.76 & 73.55 & 86.25 & 85.56 & 81.84 & 82.75 & 79.23 & 82.42 & 88.02 & 76.47 & 72.65 & 83.17 & 89.12 & \textbf{90.44} \\ 
\hline

    \end{tabular}%
    }
    \label{tab:classRcllVal}
\end{table*}

\begin{table*}[!t]
    \centering
    \caption{\textbf{Per-class AP score - Validation Split}. The metrics are presented as percentages, and the highest score in each column is denoted in bold.}
    
    \resizebox{\textwidth}{!}{%
    \begin{tabular}{clrrrrrrrrrrrrrrrrr}\hline
    & \textbf{Model} & \textbf{RB} & \textbf{OB} & \textbf{PF} & \textbf{DE} & \textbf{FS} & \textbf{IS} & \textbf{RO} & \textbf{IN} & \textbf{AF} & \textbf{BE} & \textbf{FO} & \textbf{GR} & \textbf{PH} & \textbf{PB} & \textbf{OS} & \textbf{OP} & \textbf{OK} \\ \hline
\parbox[t]{2mm}{\multirow{4}{*}{\rotatebox[origin=c]{90}{Sewer}}}                                             & Xie \etal \cite{Xie2019HierCNN} & 29.82 & 83.81 & 83.82 & 63.25 & 87.48 & 31.80 & 62.03 & 54.60 & 60.44 & 66.43 & 48.39 & 84.66 & 78.56 & 68.85 & 61.24 & 77.41 & 86.25 \\ 
& Chen \etal \cite{Chen2018} & 48.30 & 73.98 & 49.98 & 45.03 & 83.03 & 56.31 & 78.12 & 45.85 & 61.99 & 71.01 & 42.94 & 74.62 & 78.27 & 56.95 & 51.00 & 56.89 & 80.80 \\ 
& Hassan \etal \cite{Hassan2019AlexNet} & 7.20 & 32.00 & 1.09 & 0.26 & 37.38 & 1.26 & 5.90 & 6.25 & 7.71 & 11.05 & 1.44 & 10.27 & 4.34 & 0.00 & 0.00 & 5.85 & 19.14 \\ 
& Myrans \etal \cite{Myrans2018MultiClass} & 0.05 & 0.62 & 0.10 & 0.64 & 2.57 & 0.13 & 0.12 & 0.19 & 1.02 & 0.28 & 0.00 & 0.35 & 0.00 & 1.17 & 0.00 & 0.00 & 1.96 \\ \hline
\parbox[t]{2mm}{\multirow{5}{*}{\rotatebox[origin=c]{90}{General}}}                                             & ResNet-101 \cite{ResNet} & 54.54 & 90.21 & 84.15 & 76.73 & 93.56 & 49.33 & 81.88 & 67.13 & 74.85 & 80.20 & 64.11 & 90.83 & 87.63 & 66.49 & 59.77 & 81.58 & 93.57 \\ 
& KSSNet \cite{KSSNet20} & 56.86 & 90.74 & 85.42 & 76.50 & 94.05 & 56.75 & \textbf{83.43} & 68.86 & 75.14 & 81.40 & \textbf{65.51} & 91.27 & 89.20 & 66.49 & 64.58 & 79.66 & 93.87 \\ 
& TResNet-M \cite{TResNet20} & \textbf{57.22} & 90.90 & 87.74 & 77.69 & 93.98 & 58.68 & 80.56 & \textbf{69.94} & \textbf{76.17} & \textbf{82.39} & 60.67 & 91.55 & \textbf{89.90} & 69.27 & 65.58 & 84.39 & 94.32 \\ 
& TResNet-L \cite{TResNet20} & 56.76 & 90.75 & 88.32 & 78.36 & 93.95 & 60.42 & 80.88 & 69.21 & 75.64 & 81.99 & 64.62 & 91.37 & 89.38 & 69.75 & 69.39 & 83.57 & 94.44 \\ 
& TResNet-XL \cite{TResNet20} & 57.15 & 90.81 & 87.36 & 78.34 & 94.04 & 56.91 & 80.92 & 69.84 & 76.01 & 82.00 & 63.28 & 91.69 & 88.97 & 69.66 & 68.16 & 82.09 & 94.23 \\ \hline
& \textit{Benchmark} & 56.68 & \textbf{90.93} & \textbf{90.12} & \textbf{80.30} & \textbf{94.06} & \textbf{60.55} & 80.79 & 69.45 & 75.99 & 82.27 & 65.32 & \textbf{92.06} & 89.89 & \textbf{75.70} & \textbf{72.97} & \textbf{84.81} & \textbf{94.57} \\ 
\hline
    \end{tabular}%
    }
    \label{tab:classAPVal}
\end{table*}

\clearpage

\begin{table*}[!t]
    \centering
    \caption{\textbf{Performance metrics for each method - Test Split.} The metrics are presented as percentages, and the highest score in each column is denoted in bold.}
    \begin{tabular}{clrrrrrrrrrr} \hline
    & \textbf{Model} & \textbf{m-F1} & \textbf{M-F1} & \textbf{OV-F1} & \textbf{OV-P} & \textbf{OV-R} & \textbf{PC-F1} & \textbf{PC-P} & \textbf{PC-R}& \textbf{0-1} & \textbf{mAP} \\ \hline
\parbox[t]{2mm}{\multirow{4}{*}{\rotatebox[origin=c]{90}{Sewer}}}                                             & Xie \etal \cite{Xie2019HierCNN} & 59.05 & 37.94 & 59.05 & 46.06 & 82.24 & 42.16 & 29.48 & 73.95 & 51.55 & 65.32 \\ 
& Chen \etal \cite{Chen2018} & 33.49 & 26.23 & 33.49 & 26.03 & 46.94 & 34.55 & 23.60 & 64.48 & 7.63 & 59.89 \\ 
& Hassan \etal \cite{Hassan2019AlexNet} & 12.57 & 6.27 & 12.57 & 7.33 & 44.12 & 6.83 & 3.67 & 50.00 & 0.00 & 7.35 \\ 
& Myrans \etal \cite{Myrans2018MultiClass} & 5.66 & 3.88 & 5.66 & 3.36 & 18.07 & 5.02 & 3.04 & 14.43 & 14.51 & 0.59 \\ \hline
\parbox[t]{2mm}{\multirow{5}{*}{\rotatebox[origin=c]{90}{General}}}                                             & ResNet-101 \cite{ResNet} & 53.91 & 37.94 & 53.91 & 40.19 & 81.85 & 43.46 & 28.89 & 87.70 & 39.38 & 74.99 \\ 
& KSSNet \cite{KSSNet20} & 55.64 & 39.22 & 55.64 & 42.12 & 81.96 & 44.68 & 30.02 & 87.32 & 40.46 & 75.70 \\ 
& TResNet-M \cite{TResNet20} & 54.62 & 38.53 & 54.62 & 40.72 & 82.94 & 43.96 & 29.24 & \textbf{88.56} & 40.23 & 76.55 \\ 
& TResNet-L \cite{TResNet20} & 55.34 & 39.45 & 55.34 & 41.56 & 82.79 & 44.72 & 29.97 & 88.05 & 40.42 & 76.82 \\ 
& TResNet-XL \cite{TResNet20} & 55.08 & 38.98 & 55.08 & 41.21 & 83.01 & 44.34 & 29.61 & 88.23 & 40.74 & 76.61 \\ \hline
& \textit{Benchmark} & \textbf{61.26} & \textbf{42.35} & \textbf{61.26} & \textbf{46.90} & \textbf{88.30} & \textbf{46.22} & \textbf{32.24} & 81.61 & \textbf{51.59} & \textbf{77.79} \\ 
\hline
\end{tabular}
    \label{tab:mainResTest}
\end{table*}

\begin{table*}[!t]
    \centering
    \caption{\textbf{Per-class F1 score - Test Split}. The metrics are presented as percentages, and the highest score in each column is denoted in bold.}
    
    \resizebox{\textwidth}{!}{%
    \begin{tabular}{clrrrrrrrrrrrrrrrrrr}\hline
    & \textbf{Model} & \textbf{RB} & \textbf{OB} & \textbf{PF} & \textbf{DE} & \textbf{FS} & \textbf{IS} & \textbf{RO} & \textbf{IN} & \textbf{AF} & \textbf{BE} & \textbf{FO} & \textbf{GR} & \textbf{PH} & \textbf{PB} & \textbf{OS} & \textbf{OP} & \textbf{OK}& \textbf{Normal} \\ \hline
\parbox[t]{2mm}{\multirow{4}{*}{\rotatebox[origin=c]{90}{Sewer}}}                                             & Xie \etal \cite{Xie2019HierCNN} & 23.47 & 71.12 & 34.63 & 23.51 & 77.25 & 9.99 & 16.14 & 31.33 & 39.59 & 39.93 & \textbf{6.29} & 49.22 & 34.61 & 24.09 & 14.27 & 27.05 & 69.85 & 90.62 \\ 
& Chen \etal \cite{Chen2018} & 24.54 & 57.24 & 16.85 & 11.65 & 67.47 & 6.09 & \textbf{29.99} & 23.03 & 26.47 & 37.67 & 5.62 & 24.84 & 31.53 & 8.58 & 11.01 & 29.58 & 56.46 & 3.59 \\ 
& Hassan \etal \cite{Hassan2019AlexNet} & 0.00 & 30.35 & 2.95 & 3.49 & 43.16 & 1.41 & 4.04 & 0.00 & 13.19 & 0.00 & 0.00 & 9.84 & 4.45 & 0.00 & 0.00 & 0.00 & 0.00 & 0.00 \\ 
& Myrans \etal \cite{Myrans2018MultiClass} & 1.63 & 5.60 & 1.71 & 1.46 & 8.18 & 0.36 & 0.60 & 1.02 & 4.03 & 2.93 & 0.21 & 3.16 & 0.82 & 0.89 & 0.31 & 0.16 & 9.28 & 27.48 \\ \hline
\parbox[t]{2mm}{\multirow{5}{*}{\rotatebox[origin=c]{90}{General}}}                                             & ResNet-101 \cite{ResNet} & 24.42 & 72.52 & 28.62 & 19.75 & 79.22 & 9.98 & 19.07 & 30.13 & 38.93 & 42.15 & 4.87 & 47.69 & 37.80 & 26.86 & 18.56 & 30.54 & 73.28 & 78.57 \\ 
& KSSNet \cite{KSSNet20} & \textbf{26.06} & 73.34 & 29.89 & 19.64 & \textbf{80.56} & 10.83 & 19.84 & 31.47 & 40.59 & 43.50 & 5.24 & 50.88 & 40.74 & 26.39 & 20.55 & 32.84 & 74.38 & 79.29 \\ 
& TResNet-M \cite{TResNet20} & 24.78 & 72.54 & 28.94 & 20.96 & 79.87 & 10.89 & 18.52 & 31.14 & 39.64 & 41.39 & 4.90 & 47.97 & 40.11 & 24.99 & 18.34 & 34.68 & 73.90 & 79.91 \\ 
& TResNet-L \cite{TResNet20} & 24.78 & 72.93 & 28.68 & 20.62 & 79.60 & 12.01 & 18.20 & 32.29 & 40.43 & 42.56 & 5.11 & 49.97 & 43.33 & 28.13 & 19.43 & 38.68 & 73.41 & 79.88 \\ 
& TResNet-XL \cite{TResNet20} & 24.76 & 72.66 & 30.24 & 21.49 & 79.71 & 10.46 & 18.32 & 31.51 & 39.59 & 41.94 & 5.13 & 48.32 & 41.21 & 27.12 & 19.25 & 35.10 & 74.41 & 80.42 \\ \hline
& \textit{Benchmark} & 25.11 & \textbf{74.40} & \textbf{35.58} & \textbf{25.01} & 80.50 & \textbf{12.26} & 18.59 & \textbf{34.26} & \textbf{41.93} & \textbf{44.16} & 5.26 & \textbf{51.28} & \textbf{45.09} & \textbf{31.60} & \textbf{22.20} & \textbf{49.17} & \textbf{75.04} & \textbf{90.94} \\ 
\hline
\end{tabular}%
    }
    \label{tab:classf1Test}
\end{table*}

\begin{table*}[!t]
    \centering
    \caption{\textbf{Per-class F2 score - Test Split}. The metrics are presented as percentages, and the highest score in each column is denoted in bold.}
    
    \resizebox{\textwidth}{!}{%
    \begin{tabular}{clrrrrrrrrrrrrrrrrrr}\hline
    & \textbf{Model} & \textbf{RB} & \textbf{OB} & \textbf{PF} & \textbf{DE} & \textbf{FS} & \textbf{IS} & \textbf{RO} & \textbf{IN} & \textbf{AF} & \textbf{BE} & \textbf{FO} & \textbf{GR} & \textbf{PH} & \textbf{PB} & \textbf{OS} & \textbf{OP} & \textbf{OK}& \textbf{Normal} \\ \hline
\parbox[t]{2mm}{\multirow{4}{*}{\rotatebox[origin=c]{90}{Sewer}}}                                             & Xie \etal \cite{Xie2019HierCNN} & 39.77 & 77.88 & 51.49 & 37.53 & 79.90 & 20.42 & 30.19 & 49.28 & 53.16 & 54.67 & \textbf{13.50 }& 59.40 & 53.74 & 39.57 & 26.30 & 45.24 & 76.63 & 89.77 \\ 
& Chen \etal \cite{Chen2018} & 38.91 & 66.52 & 29.90 & 23.33 & 73.08 & 12.94 & \textbf{42.11} & 35.50 & 41.72 & 44.01 & 12.11 & 42.63 & 50.53 & 18.73 & 21.87 & 48.37 & 65.92 & 2.28 \\ 
& Hassan \etal \cite{Hassan2019AlexNet} & 0.00 & 52.14 & 7.07 & 8.28 & 65.50 & 3.45 & 9.53 & 0.00 & 27.53 & 0.00 & 0.00 & 21.43 & 10.44 & 0.00 & 0.00 & 0.00 & 0.00 & 0.0 \\ 
& Myrans \etal \cite{Myrans2018MultiClass} & 3.27 & 7.70 & 3.89 & 2.47 & 9.75 & 0.87 & 1.33 & 2.22 & 7.27 & 5.35 & 0.51 & 6.11 & 1.80 & 2.15 & 0.77 & 0.38 & 13.51 & 27.35 \\ \hline
\parbox[t]{2mm}{\multirow{5}{*}{\rotatebox[origin=c]{90}{General}}}                                             & ResNet-101 \cite{ResNet} & 42.95 & 83.92 & 48.25 & 37.06 & 87.22 & 21.11 & 35.87 & 50.06 & 58.23 & 60.37 & 11.18 & 64.81 & 58.83 & 47.12 & 35.24 & 51.76 & 82.63 & 70.41 \\ 
& KSSNet \cite{KSSNet20} & \textbf{44.79} & \textbf{84.52} & 49.68 & 36.93 & 87.36 & 22.42 & 36.96 & 51.45 & 59.54 & \textbf{61.55} & 11.96 & \textbf{66.99} & 61.57 & 46.35 & \textbf{37.92} & 54.59 & 83.00 & 71.32 \\ 
& TResNet-M \cite{TResNet20} & 43.39 & 84.36 & 48.99 & 38.84 & \textbf{87.48} & 22.70 & 35.12 & 51.09 & 59.13 & 60.27 & 11.27 & 65.41 & 60.99 & 44.71 & 35.04 & 56.55 & \textbf{83.72} & 72.08 \\ 
& TResNet-L \cite{TResNet20} & 43.50 & 84.42 & 48.55 & 38.39 & 87.45 & 24.45 & 34.67 & 52.27 & \textbf{59.55} & 61.13 & 11.70 & 66.37 & 63.54 & \textbf{48.58} & 36.69 & 60.42 & 83.50 & 72.03 \\ 
& TResNet-XL \cite{TResNet20} & 43.39 & 84.22 & 50.14 & 39.42 & 87.43 & 22.00 & 34.89 & 51.39 & 59.15 & 60.64 & 11.74 & 65.47 & 61.83 & 47.45 & 36.31 & 56.76 & 83.52 & 72.74 \\ \hline
& \textit{Benchmark} & 43.35 & 83.82 & \textbf{52.94} & \textbf{39.69} & 86.76 & \textbf{24.70} & 34.96 & \textbf{53.41} & 59.45 & 61.05 & 11.94 & 66.05 & \textbf{64.00} & 47.39 & 36.79 & \textbf{65.41} & 82.72 & \textbf{90.35} \\ 
\hline
    \end{tabular}%
    }
    \label{tab:classF2test}
\end{table*}

\begin{table*}[!t]
    \centering
    \caption{\textbf{Per-class Precision score - Test Split}. The metrics are presented as percentages, and the highest score in each column is denoted in bold.}
    
    \resizebox{\textwidth}{!}{%
    \begin{tabular}{clrrrrrrrrrrrrrrrrrr}\hline
    & \textbf{Model} & \textbf{RB} & \textbf{OB} & \textbf{PF} & \textbf{DE} & \textbf{FS} & \textbf{IS} & \textbf{RO} & \textbf{IN} & \textbf{AF} & \textbf{BE} & \textbf{FO} & \textbf{GR} & \textbf{PH} & \textbf{PB} & \textbf{OS} & \textbf{OP} & \textbf{OK}& \textbf{Normal} \\ \hline
    \parbox[t]{2mm}{\multirow{4}{*}{\rotatebox[origin=c]{90}{Sewer}}}                                             & Xie \etal \cite{Xie2019HierCNN} & 13.95 & 62.14 & 22.40 & 14.49 & \textbf{73.20} & 5.40 & 9.09 & 19.50 & 27.77 & 27.54 & \textbf{3.33} & \textbf{38.28} & 21.72 & 14.58 & 8.10 & 16.20 & 60.87 & 92.09 \\ 
& Chen \etal \cite{Chen2018} & 15.19 & 46.43 & 9.75 & 6.35 & 59.82 & 3.24 & \textbf{20.27} & 14.52 & 16.45 & \textbf{30.38} & 2.97 & 14.65 & 19.38 & 4.51 & 6.02 & 17.96 & 45.57 & 91.35 \\ 
& Hassan \etal \cite{Hassan2019AlexNet} & 0.00 & 17.89 & 1.50 & 1.77 & 27.52 & 0.71 & 2.06 & 0.00 & 7.06 & 0.00 & 0.00 & 5.17 & 2.28 & 0.00 & 0.00 & 0.00 & 0.00 & 0.00 \\ 
& Myrans \etal \cite{Myrans2018MultiClass} & 0.89 & 3.86 & 0.88 & 0.87 & 6.46 & 0.18 & 0.31 & 0.54 & 2.32 & 1.67 & 0.11 & 1.75 & 0.43 & 0.45 & 0.16 & 0.08 & 6.09 & 27.70 \\ \hline
\parbox[t]{2mm}{\multirow{5}{*}{\rotatebox[origin=c]{90}{General}}}                                             & ResNet-101 \cite{ResNet} & 14.20 & 59.14 & 17.06 & 11.10 & 68.72 & 5.31 & 10.71 & 18.11 & 25.08 & 28.04 & 2.51 & 33.12 & 23.69 & 15.65 & 10.37 & 18.14 & 61.65 & 97.37 \\ 
& KSSNet \cite{KSSNet20} & \textbf{15.36} & 60.10 & 17.96 & 11.03 & 71.31 & 5.82 & 11.20 & 19.11 & 26.52 & 29.22 & 2.71 & 36.33 & 26.05 & 15.36 & 11.66 & 19.73 & 63.39 & 97.46 \\ 
& TResNet-M \cite{TResNet20} & 14.45 & 58.81 & 17.21 & 11.86 & 69.76 & 5.83 & 10.36 & 18.86 & 25.58 & 27.19 & 2.52 & 33.21 & 25.54 & 14.40 & 10.23 & 21.09 & 61.81 & 97.59 \\ 
& TResNet-L \cite{TResNet20} & 14.43 & 59.44 & 17.05 & 11.64 & 69.23 & 6.50 & 10.15 & 19.73 & 26.33 & 28.26 & 2.64 & 35.40 & 28.32 & 16.53 & 10.89 & 24.18 & 61.10 & \textbf{97.62} \\ 
& TResNet-XL \cite{TResNet20} & 14.43 & 59.12 & 18.20 & 12.22 & 69.48 & 5.58 & 10.23 & 19.16 & 25.52 & 27.70 & 2.65 & 33.64 & 26.48 & 15.83 & 10.80 & 21.45 & 62.96 & 97.58 \\ \hline
& \textit{Benchmark} & 14.76 & \textbf{62.66} & \textbf{23.01} & \textbf{15.47} & 71.87 & \textbf{6.66} & 10.44 & \textbf{21.44} & \textbf{28.11} & 30.22 & 2.72 & 37.36 & \textbf{30.22} & \textbf{20.32} & \textbf{13.37} & \textbf{34.79} & \textbf{64.99} & 91.94 \\ 
\hline
    \end{tabular}%
    }
    \label{tab:classPrcTest}
\end{table*}

\begin{table*}[!t]
    \centering
    \caption{\textbf{Per-class Recall score - Test Split}. The metrics are presented as percentages, and the highest score in each column is denoted in bold.}
    
    \resizebox{\textwidth}{!}{%
    \begin{tabular}{clrrrrrrrrrrrrrrrrrr}\hline
    & \textbf{Model} & \textbf{RB} & \textbf{OB} & \textbf{PF} & \textbf{DE} & \textbf{FS} & \textbf{IS} & \textbf{RO} & \textbf{IN} & \textbf{AF} & \textbf{BE} & \textbf{FO} & \textbf{GR} & \textbf{PH} & \textbf{PB} & \textbf{OS} & \textbf{OP} & \textbf{OK}& \textbf{Normal} \\ \hline
\parbox[t]{2mm}{\multirow{4}{*}{\rotatebox[origin=c]{90}{Sewer}}}                                             & Xie \etal \cite{Xie2019HierCNN} & 74.02 & 83.15 & 76.24 & 62.29 & 81.77 & 66.99 & 71.98 & 79.72 & 68.91 & 72.53 & 57.16 & 68.90 & 85.11 & 69.27 & 60.00 & 81.99 & 81.93 & 89.21 \\ 
& Chen \etal \cite{Chen2018} & 63.81 & 74.59 & 61.83 & 70.35 & 77.37 & 51.41 & 57.64 & 55.55 & 67.74 & 49.56 & 52.39 & 81.56 & 84.47 & 88.48 & 63.96 & 83.86 & 74.21 & 1.83 \\ 
& Hassan \etal \cite{Hassan2019AlexNet} & 0.00 & \textbf{100.00} & \textbf{100.00} & \textbf{100.00} & \textbf{100.00} & \textbf{100.00} & \textbf{100.00} & 0.00 & \textbf{100.00} & 0.00 & 0.00 & \textbf{100.00} & \textbf{100.00} & 0.00 & 0.00 & 0.00 & 0.00 & 0.00 \\ 
& Myrans \etal \cite{Myrans2018MultiClass} & 9.96 & 10.26 & 26.37 & 4.55 & 11.17 & 12.34 & 7.19 & 10.26 & 15.62 & 11.87 & 8.63 & 16.13 & 8.98 & 33.61 & 18.49 & 7.69 & 19.42 & 27.26 \\ \hline
\parbox[t]{2mm}{\multirow{5}{*}{\rotatebox[origin=c]{90}{General}}}                                             & ResNet-101 \cite{ResNet} & 86.91 & 93.73 & 88.92 & 89.16 & 93.52 & 82.47 & 86.85 & \textbf{89.55} & 86.96 & 84.82 & 81.66 & 85.19 & 93.52 & 94.72 & 87.92 & 96.44 & 90.31 & 65.85 \\ 
& KSSNet \cite{KSSNet20} & 85.98 & 94.07 & 88.92 & 89.34 & 92.58 & 78.25 & 87.07 & 89.21 & 86.46 & 85.09 & 81.66 & 84.91 & 93.42 & 93.52 & 86.79 & \textbf{97.75} & 89.96 & 66.83 \\ 
& TResNet-M \cite{TResNet20} & 86.95 & 94.64 & 91.02 & 90.03 & 93.42 & 82.03 & 87.22 & 89.21 & 87.95 & \textbf{86.62} & \textbf{84.75} & 86.34 & 93.38 & 94.36 & 89.06 & 97.56 & 91.85 & 67.66 \\ 
& TResNet-L \cite{TResNet20} & \textbf{87.64} & 94.34 & 90.20 & 90.16 & 93.61 & 79.00 & 87.48 & 88.93 & 86.99 & 86.19 & 82.74 & 84.94 & 92.20 & 94.24 & \textbf{90.00} & 96.62 & \textbf{91.93} & 67.60 \\ 
& TResNet-XL \cite{TResNet20} & 87.04 & 94.22 & 89.33 & 88.86 & 93.47 & 83.23 & 87.89 & 88.72 & 88.22 & 86.27 & 83.05 & 85.74 & 92.81 & \textbf{94.84} & 88.68 & 96.44 & 90.94 & 68.39 \\ \hline
& \textit{Benchmark} & 84.04 & 91.54 & 78.45 & 65.19 & 91.50 & 76.52 & 84.72 & 85.16 & 82.42 & 81.95 & 77.81 & 81.74 & 88.83 & 71.07 & 65.47 & 83.86 & 88.78 & \textbf{89.96} \\ 
\hline

    \end{tabular}%
    }
    \label{tab:classRclltest}
\end{table*}

\begin{table*}[!t]
    \centering
    \caption{\textbf{Per-class AP score - Test Split}. The metrics are presented as percentages, and the highest score in each column is denoted in bold.}
    
    \resizebox{\textwidth}{!}{%
    \begin{tabular}{clrrrrrrrrrrrrrrrrr}\hline
    & \textbf{Model} & \textbf{RB} & \textbf{OB} & \textbf{PF} & \textbf{DE} & \textbf{FS} & \textbf{IS} & \textbf{RO} & \textbf{IN} & \textbf{AF} & \textbf{BE} & \textbf{FO} & \textbf{GR} & \textbf{PH} & \textbf{PB} & \textbf{OS} & \textbf{OP} & \textbf{OK} \\ \hline
\parbox[t]{2mm}{\multirow{4}{*}{\rotatebox[origin=c]{90}{Sewer}}}                                             & Xie \etal \cite{Xie2019HierCNN} & 35.07 & 83.48 & 85.86 & 62.97 & 87.30 & 36.64 & 59.04 & 58.52 & 58.15 & 62.44 & 36.26 & 82.11 & 80.02 & 65.49 & 53.70 & 77.47 & 85.85 \\ 
& Chen \etal \cite{Chen2018} & 48.49 & 74.08 & 48.09 & 47.43 & 81.76 & 57.95 & 74.70 & 48.77 & 60.86 & 65.30 & 31.56 & 74.91 & 80.58 & 43.37 & 49.87 & 49.86 & 80.61 \\ 
& Hassan \etal \cite{Hassan2019AlexNet} & 6.16 & 26.79 & 0.33 & 3.60 & 35.10 & 1.17 & 1.25 & 3.16 & 5.62 & 5.14 & 0.44 & 9.23 & 4.17 & 0.00 & 1.57 & 2.52 & 18.69 \\ 
& Myrans \etal \cite{Myrans2018MultiClass} & 0.11 & 0.69 & 0.00 & 0.63 & 2.91 & 0.00 & 0.09 & 0.16 & 1.80 & 0.43 & 0.18 & 0.69 & 0.00 & 0.00 & 0.00 & 0.24 & 2.09 \\  \hline
\parbox[t]{2mm}{\multirow{5}{*}{\rotatebox[origin=c]{90}{General}}}                                             & ResNet-101 \cite{ResNet} & 55.25 & 90.20 & 88.04 & 71.96 & 93.32 & 65.63 & 78.98 & 65.69 & 71.40 & 77.78 & 47.33 & 90.72 & 88.05 & 65.34 & 52.55 & 79.33 & 93.32 \\ 
& KSSNet \cite{KSSNet20} & \textbf{58.43} & \textbf{90.59} & 86.80 & 71.99 & 93.68 & 69.97 & \textbf{80.75} & 68.28 & 72.57 & 78.92 & 44.03 & 91.11 & 88.46 & 62.31 & 55.86 & 79.26 & 93.83 \\ 
& TResNet-M \cite{TResNet20} & 55.61 & 90.16 & 89.82 & \textbf{76.00} & 93.65 & 65.85 & 78.58 & 69.71 & 73.45 & 79.82 & 50.44 & 91.03 & \textbf{89.41} & 67.57 & 57.79 & 78.11 & 94.32 \\ 
& TResNet-L \cite{TResNet20} & 56.95 & 90.38 & 89.28 & 75.03 & 93.64 & 68.61 & 80.04 & 70.09 & 73.59 & 79.43 & \textbf{48.74} & 91.36 & 88.77 & 67.29 & 59.38 & 79.00 & 94.40 \\ 
& TResNet-XL \cite{TResNet20} & 56.64 & 90.00 & 89.17 & 75.66 & 93.68 & 63.87 & 79.70 & 68.90 & 73.62 & 79.80 & 48.14 & 91.33 & 88.83 & 69.51 & 59.38 & 79.96 & 94.17 \\ \hline
& \textit{Benchmark} & 56.99 & 90.46 & \textbf{89.89} & 75.09 & \textbf{93.70} & \textbf{70.74} & 80.20 & \textbf{70.81} & \textbf{73.99} & \textbf{79.96} & 48.21 & \textbf{92.21} & 89.08 & \textbf{72.70} & \textbf{62.57} & \textbf{81.33} & \textbf{94.55} \\ 
\hline
    \end{tabular}%
    }
    \label{tab:classAPTest}
\end{table*}

\clearpage
{\small
\bibliographystyle{ieee_fullname}
\bibliography{main}
}

\end{document}